%% file: iclr2026_conference.tex
\documentclass{article} 
\usepackage{iclr2026_conference,times}

\input{math_commands.tex}

\usepackage[toc,page,header]{appendix}
\usepackage{amssymb}
\usepackage{hyperref}
\usepackage{minitoc}
\hypersetup{
    colorlinks,
    linkcolor=blue,
    citecolor=magenta,
    urlcolor=blue
}
\usepackage{url}
\usepackage{graphicx}
\usepackage{subfigure}
\usepackage[ruled]{algorithm2e}
\usepackage{multirow}
\usepackage{xcolor}
\usepackage{tcolorbox}
\usepackage{bbm}
\usepackage{booktabs}

\title{\textbf{GaLLoP}: \textbf{G}radient-based Sp\textbf{a}rse \textbf{L}earning on \textbf{Lo}w-Magnitude \textbf{P}arameters}

\author{Anand Choudhary$^{1,2}$ \quad Yasser Sulaıman$^{1,3}$ \quad Lukas Mauch$^1$ \quad Ghouthi Boukli Hacene$^1$\\
\textbf{Fabien Cardinaux}$^1$ \quad \hspace{0.9em}\textbf{Antoine Bosselut}$^2$\\
$^1$Sony Europe Ltd., Stuttgart Technology Center, EUREC \quad $^2$EPFL, Switzerland\\
$^3$University of Stuttgart, Germany\\
\texttt{anand.choudhary@alumni.epfl.ch, antoine.bosselut@epfl.ch}\\
\texttt{\{lukas.mauch, ghouthi.bouklihacene, fabien.cardinaux\}@sony.com}\\
\texttt{sulaiman17@itu.edu.tr}
}

\iclrfinalcopy 
\begin{document}
\doparttoc 
\faketableofcontents 
\maketitle

\begin{abstract}
Sparse fine-tuning techniques adapt LLMs to downstream tasks by only tuning a sparse subset of model parameters. However, the effectiveness of sparse adaptation depends on optimally selecting the model parameters to be fine-tuned. In this work, we introduce a novel sparse fine-tuning technique named \textbf{GaLLoP}: \textbf{G}radient-based Sp\textbf{a}rse \textbf{L}earning on \textbf{Lo}w-Magnitude \textbf{P}arameters, which fine-tunes only those model parameters which have the largest gradient magnitudes on downstream tasks and the smallest pre-trained magnitudes, intuitively prioritizing parameters that are highly task-relevant, but minimally disruptive to pre-trained knowledge. Our experimentation with LLaMA3 8B and Gemma 2B as base models shows that GaLLoP consistently improves or matches the in-distribution as well as out-of-distribution performance obtained via the usage of other leading parameter-efficient fine-tuning techniques, including LoRA, DoRA, and SAFT. Our analysis demonstrates that GaLLoP mitigates catastrophic forgetting and memorization of task data, as important pre-trained parameters remain unchanged, and stabilizes performance relative to other fine-tuning techniques, robustly generalizing across most random seeds.
\end{abstract}

\section{Introduction}

Noting that pre-trained LLMs have a low intrinsic dimension~\citep{intrinsic_dim_2}, efficient reparametrization of fine-tuning via a low-rank decomposition made techniques such as LoRA, DoRA, and their variants~\citep{lora, dora, vera} gain widespread popularity. However, such techniques are still susceptible to overfitting~\citep{adalora, lora_dropout} and hence, fine-tuning models using them can also result in the \emph{memorization} of patterns in the training dataset(s) and the loss of pre-trained knowledge, i.e., \emph{catastrophic forgetting}~\citep{cfr}. This can further impair their \emph{generalizability}, i.e., their ability to perform well on tasks related to yet unseen during fine-tuning, not only for cases wherein the task data used for testing possesses the same distribution as the fine-tuning data and differs only in content (\emph{In-Distribution} (ID)) but also for cases wherein the test data and the fine-tuning data possess different distributions altogether (\emph{Out-of-Distribution} (OOD))~\citep{ID_OOD_study}. Sparse Fine-Tuning (SpFT) techniques~\citep{spiel,saft} have recently emerged as promising alternatives to overcome these issues since they leverage the low intrinsic dimensionality of LLMs by directly fine-tuning only a small fraction of the original model parameters without introducing any additional parameters. However, the effectiveness of sparse adaptation crucially depends upon the selection criterion used to decide which parameters to update. 

In this work, we introduce a novel SpFT technique named \textbf{GaLLoP}: \textbf{G}radient-based Sp\textbf{a}rse \textbf{L}earning on \textbf{Lo}w-Magnitude \textbf{P}arameters. To enhance both ID as well as OOD generalizability, GaLLoP fine-tunes only a selected fraction of the model parameters which have the largest gradients on the downstream task (indicating strong ID task relevance~\citep{saft}) and the smallest pre-trained magnitudes (preserving the pre-trained knowledge~\citep{attn_small_wts,blockllm} for OOD tasks). 

Through experiments on eight datasets with LLaMA3 8B~\citep{llama} and Gemma 2B~\citep{gemma} as base models, we show that fine-tuning models with GaLLoP consistently enhances both their ID and OOD generalizability by preventing catastrophic forgetting and memorization as compared to state-of-the-art (SOTA) PEFT and post-training model editing techniques. Furthermore, we show that GaLLoP demonstrates robustness to overtraining and stabilizes performance.

\section{Related Work}

\subsection{Parameter-Efficient Fine-Tuning (PEFT)}
Given the large number of trainable parameters in LLMs, PEFT techniques facilitate the compute- and memory-efficient adaptation of LLMs to downstream tasks by only updating a small number of model parameters while keeping the rest frozen~\citep{adapter_orig}. They can be broadly classified into the following three categories:

\textbf{Additive Fine-Tuning (AFT):} Additional modules (\textit{adapter} layers) are connected to or introduced into the original \textit{frozen} LLM and these new modules (with a lower number of parameters than the original model) are then fine-tuned~\citep{adapter_orig, adapter_new}. However, their sequential processing introduces unwanted latency during training as well as inference~\citep{lora}.

\textbf{Reparametrized Fine-Tuning (RFT):} A low-rank decomposition-based reparametrization of the update matrix can be performed with almost the same effectiveness as the original full-rank representation~\citep{lora}. The fine-tuned parameters can then be merged with the pre-trained parameters prior to inference, doing away with any additional latency as in AFT. This category includes \textbf{LoRA}~\citep{lora}, the original RFT technique; \textbf{DoRA}, which fine-tunes magnitude and directional components separately to close LoRA's performance gap with Full Fine-Tuning (FFT)~\citep{dora}; as well as other variants which either increase LoRA's efficiency~\citep{vera,dylora} or its expressivity~\citep{hira,lora_plus,rosa}. However, in contrast to GaLLoP, these techniques are still susceptible to overfitting~\citep{adalora, lora_dropout,lines} since all the newly introduced parameters are fine-tuned on the downstream task. 

\textbf{Sparse Fine-Tuning (SpFT):} Only a selected fraction of the model parameters is fine-tuned on the downstream task to ensure high memory and compute efficiency. Diff pruning~\citep{diff_pruning} and LT-SFT~\citep{lt_sft} incorporate FFT phases, defeating the very purpose of SpFT. PaFi~\citep{pafi} selects the parameters with the smallest pre-trained magnitudes while FISH Mask~\citep{fish_mask}, SIFT~\citep{sift}, SAFT~\citep{saft}, SMT~\citep{smt}, and SpIEL~\citep{spiel} select the parameters with the highest gradient magnitudes on a downstream task for fine-tuning. While FISH Mask, SIFT, SAFT, and SMT generate a \emph{static} (fixed before fine-tuning) mask of the parameters to be fine-tuned via a single pass on the downstream task dataset, SpIEL iteratively generates \emph{dynamic} masks by alternating between the updation, deletion, and growth of the candidate parameter set. To the best of our knowledge, GaLLoP is the only SpFT technique which incorporates dual parameter selection criteria: high task gradient magnitudes and low pre-trained magnitudes, which consistently improves generalizability and ensures stability.

\subsection{Post-Training Model Editing}
Most fine-tuning techniques (except SAFT~\citep{saft}) focus solely on improving ID accuracy on a target downstream dataset at the risk of OOD performance degradation~\citep{ID_OOD_study,distr_shift_imagenet}. Consequently, weight-space modifications to a single fine-tuned model can serve as a robust and efficient alternative to complex modifications in fine-tuning. \textbf{WiSE-FT} (\textbf{W}e\textbf{i}ght-\textbf{S}pace \textbf{E}nsembles for \textbf{F}ine-\textbf{T}uning)~\citep{wise_ft} performs a linear ensembling of zero-shot and end-to-end fine-tuned model weights to combine robustness with target-specific accuracy. \textbf{LiNeS} (\textbf{L}ayer-\textbf{i}ncreasing \textbf{Ne}twork \textbf{S}caling) performs layer-wise parameter scaling by maintaining shallower layers close to their pre-trained values for preserving generality while allowing deeper layers to retain task-specific characteristics~\citep{lines}. However, the dependence of editing techniques on the fine-tuned model limits their own performance improvements (if any).

\section{Motivation}
\label{sec:motivation}

Fine-tuning a model with our SpFT algorithm must not only minimize the model's loss on the downstream task (for enhancing ID generalizability) but at the same time, also preserve the pre-trained knowledge stored in the model parameters (for enhancing OOD generalizability). The first criterion is straightforward to realize: since parameters with the largest gradients generally lead to fast convergence to the optimum on the downstream task, they must be selected for fine-tuning so as to minimize the loss on the downstream task~\citep{saft}. However, the second criterion does not directly follow from the first since parameters with high gradients on the downstream task can still be highly relevant for the pre-trained model and fine-tuning them can potentially lead to catastrophic forgetting~\citep{repr_collapse}. While the model pruning literature regards parameters with the smallest pre-trained magnitudes as the least important~\citep{prune1}, fine-tuning them does not necessarily lead to the least impact on the pre-trained loss~\citep{snip}. 

Therefore, we analyze the differences in ID and OOD performance upon fine-tuning only the top-$\rho\%$ of the model parameters based on whether they have a) the smallest or b) the largest pre-trained magnitudes. We consider eight reasoning datasets~\citep{llm-adapters} and fine-tune Gemma 2B~\citep{gemma} models, considering four different density levels. Models are fine-tuned on the training set of each of these datasets (\emph{individually}) and their performance is evaluated on the corresponding dataset's test set (unseen during fine-tuning yet possesses the training set's distribution), which measures their ID accuracy, and the test sets of the remaining seven datasets (unseen during fine-tuning and possess distributions which differ from the training set's distribution), which measures their OOD accuracy (averaged over the seven datasets). The averaged (over all possible ID and OOD task combinations) ID and OOD accuracies of these models are shown in~\autoref{fig_ablations_motivation} (see~\autoref{Appendix_abl} for further details). 

\begin{figure*}[htbp]
\centering
\includegraphics[scale=0.40]{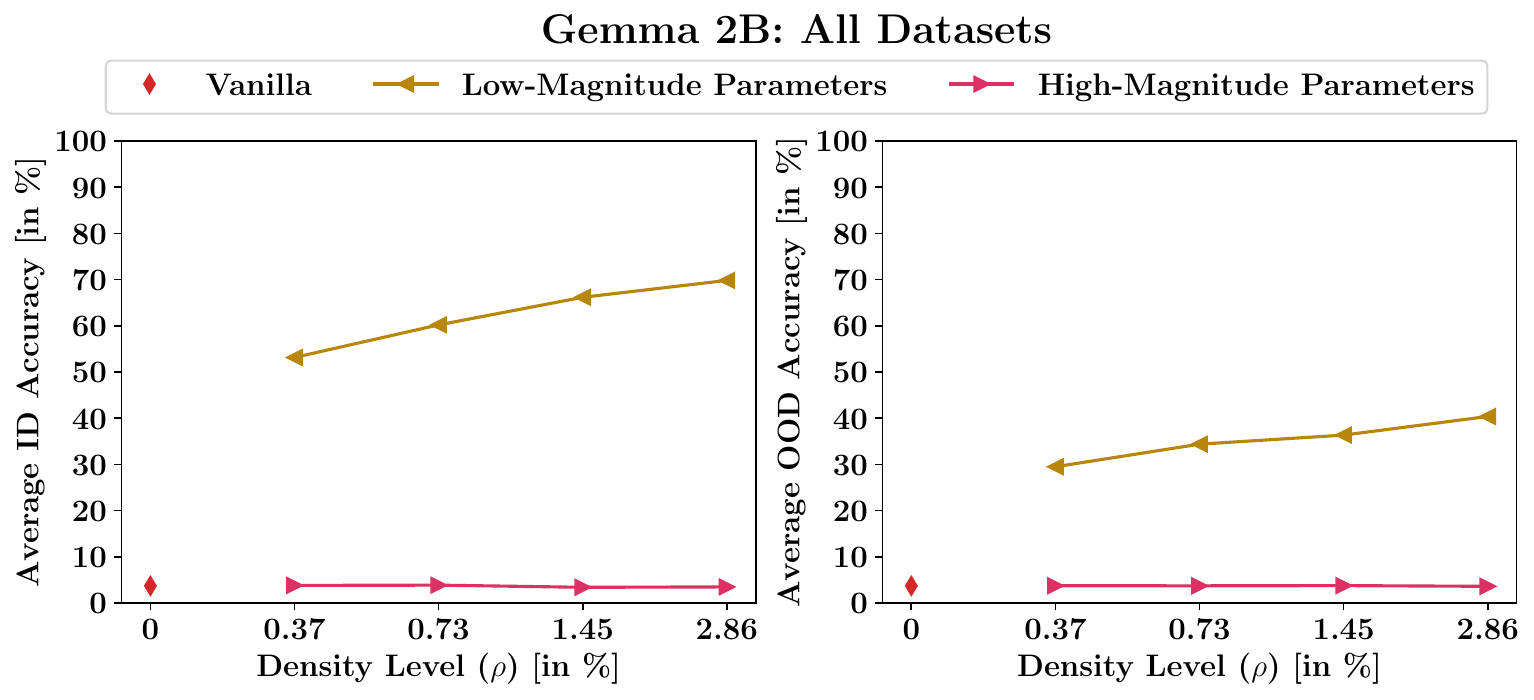}
\caption{Fine-tuning the top-$\rho\%$ of model parameters with the largest pre-trained magnitudes leads to no performance improvements over the non fine-tuned, vanilla model while fine-tuning the top-$\rho\%$ of model parameters with the smallest pre-trained magnitudes enhances the pre-trained knowledge, leading to significantly increased ID (left) and OOD (right) generalizability.}
\label{fig_ablations_motivation}
\end{figure*}

While fine-tuning the parameters with the largest pre-trained magnitudes leads to no performance improvements over the non fine-tuned, vanilla model (only 3-4\% ID and OOD accuracies), fine-tuning the parameters with the smallest pre-trained magnitudes leads to increasingly high ID ($\approx$ 50 - 70\% accuracy) and OOD ($\approx$ 30 - 40\% accuracy) generalizability. These findings are also in agreement with recent studies~\citep{attn_small_wts,blockllm} and support our hypothesis that fine-tuning low-magnitude parameters enhances pre-trained knowledge.

\section{Methodology}
\label{sec:method}

Motivated by the results discussed in~\autoref{sec:motivation}, we now present our SpFT algorithm, \textbf{GaLLoP}: \textbf{G}radient-based Sp\textbf{a}rse \textbf{L}earning on \textbf{Lo}w-Magnitude \textbf{P}arameters. Given a density level $\rho$, GaLLoP fine-tunes the top-$\rho\%$ of the model parameters with the largest gradient magnitudes (to enhance ID generalizability) and the smallest pre-trained magnitudes (to enhance OOD generalizability). We now give a formal description of how GaLLoP works.

\begin{figure*}[htbp]
    \begin{center}
        \subfigure[Phase 1]{
            \label{phase1}
            \includegraphics[scale = 0.26]{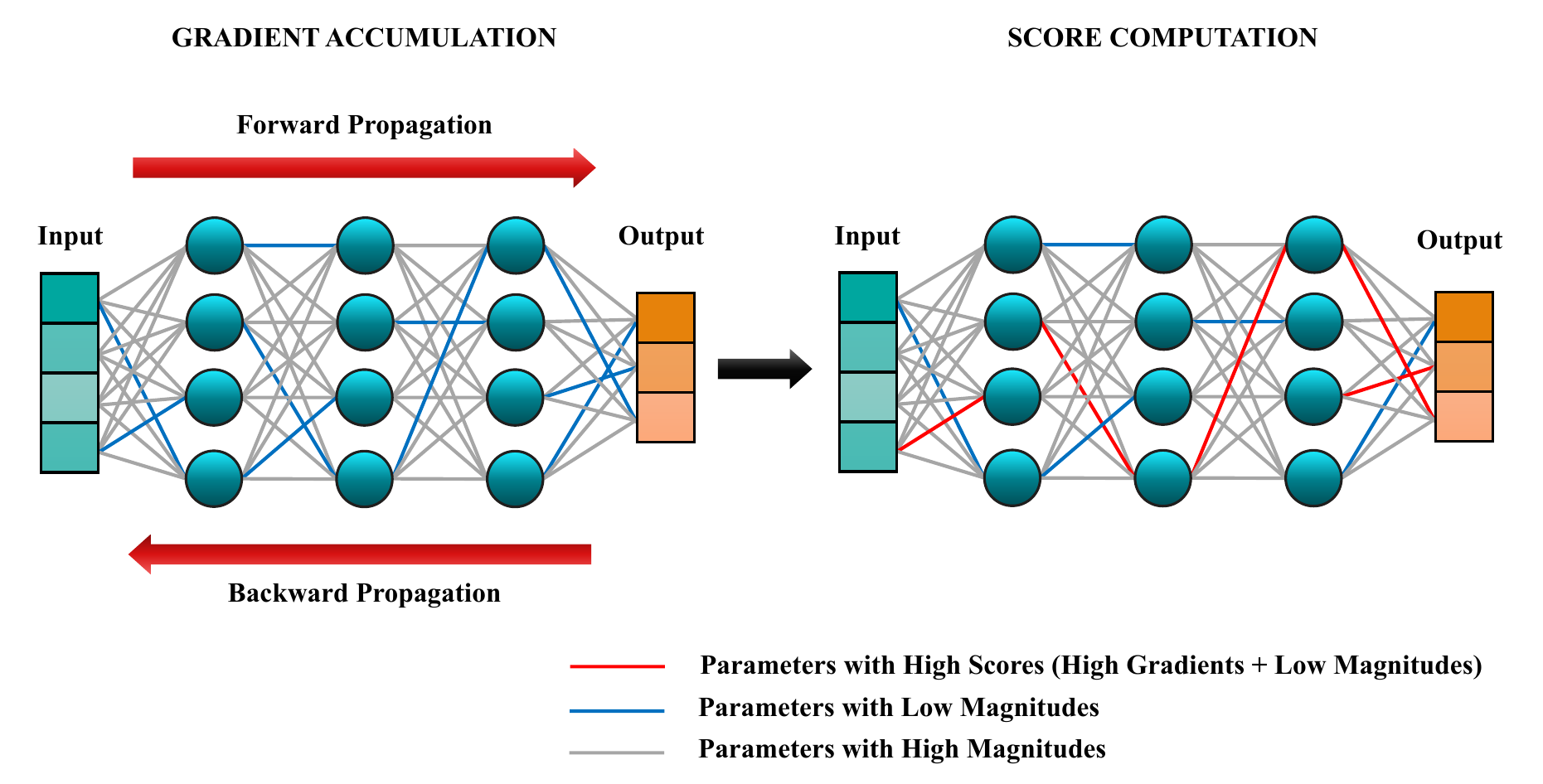}
        }
        \subfigure[Phase 2]{
            \label{phase2}
            \includegraphics[scale = 0.26]{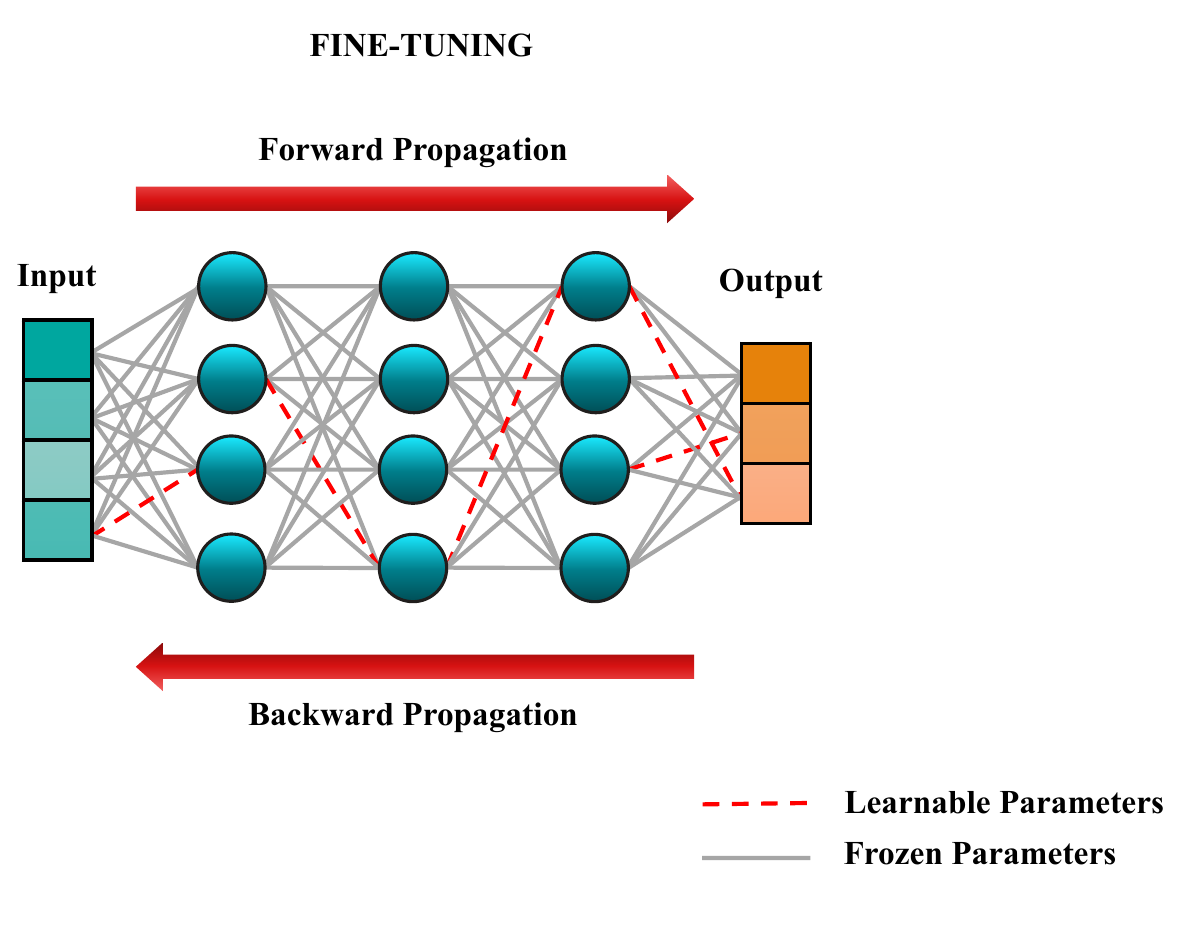}
        }
    \end{center}
    \caption{A visual overview of GaLLoP: (a) Phase 1 involves the selection of learnable parameters and (b) Phase 2 involves their fine-tuning. Note that the example of a fully connected neural network has only been considered for ensuring simplicity in visualization; GaLLoP is model-agnostic and can be applied to other models such as transformers and CNNs as well.}
    \label{fig_gallop}
\end{figure*}

GaLLoP operates in two phases (see~\autoref{fig_gallop}). Consider a model with a parameter vector $\boldsymbol{\theta} \in \mathbb{R}^D$ which needs to be fine-tuned on a dataset $\mathcal{D} = {\{(\bm{x}_n, \bm{y}_n)\}}_{n = 1}^{N}$, where, $\bm{x}_n \in \mathbb{R}^{P}$ and $\bm{y}_n \in \mathbb{R}^{Q}$. In the first phase, given a density level $\rho$, GaLLoP selects the learnable parameters using a dataset sample proportion $d_s$ (corresponding dataset sample size = $d_sN$). To do so, it proceeds by computing the gradient vector $\bm{g}$ of the net fine-tuning loss $\mathcal{L}(\bm{x}_n, \bm{y}_n; \boldsymbol{\theta})$ which is given by:
\begin{align}
     \bm{g} = \dfrac{1}{d_s N}\sum_{n = 1}^{d_sN}\nabla_{\boldsymbol{\theta}}\mathcal{L}(\bm{x}_n, \bm{y}_n; \boldsymbol{\theta}).
     \label{grad_eqn}
\end{align}

This allows us to enforce our first parameter selection criterion (high-magnitude gradients). In order to simultaneously enforce our second parameter selection criterion (low- (pre-trained) magnitude parameters), we compute a score vector $\bm{s}$ such that:
\begin{align}
    \bm{s} =  \left(\dfrac{\text{abs}(\bm{g})}{\text{abs}(\boldsymbol{\theta}) + \epsilon}\right),
    \label{score_eqn}
\end{align}
where, \text{abs}(.) computes the element-wise absolute value of a vector and a small value $\epsilon \approx 10^{-8}$ has been added to the denominator so as to prevent numerical overflows. 

Subsequently, in order to select only the top-$\rho\%$ of all model parameters $\boldsymbol{\theta}$ for fine-tuning based on their corresponding scores $\bm{s}$, we compute a binary mask vector $\bm{m}$ such that, for a given parameter $\boldsymbol{\theta}_{i}$ with a score $\bm{s}_{i}$, the mask value $\bm{m}_{i}$ is given by:
\begin{align}
    \bm{m}_{i} = \begin{cases}1 &\text{if } \bm{s}_{i} \geq s_t,\\0 &\text{otherwise.}\end{cases}
    \label{mask_eqn}
\end{align}
where, the score threshold $s_t$ is computed as follows:
\begin{align}
    s_t = \mathtt{sorted}_d(\bm{s}_0, \bm{s}_1, \dots ,\bm{s}_{\text{end}})[k]
    \label{threshold_eqn}
\end{align}
where, $k = \lfloor \rho D \rfloor$ and $\mathtt{sorted}_d(.)$ is the function used for sorting an array in the descending order.

Finally, in the second phase, GaLLoP only updates the selected, \emph{unmasked}, parameters using mini-batch gradient descent while rendering the values of the remaining (unselected), \emph{masked}, parameters unchanged by masking out their gradients during the update using $\bm{m}$. We give a practical implementation of GaLLoP in~\autoref{Appendix_alg}.

\section{Experimental Settings}

\subsection{Datasets}
To examine the effectiveness of GaLLoP, we perform fine-tuning experiments on eight commonsense reasoning datasets with predefined training and test sets~\citep{llm-adapters}: ARC-c~\citep{arc}, ARC-e~\citep{arc}, BoolQ~\citep{boolq}, HellaSwag~\citep{hellaswag}, OBQA~\citep{obqa}, PIQA~\citep{piqa}, SIQA~\citep{siqa}, and WinoGrande~\citep{winogrande}. Further details are provided in~\autoref{AppendixA}.

For the purpose of experimentation, we consider all possible ID and OOD combinations involving these datasets by following a round-robin approach. Thus, for each experimental run, a model is fine-tuned on the training set of one of these eight datasets and then evaluated on its test set as well as the test sets of all the remaining seven datasets. Accordingly, eight different experimental runs are performed as part of an experiment for a given density level $\rho$ that is enforced during the fine-tuning of a model with a given algorithm. Performance evaluations on the test set of the dataset used for fine-tuning serve as a measure of ID generalizability while those on the test sets of the remaining datasets (not used for fine-tuning) serve as a measure of the OOD generalizability.

\subsection{Model Architectures}
We perform our experiments with Gemma 2B~\citep{gemma} (\textit{relatively small-sized}) and LLaMA3 8B~\citep{llama} (\textit{relatively large-sized}) as base models.

\subsection{Baselines} 
For a rigorous evaluation of performance, we compare GaLLoP with several SOTA fine-tuning algorithms for LLMs. We employ Full Fine-Tuning (FFT) as our fine-tuning baseline and the Zero-Shot (Vanilla) model performance as an overall, non fine-tuning (pre-trained) baseline. We also include LoRA~\citep{lora} and DoRA~\citep{dora} from the RFT category, SAFT~\citep{saft} and SpIEL~\citep{spiel} from the SpFT category, and WiSE-FT~\citep{wise_ft} and LiNeS~\citep{lines} from the post-training model editing category.

\subsection{Evaluation Metrics}
Since we consider all possible ID and OOD task combinations to ensure a robust evaluation, a fine-tuning experiment with an algorithm $\mathcal{A}$ and a density level $\rho$ on $N^D$ datasets involves $N^D$ experimental runs, such that, in each run $r$, fine-tuning is performed on a dataset $\mathcal{D}^{f_r}$. For a given experimental run $r$, we evaluate the ID and OOD performance of all the fine-tuned models using ID accuracy and OOD accuracy which are defined as follows:
\begin{equation}
\text{Accuracy}\strut_{r}^{\text{ID}} = \text{accuracy($\mathcal{D}_{test}^{f_r}$)}\quad;\quad\text{Accuracy}\strut_{r}^{\text{OOD}} = \dfrac{1}{N^D - 1} \sum\limits_{\substack{n = 1\\n \neq f_r}}^{N^D}\text{accuracy($\mathcal{D}_{test}^n$)},
\label{eqn:per_run_acc}
\end{equation}
where, $\text{accuracy}(\mathcal{D}_{test})$ computes the percentage of correct responses generated by a fine-tuned model on the test set $\mathcal{D}_{test}$ of a dataset $\mathcal{D}$.

In addition to the standard accuracy metrics, we introduce two new metrics, the Forget Ratio and the Collapse Rate, which aim to quantify the extent of catastrophic forgetting and memorization incurred upon fine-tuning (respectively). For a given experimental run $r$, while fine-tuning leads to performance improvements on the (ID) downstream task, it may result in the degradation of zero-shot (vanilla) performance on the remaining OOD tasks due to the loss of pre-trained knowledge, i.e., catastrophic forgetting. Accordingly, we define the forget ratio as a measure of this relative drop in OOD performance such that:
\begin{equation}
    \text{Forget Ratio}_r = \max\left(0, \dfrac{\text{Accuracy}\strut_{\text{Vanilla},\,r}^{\text{OOD}} - \text{Accuracy}\strut_{r}^{\text{OOD}}}{\text{Accuracy}\strut_{\text{Vanilla},\,r}^{\text{OOD}}}\right),
    \label{per_run_cfr}
\end{equation}
where, $\text{Accuracy}\strut_{\text{Vanilla},\,r}^{\text{OOD}}$ refers to the accuracy of the zero-shot (vanilla) model on the OOD test sets for the $r\textsuperscript{th}$ experimental run. From the above equation, it follows that gains in OOD performance, relative to the performance of the vanilla model, lead to a 0\% forget ratio as desired.

The collapse rate for a given experimental run $r$ measures the extent to which fine-tuning results in severe memorization of patterns present in $\mathcal{D}^{f_r}$. It is computed by determining the total number of OOD datasets, on the test sets of which, the accuracy drops to $\approx 0\%$ (indicative of a complete collapse) and is hence, defined as follows:
\begin{equation}
    \text{Collapse Rate}_r = \sum\limits_{\substack{n = 1 \\ n \neq {f_r}}}^{N^D}\mathbbm{1}[\lfloor \text{accuracy}({\mathcal{D}_{test}^n}) \rfloor = 0\%],
    \label{per_run_mem_rate}
\end{equation}
where, $\mathbbm{1}[.]$ denotes the indicator function which evaluates to 1 when its argument is True and 0 otherwise.

For each fine-tuning algorithm $\mathcal{A}$ and density level $\rho$, the obtained performance and the extent of catastrophic forgetting and memorization incurred across the entire experiment consisting of $N^D$ experimental runs is given by the average of the aforementioned four metrics across all the runs.

\subsection{Implementation Details}
We perform all our experiments using Torchtune~\citep{torchtune} and Huggingface's PEFT library~\citep{hf_peft}. We use NVIDIA RTX A6000 and/or NVIDIA RTX 6000 Ada GPUs with 48 GB of internal memory and perform mixed-precision (using the BF16 datatype), distributed fine-tuning of all our models (except SpIEL, which can be only run on a single GPU~\citep{spiel}) using Fully Sharded Data Parallel (FSDP)~\citep{fsdp}, activation checkpointing, and gradient accumulation. To ensure a fair comparison, we apply each fine-tuning algorithm across all the transformer layers and maintain the same density level ($\rho$) for a given experiment. We explore four density levels:  \{0.24\%, 0.47\%, 0.93\%, 1.85\%\} for LLaMA3 8B, and \{0.37\%, 0.73\%, 1.45\%, 2.86\%\} for Gemma 2B which correspond to ranks \{8, 16, 32, 64\} used for reparametrized fine-tuning. Further details on the hyperparameters are given in~\autoref{AppendixB}.

\section{Results and Discussion}

\subsection{ID and OOD Accuracy}
\label{sec:ID_OOD_acc}

\begin{figure*}[htbp]
    \begin{center}
        \subfigure[]{
            \label{fig_CR_full_avg_llama}
            \includegraphics[scale = 0.25]{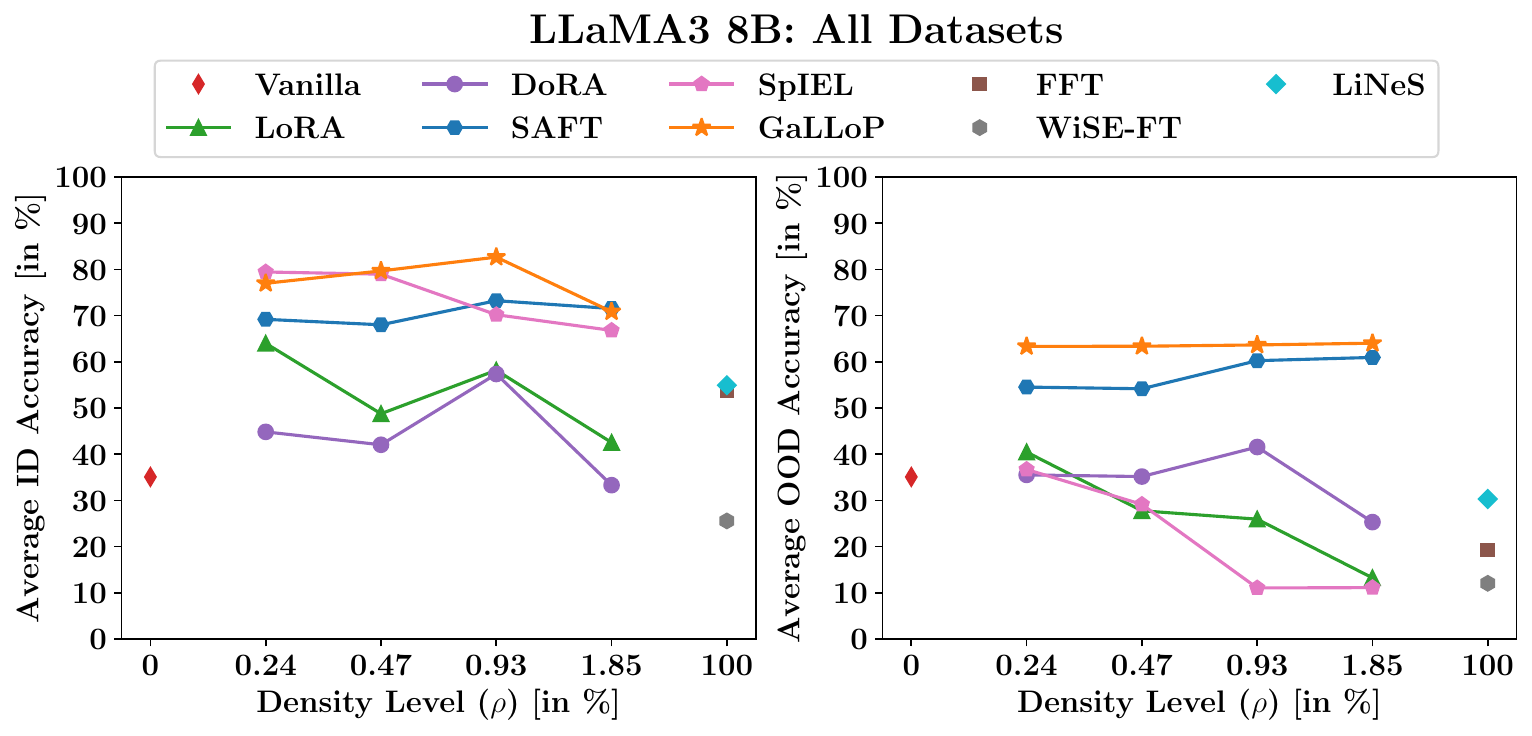}
        }
        \subfigure[]{
            \label{fig_CR_full_avg_gemma}
            \includegraphics[scale = 0.25]{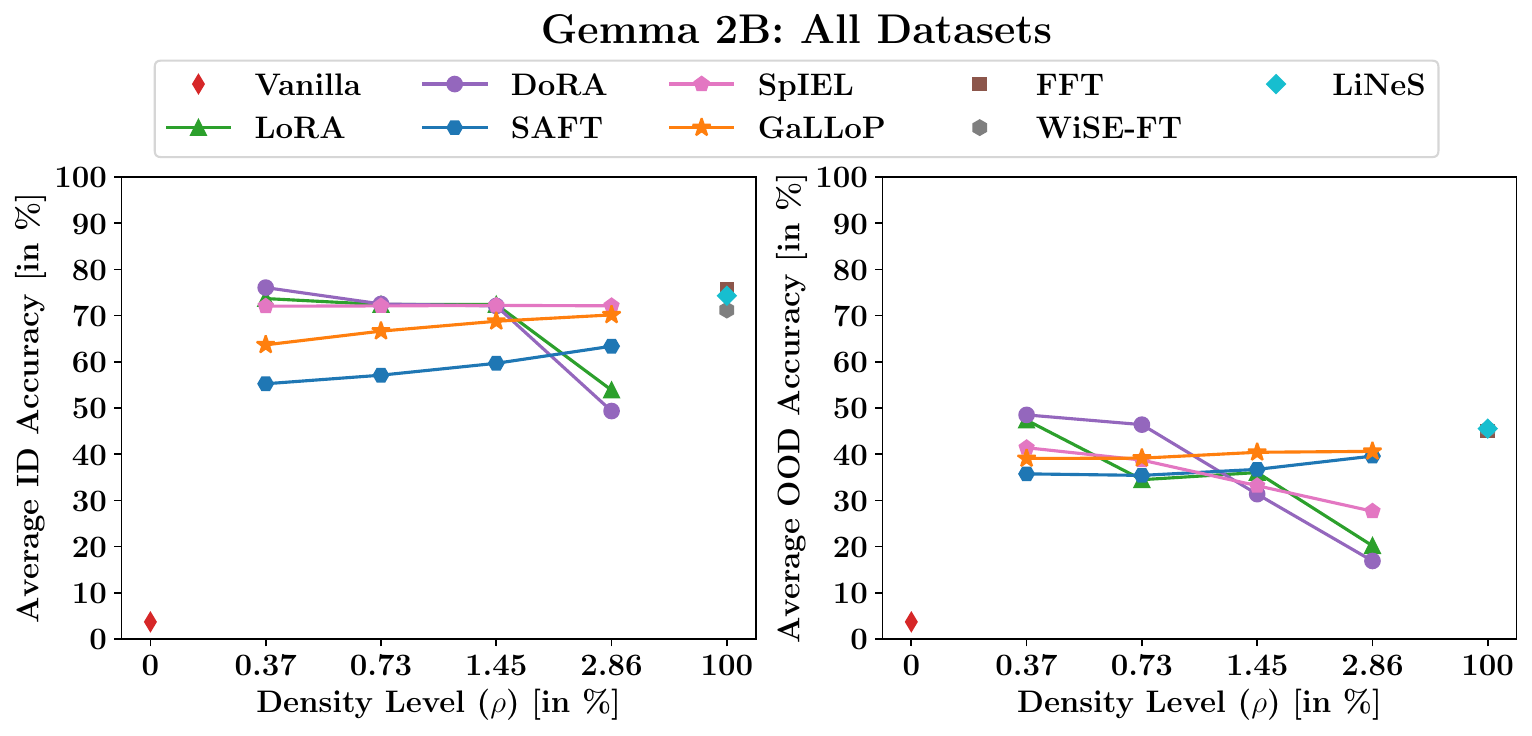}
        }
    \end{center}
    \caption{a) LLaMA3 8B models fine-tuned with GaLLoP form a dominant Pareto front for both ID and OOD performance (on average) over all the models fine-tuned and/or edited with the competing algorithms across all density levels. b) Gemma 2B models fine-tuned with GaLLoP attain consistently high and balanced ID and OOD performance (on average) across all density levels.}
    \label{fig_CR_full_avg}
\end{figure*}

\autoref{fig_CR_full_avg} shows the averaged ID and OOD accuracies of models fine-tuned with GaLLoP against those of the vanilla models and models fine-tuned and/or edited with the competing algorithms (per-run ID and OOD accuracies are discussed in detail in~\autoref{app:id_ood} of~\autoref{AppendixD}).

LLaMA3 8B and Gemma 2B models fine-tuned with GaLLoP attain high ID and OOD accuracies (on average) while maintaining relatively balanced ID and OOD performance across all the density levels. In fact, LLaMA3 8B models fine-tuned with GaLLoP form a dominant Pareto front for both ID and OOD performance over all the other fine-tuned and/or edited models. Models fine-tuned with GaLLoP consistently surpass those fine-tuned with SAFT with a high average margin of roughly 10\% (for both ID and OOD performance), which only narrows down for the highest density level due to \emph{high gradient dilution} and \emph{low-magnitude dilution}. With the increase in the density level, parameters with a relatively higher pre-trained magnitude and lower gradient magnitude also fall into the `high gradient and low-magnitude' category, resulting in an overlap between the set of parameters selected by SAFT and GaLLoP. Nevertheless, as is observed, GaLLoP is more efficient in its selection of parameters and would still form a performance upper bound over SAFT even as $\rho \rightarrow 100\%$ (asymptotic limit), particularly because it would still select a larger/at least as large a concentration of low-magnitude high gradient parameters than/as the latter.

While models fine-tuned with SpIEL show high ID accuracies, they perform poorly on OOD tasks with a huge average gap ($\gtrapprox 30\%$) in their own ID and OOD accuracies (on average), which widens with the increase in the density level. Contrary to the other SpFT methods, SpIEL iteratively selects parameters, potentially updating a much larger number of parameters overall. Hence, it leads to increasingly high overfitting on the in-domain distribution with the increase in the density level. Models fine-tuned with RFT techniques also exhibit high levels of overfitting since all the newly introduced trainable parameters are fine-tuned on the training set. Further, while the ID and OOD accuracies obtained via RFT are greater than those obtained via GaLLoP for Gemma 2B for the first two density levels, these gains are only confined to certain datasets and not to others due to catastrophic forgetting and memorization, as we show subsequently in~\autoref{sec:cfr} and~\autoref{sec:mem}. 

Moreover, the performance of FFTed models is highly dependent on their zero-shot performance. While FFT works quite well for the small Gemma 2B model by yielding high performance levels comparable to those yielded by GaLLoP, it completely fails for the 4X larger LLaMA3 8B model. Overtraining, i.e., considerably higher pre-training (6T pre-training tokens for Gemma 2B~\citep{gemma} versus 15T pre-training tokens for LLaMA3 8B~\citep{llama}), makes models much more sensitive to parameter updates, leading to severe catastrophic forgetting and losses in OOD as well as ID generalizability~\citep{FFT_distortion, overtraining}. In fact, this is also the reason why the performance gap between GaLLoP and the competing algorithms decreases for Gemma 2B as compared to LLaMA3 8B. However, GaLLoP exhibits superior performance in both pre-training regimes which implies that it is robust to overtraining, a pre-training regime wherein the competing algorithms simply fail. Finally, the fine-tuned performance dependence and meagre performance improvements (\emph{if any}) yielded by WiSE-FT and LiNeS defeats the very purpose of these post-training model editing techniques which were developed with an aim to provide higher robustness to generalization while avoiding the complexities of fine-tuning.

\subsection{Catastrophic Forgetting}
\label{sec:cfr}

\begin{figure*}[htbp]
    \begin{center}
        \subfigure[]{
            \label{fig_CR_cfr_llama}
            \includegraphics[scale = 0.25]{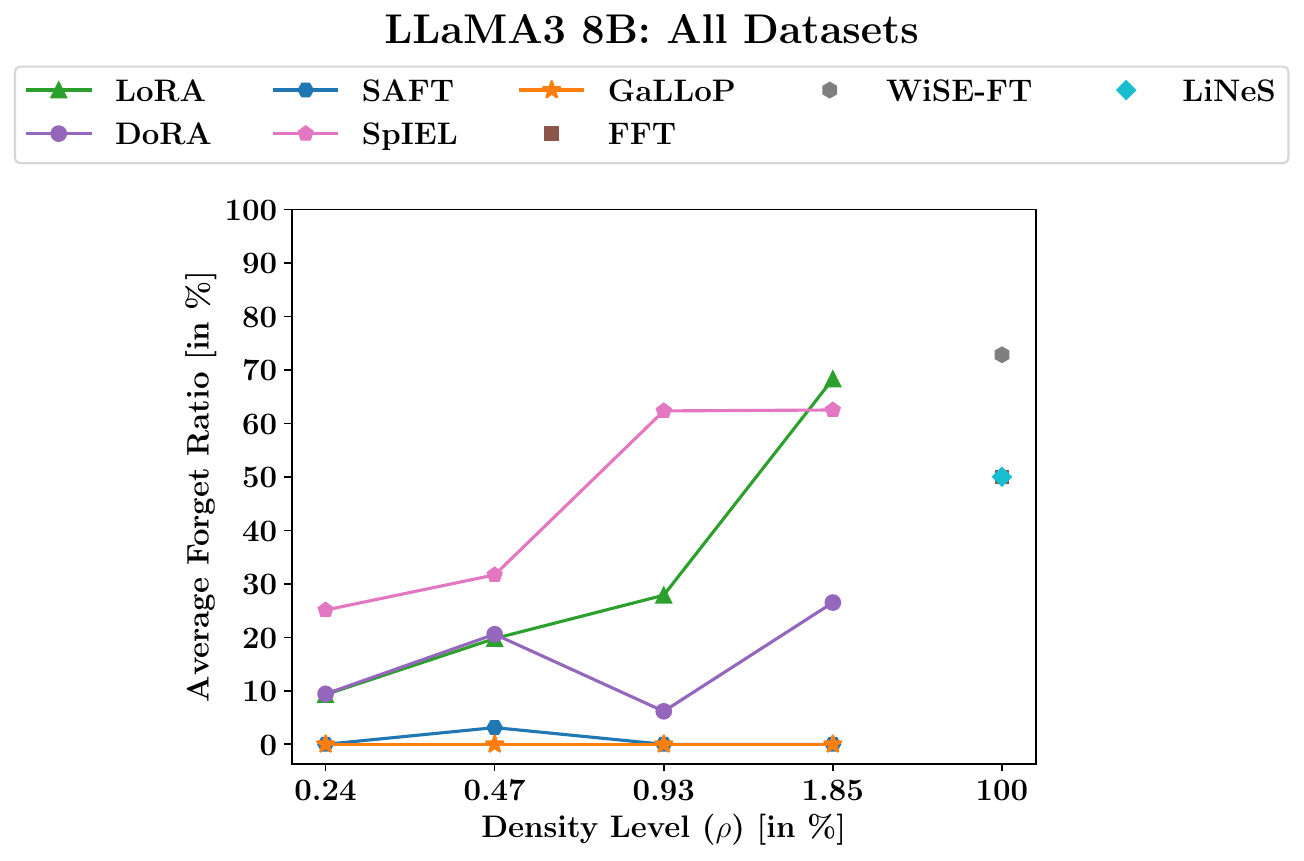}
        }
        \subfigure[]{
            \label{fig_CR_cfr_gemma}
            \includegraphics[scale = 0.25]{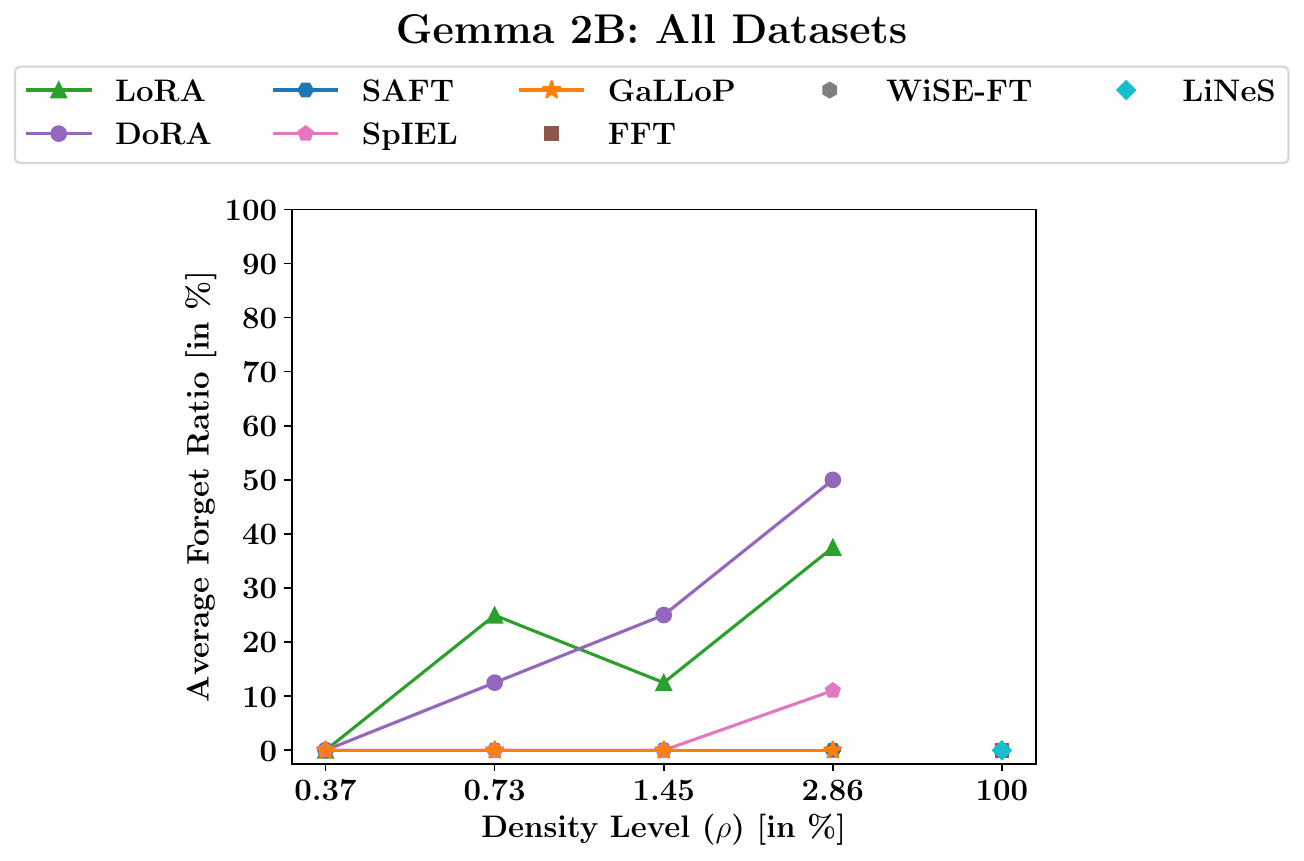}
        }
    \end{center}
    \caption{Models fine-tuned with GaLLoP show 0\% forget ratios across all density levels.}
    \label{fig_CR_cfr}
\end{figure*}

\autoref{fig_CR_cfr} shows the averaged forget ratios of models fine-tuned with GaLLoP against those fine-tuned and/or edited with the competing algorithms (per-run forget ratios are discussed in~\autoref{app:cfr} of~\autoref{AppendixD}). In an ideal scenario, the forget ratio must be zero since fine-tuning a model must not lead to any loss in the pre-trained knowledge stored in the model.

LLaMA3 8B and Gemma 2B models fine-tuned with GaLLoP do not undergo catastrophic forgetting since they consistently show 0\% forget ratios. In contrast, all the competing algorithms show high forget ratios which may even take on  increasingly high values with the increase in the density level due to overfitting. In general, for all competing algorithms, fine-tuned Gemma 2B models show significantly decreased forget ratios as compared to their LLaMA3 8B counterparts. While this does happen because of overtraining, there are, in actuality, \emph{two sides} of the same coin. While lesser pre-training of Gemma 2B (as compared to LLaMA3 8B) lowers the sensitivity of updates to its parameters and allows fine-tuning it to be more stable and performant (which decreases the risk of catastrophic forgetting)~\citep{overtraining}, it also leads to the decreased and (here) near-zero, vanilla (zero-shot) performance of the former as compared to the latter, in the first place itself (see~\autoref{vanilla_case} in~\autoref{AppendixD}). Since forget ratios are defined relative to vanilla performance (\autoref{per_run_cfr}), even though the OOD performance of certain fine-tuned Gemma 2B models is quite low, it never goes below that of their vanilla counterpart, which leads to all of them attaining a 0\% forget ratio. Nevertheless, these fine-tuning algorithms may lead to a more severe phenomenon which impairs generalizability: \emph{memorization}. This forms the subject of our subsequent discussion.

\subsection{Memorization}
\label{sec:mem}

\begin{figure*}[htbp]
    \begin{center}
        \subfigure[]{
            \label{fig_CR_fr_llama}
            \includegraphics[scale = 0.25]{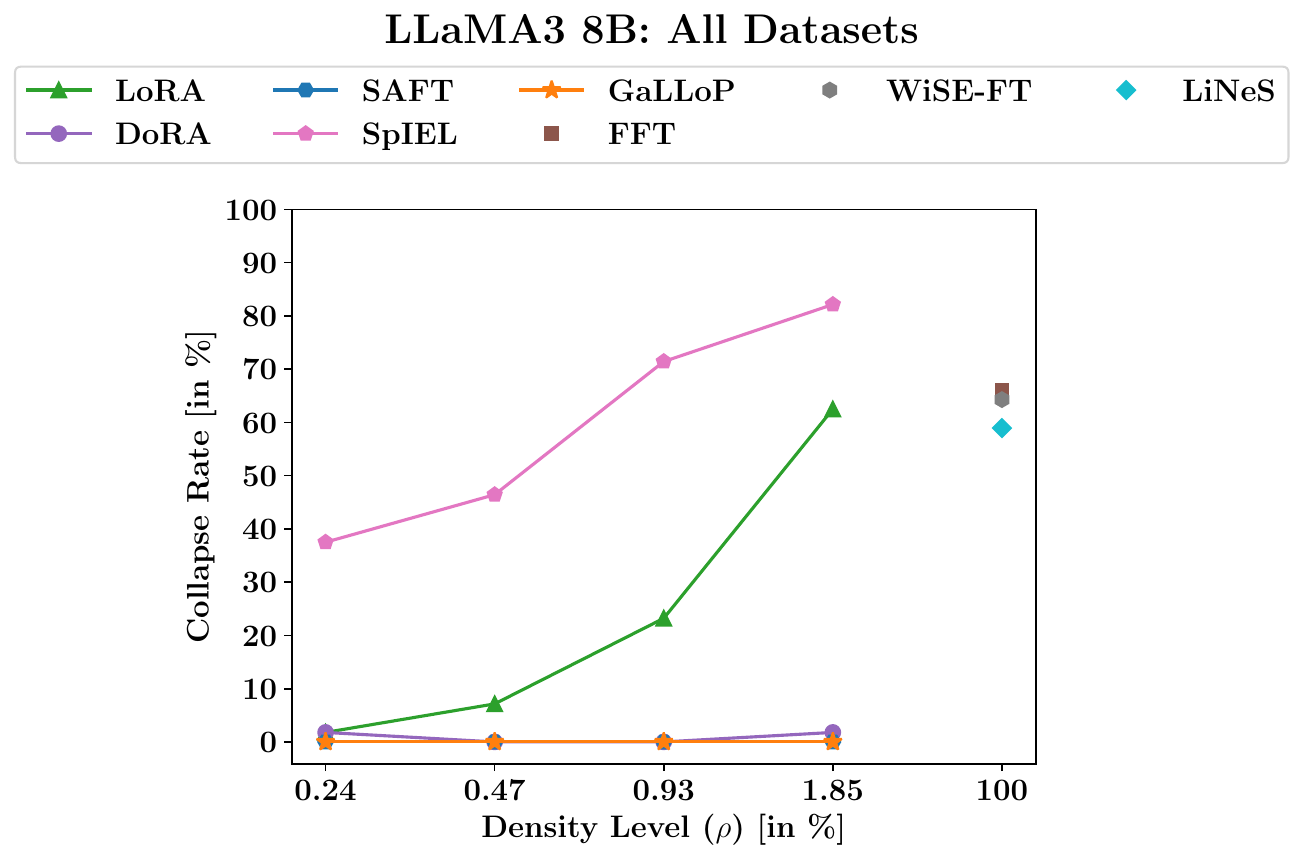}
        }
        \subfigure[]{
            \label{fig_CR_fr_gemma}
            \includegraphics[scale = 0.25]{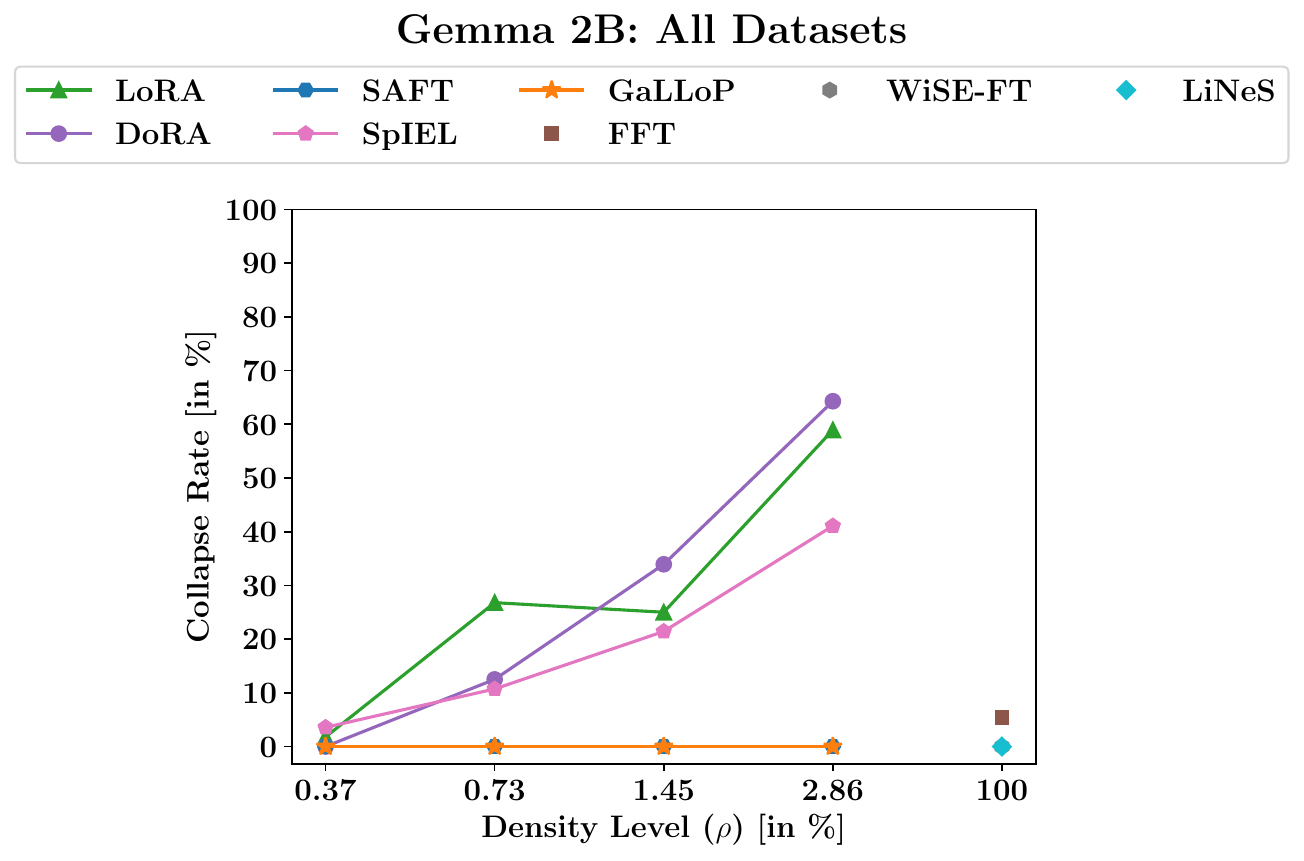}
        }
    \end{center}
    \caption{Models fine-tuned with GaLLoP show 0\% collapse rates across all density levels.}
    \label{fig_CR_fr}
\end{figure*}

\autoref{fig_CR_fr} shows the total collapse rates of models fine-tuned with GaLLoP against those fine-tuned and/or edited with the competing algorithms. In an ideal scenario, the collapse rate must be zero since fine-tuning a model must not lead to the memorization of any kind of patterns present in the dataset used for fine-tuning it.

LLaMA3 8B and Gemma 2B models fine-tuned with GaLLoP and SAFT exhibit 0\% collapse rates across all density levels. Combined with the dominant ID as well as OOD performance and 0\% forget ratios exhibited by models fine-tuned with GaLLoP over those fine-tuned with SAFT (see~\autoref{fig_CR_full_avg} and~\autoref{fig_CR_cfr}), it is clear that fine-tuning models with GaLLoP allows for the attainment of high yet balanced generalizability.

For the other algorithms, fine-tuned LLaMA3 8B models show higher collapse rates than their Gemma 2B counterparts due to overtraining. Furthermore, while memorization inevitably occurs due to overfitting, it can manifest in different ways. Throughout our experiments, the most dominant form of memorization observed by us is that of the memorization of the response format of the dataset used for fine-tuning. With the increase in the density level, this kind of memorization leads to models fine-tuned with these algorithms increasingly failing on an OOD task which does not share the same response format as the ID task. Another, more severe form of memorization seems to be more pervasive since it affects the performance of fine-tuned LLaMA3 8B models across all the datasets (ID as well as OOD) irrespective of whether they share response formats or not: memorization of the most frequently occurring words/phrases in the fine-tuning dataset. This leads to the generation of repetitive sequences as answers and consequently, degrades their intelligibility. Since we find that repetition is only restricted to RFT and does not occur on performing SpFT, we attribute its occurrence to the fact that the former class of algorithms restrict fine-tuning to be performed on newly introduced parameters, tied to specific positions in the model, which leads to \emph{concerted overfitting} while the latter class of algorithms allow for much more flexibility in fine-tuning via a position-independent selection of parameters to be fine-tuned and lead to \emph{scattered (unstructured) learning}. Finally, LLaMA3 8B models fine-tuned with SpIEL and FFT (and hence, even those edited with WiSE-FT and LiNeS) occasionally undergo the worst form of memorization: generation of the EOS token. In fact, not only FFT but also SpIEL (as explained earlier in~\autoref{sec:ID_OOD_acc}) fine-tunes a much larger number of parameters than the other algorithms and hence, both of them are more prone to lead to instability in the learning dynamics in the overtrained regime which can lead to such a detrimental form of memorization. We discuss all these interesting cases in more detail with representative samples of LLM-generated responses in~\autoref{app:mem}.

\subsection{Stability}

\begin{figure*}[htbp]
\centering
\includegraphics[scale=0.40]{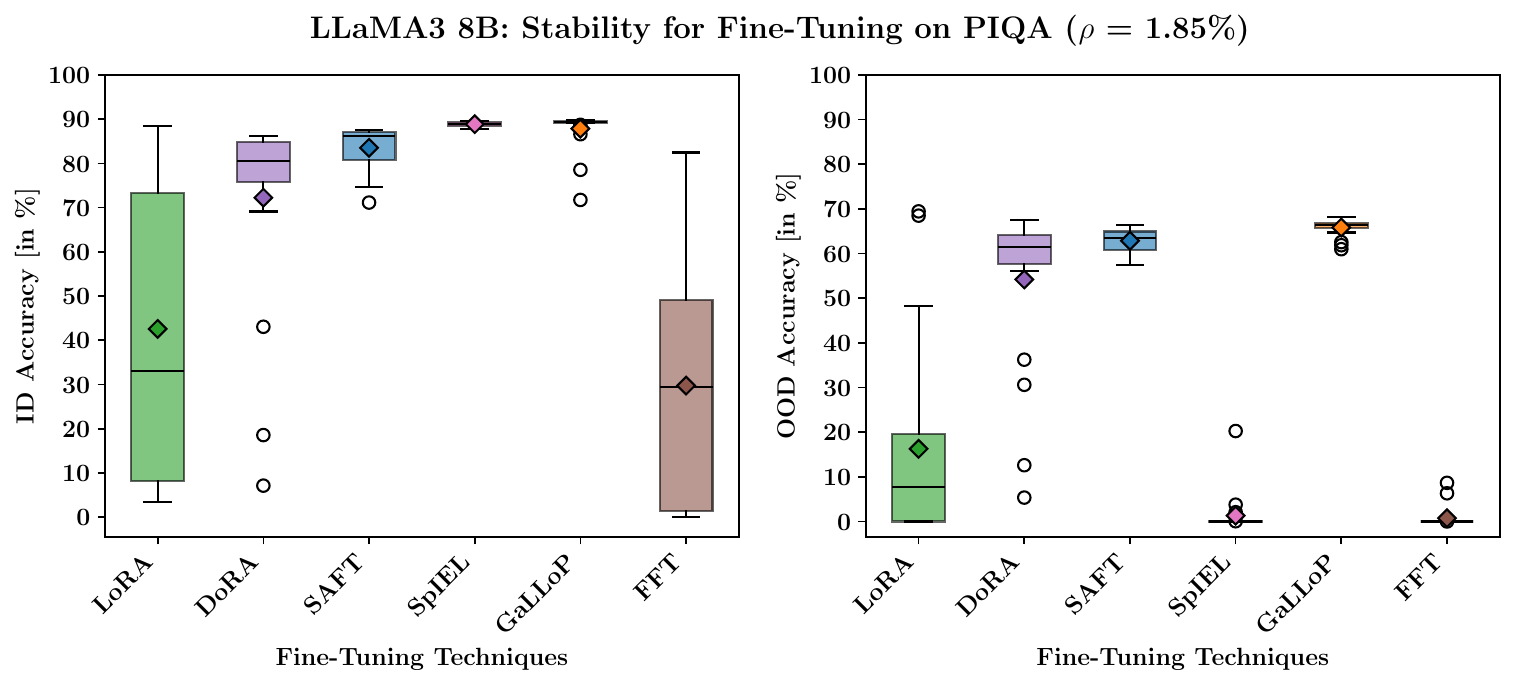}
\caption{LLaMA3 8B models fine-tuned with GaLLoP are the most stable and consistently attain the highest ID and OOD accuracies across 20 different random seeds upon fine-tuning on the PIQA dataset with the highest density level ($\rho = 1.85\%$).}
\label{fig_stability}
\end{figure*}

\autoref{fig_stability} shows the boxplots of ID and OOD performance for LLaMA3 8B models fine-tuned with GaLLoP against models fine-tuned with the competing algorithms across 20 different random seeds. We deliberately choose the PIQA dataset here because the FFTed LLaMA3 8B model attains the lowest ID and OOD performance on this dataset (see~\autoref{fig_CR_llama_piqa} in~\autoref{AppendixD}). Moreover, we only consider the highest $\rho$ ($= 1.85\%$) so as to rigorously analyze performance stability in the overtrained regime with the largest examined number of update-sensitive parameters. Note that we do not consider editing techniques here since their stability is inherently dependent upon the stability of the fine-tuned model (see~\autoref{sec:ID_OOD_acc}). More details are given in~\autoref{app:var_cfr_mem}. 

Models fine-tuned with GaLLoP consistently attain the highest median ID and OOD accuracies with the least interquartile ranges. Furthermore, the difference between the farthest outlier and median performance is amongst the lowest for GaLLoP and never falls below the lowest performance level of SAFT. Hence, fine-tuning with GaLLoP leads to the highest performance stability. In contrast, models fine-tuned with competing algorithms show high performance instability and lower median accuracies, with instability being the highest on performing RFT or FFT~\citep{fft_instability1, fft_instability2}. Directly leveraging low intrinsic dimensionality by infusing sparsity (as in SpFT) effectively regularizes fine-tuning as compared to reparametrization (as in RFT) which does not constrain updates to the newly introduced low-rank matrices and hence, leads to instability.

\section{Conclusion}
In this work, we have thus developed a novel SpFT technique named \textbf{GaLLoP}: \textbf{G}radient-based Sp\textbf{a}rse \textbf{L}earning on \textbf{Lo}w-Magnitude \textbf{P}arameters which enhances the ID as well as the OOD generalizability of models, prevents catastrophic forgetting and memorization, ensures robustness to overtraining, and stabilizes performance. Nevertheless, as is the case for other SpFT techniques, GaLLoP leads to unstructured sparsity. Accelerating unstructured fine-tuning poses a challenge for current hardware that is optimized for performing dense and/or structured computations~\citep{hardware_lottery, saft}. An interesting direction for future work could therefore be to perform densification of this unstructured sparsity by following an aggregation scheme~\citep{smt} so as to make it more structured (only) for fine-tuning and further increase the memory and compute efficiency of GaLLoP. 

\bibliography{iclr2026_conference}
\bibliographystyle{iclr2026_conference}

\clearpage

\appendix
\addcontentsline{toc}{section}{Appendix} 
\part{Appendix} 
\parttoc 

\clearpage

\section{Practical Implementation of GaLLoP}
\label{Appendix_alg}

Following the method outlined in~\autoref{sec:method}, in practice, since we fine-tune LLMs with billions of parameters, sorting the corresponding set of billions of scores in order to compute the score threshold $s_t$ (see~\autoref{threshold_eqn}) is computationally inefficient. Therefore, we perform layer-wise uniform random sampling of the scores and utilize only a small proportion $s$ of the total number of scores in order to compute $s_t$. 

Accordingly, we now use a more specific, layer-wise notation for the model to be fine-tuned.

Consider a deep neural network model with $L$ layers in total (1 input layer of width $p$, $L-2$ hidden layers of width $k$, and 1 output layer of width $q$), represented by $\bm{y} = F_{\{\boldsymbol{\Theta^l}\}_{l = 1}^L}(\bm{x})$, such that the parameter matrices $\boldsymbol{\Theta}^l$ possess the following set of dimensions:
\begin{align*}
    \boldsymbol{\Theta}^1 &\in \mathbb{R}^{p \times k}\\
    \boldsymbol{\Theta}^l &\in \mathbb{R}^{k \times k}\quad\forall\quad l \in \{2, \dots, L-1\}\\
    \boldsymbol{\Theta}^L &\in \mathbb{R}^{k \times q}
\end{align*}
This model is to be fine-tuned on a dataset $\mathcal{D} = {\{(\bm{x}_n, \bm{y}_n)\}}_{n = 1}^{N}$, where, $\bm{x}_n \in \mathbb{R}^{P}$ and $\bm{y}_n \in \mathbb{R}^{Q}$. 

In line with this notation,~\autoref{grad_eqn},~\autoref{score_eqn}, and~\autoref{mask_eqn} can be re-written as follows:
\begin{align}
     \bm{G}^l &= \dfrac{1}{d_s N}\sum_{n = 1}^{d_sN}\nabla_{\boldsymbol{\Theta}^l}\mathcal{L}(\bm{x}_n, \bm{y}_n; \{\boldsymbol{\Theta}^l\}_{l = 1}^L),\label{layerwise_grad_eqn}\\
     \bm{S}^l &= \left(\dfrac{\text{abs}(\bm{G}^l)}{\text{abs}(\boldsymbol{\Theta}^l) + \epsilon}\right),\label{layerwise_score_eqn}\\
     \bm{M}_{ij}^l &= \begin{cases}1 &\text{if } \bm{S}_{ij}^l \geq s_t,\\0 &\text{otherwise,}\end{cases}\label{layerwise_mask_eqn}
\end{align}
where, \text{abs}(.) computes the element-wise absolute value of a matrix.

Now, on performing the layer-wise uniform random sampling of scores (with a small proportion $s$ of the total number of scores), since the full set of scores is still required for the computation of the binary mask (note that an element-wise comparison: $\mathcal{O}({D})$, is much less compute intensive than sorting: $\mathcal{O}({D\log{D}})$, for large $D$), we modify~\autoref{layerwise_grad_eqn},~\autoref{layerwise_score_eqn}, and~\autoref{layerwise_mask_eqn} such that for each layer $l$, $\bm{G}^l$, $\bm{S}^l$, and $\bm{M}^l$ respectively become:
\begin{align}
     \bm{G}^l &= \left(\dfrac{\dfrac{1}{d_s N}\sum\limits_{n = 1}^{d_sN}\text{abs}(\nabla_{\boldsymbol{\Theta}^l}\mathcal{L}(\bm{x}_n, \bm{y}_n; \{\boldsymbol{\Theta}^l\}_{l = 1}^L))}{\text{abs}(\boldsymbol{\Theta}^l) + \epsilon}\right),\label{new_grad_eqn}\\
     \bm{S}^l &\overset{{s}}{\sim} \mathcal{U}\left(\bm{G}^l\right),\label{new_score_eqn}\\
     \bm{M}^l &= \begin{cases}1 &\text{if } \bm{G}^l \geq s_t,\\0 &\text{otherwise.}\end{cases}\label{new_mask_eqn}
\end{align}

Moreover, since we now modify the gradients \emph{in place} using~\autoref{new_grad_eqn} and instead of storing all the $D$ scores (as in~\autoref{score_eqn}), we now store only $s\%$ of them (as in~\autoref{new_score_eqn}), we now save on $(100 - s)\%$ of the memory as well. 

\begin{figure*}[htbp]
    \begin{center}
        \subfigure[]{
            \label{fig_sss_piqa}
            \includegraphics[scale = 0.35]{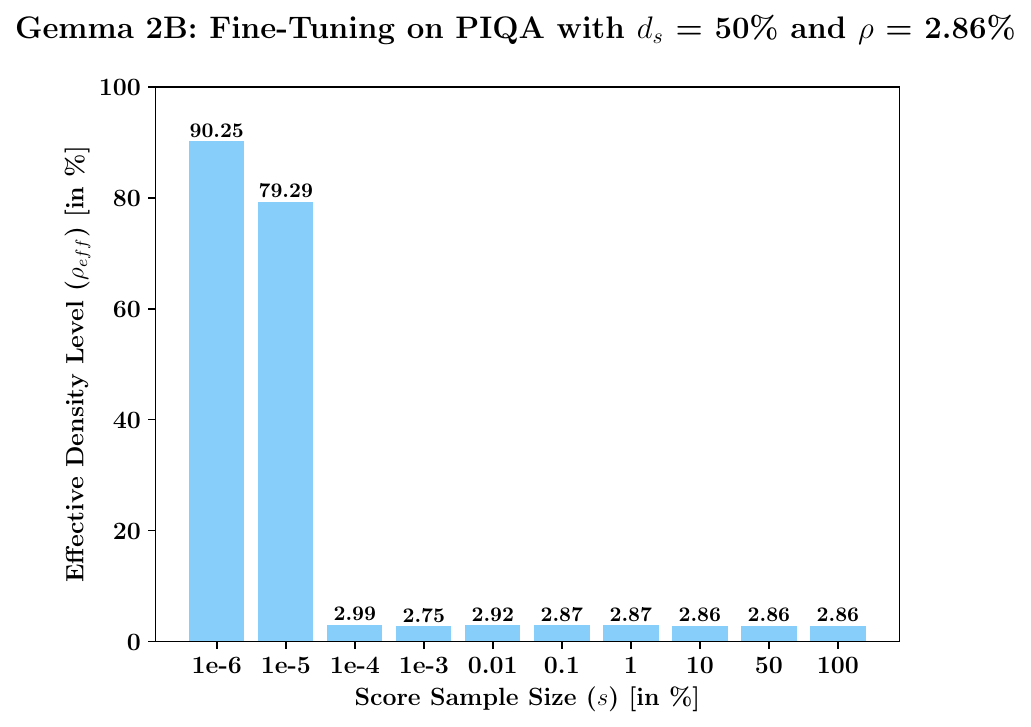}
        }
        \subfigure[]{
            \label{fig_sss_winogrande}
            \includegraphics[scale = 0.35]{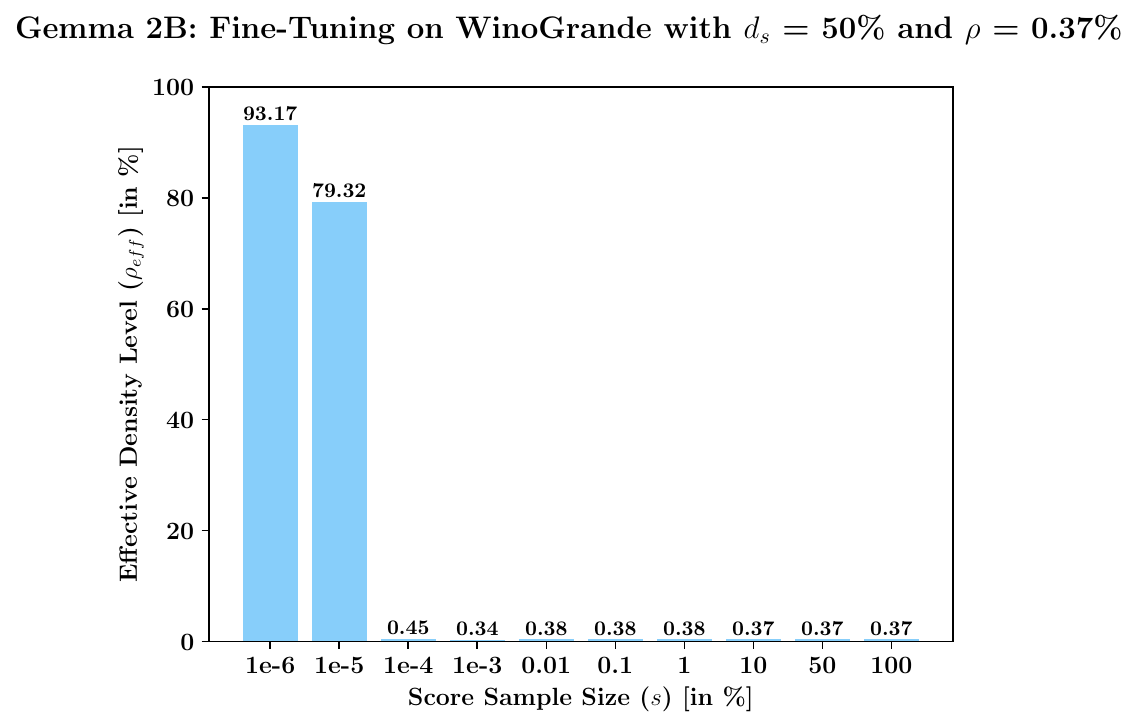}
        }
    \end{center}
    \caption{Representative experiments show that sampling only $s = 10\%$ of the scores allows us to attain a stable effective density level $\rho_{eff}$ which matches the required density level $\rho$ while drastically reducing the memory and compute requirements of GaLLoP.}
    \label{fig_CR_sss}
\end{figure*}

Throughout our experiments (see~\autoref{fig_CR_sss} for representative cases), we observe that with a dataset sample proportion ($d_s$) of 0.5, sampling only $s = 10\%$ of the scores to compute $s_t$ allows us to obtain the required $\rho\%$ of the parameters above the threshold which are to be selected for fine-tuning, i.e., the effective density level $\rho_{eff}$ matches the required density level $\rho$. This drastically reduces the compute and memory requirements of GaLLoP and hence, makes it highly efficient. 

An algorithmic implementation of GaLLoP is given in~\autoref{GaLLoP}.

\SetDataSty{textrm}
\begin{algorithm*}
\caption{GaLLoP}
\label{GaLLoP}
\SetKw{Global}{global}
\SetKwInput{Input}{Input}
\SetKwInput{Output}{Output}
\SetKwData{Numel}{$\mathtt{numel}$}
\SetKwData{Sorted}{$\mathtt{sorted}$}
\Input{Fine-Tuning Dataset ${\{(\bm{x}_n, \bm{y}_n)\}}_{i = 1}^{N}$, Number of Layers $L$, Density Level $\rho$, \newline Pre-Trained Model $F_{\{\boldsymbol{\Theta^l}\}_{l = 1}^L}(\bm{x})$, Learning Rate $\eta$, Number of Epochs $T$, \newline Dataset Sample Proportion $d_s$, Score Sample Proportion $s$, \newline Loss Function $\mathcal{L}(\bm{x}_n, \bm{y}_n; \{\boldsymbol{\Theta}^l\}_{l = 1}^L)$}
\BlankLine
\Global $D$, $\epsilon$ \\
$\epsilon \leftarrow 10^{-8}$\\
Initialize total parameter count $D \leftarrow 0$\\
\For{$l \leftarrow 1$ \KwTo $L$}{
$D \leftarrow D + \Numel(\boldsymbol{\Theta}^l)$\\
}
\textbf{\# Phase 1: Selection of Learnable Parameters}
\BlankLine
Initialize layer-wise gradient matrices $\bm{G}^l \leftarrow 0\quad\forall\quad l \in \{1, \dots, L\}$\\
\For{$n \leftarrow 1$ \KwTo $d_s N$}{
\For{$l \leftarrow 1$ \KwTo $L$}{
$\bm{G}^l \leftarrow \bm{G} ^l+ \dfrac{1}{d_s N}\nabla_{\boldsymbol{\Theta}^l}\mathcal{L}(\bm{x}_n, \bm{y}_n; \{\boldsymbol{\Theta}^l\}_{l = 1}^L)$
}}
Initialize layer-wise score matrices $\bm{S}^l \leftarrow 0\quad\forall\quad l \in \{1, \dots, L\}$\\
\For{$l \leftarrow 1$ \KwTo $L$}{
$\bm{G}^l  \leftarrow \left(\dfrac{\text{abs}(\bm{G}^l)}{\text{abs}(\boldsymbol{\Theta}^l) + \epsilon}\right)$\\
$\bm{S}^l \overset{{s}}{\sim} \mathcal{U}\left(\bm{G}^l\right)$
}
Compute threshold $s_t \leftarrow \Sorted_d(\bm{S}_{00}^0, \bm{S}_{01}^0, \dots ,\bm{S}_{\text{end}}^L)[\lfloor \rho D \rfloor]$\\
Compute layer-wise mask matrices $\bm{M}^l \leftarrow \begin{cases}1 &\text{if } \bm{G}^l \geq s_t,\\0 &\text{otherwise.}\end{cases} \quad \forall \quad l \in \{1, \dots, L\}$
\\ \textbf{\# Phase 2: Fine-Tuning of Learnable Parameters}
\BlankLine
Initialize new layer-wise parameter matrices $\tilde{\boldsymbol{\Theta}}^l \leftarrow \boldsymbol{\Theta}^l \quad\forall\quad l \in \{1, \dots, L\}$\\
\For{$t \leftarrow 1$ \KwTo $T$}{
\For{$l \leftarrow 1$ \KwTo $L$}{
$\tilde{\boldsymbol{\Theta}}^l \leftarrow \tilde{\boldsymbol{\Theta}}^l - \dfrac{1}{\eta}\left.\left(\bm{M}^l\,\odot\,\dfrac{1}{\left|\mathcal{B}_t\right|}\displaystyle\sum\limits_{(\bm{x}_n, \bm{y}_n) \in \mathcal{B}_t}\nabla_{\boldsymbol{\Theta}^l}\mathcal{L}(\bm{x}_n, \bm{y}_n; \{\boldsymbol{\Theta}^l\}_{l = 1}^L)\right)\right|_{\boldsymbol{\Theta}^l = \tilde{\boldsymbol{\Theta}}^l}$
}}
\Output{Fine-Tuned Model $\tilde{F}_{\{\tilde{\boldsymbol{\Theta}}^l\}_{l = 1}^L}(\bm{x})$}
\end{algorithm*}

\section{Datasets}
\label{AppendixA}

A detailed overview of the eight commonsense reasoning datasets which have been used throughout our experiments is as follows:
\begin{itemize}
    \item \textbf{ARC-c}: The \textbf{A}I2 \textbf{R}easoning \textbf{C}hallenge (ARC) dataset consists of natural, grade-school level science questions, authored for standardized tests taken by humans. ARC-c is the Challenge Set of this question bank, containing only those questions which have been incorrectly answered by both a retrieval-based and a word co-occurrence algorithm. Hence, high performance on this dataset requires AI models to possess advanced reasoning capabilities~\citep{arc}.
    \item \textbf{ARC-e}: The Easy Set of the ARC dataset which consists of the questions remaining in the ARC dataset post the formation of the Challenge Set~\citep{arc}.
    \item \textbf{BoolQ}: A dataset of naturally occurring, True/False (boolean) questions which have been formed from queries directed to the Google search engine. It requires AI models to utilize complex, non-factoid information for solving them~\citep{boolq}. Note that the version of the BoolQ dataset used by us -- which is created by~\citet{llm-adapters} -- does not include the (context) passages alongside the questions at all.
    \item \textbf{HellaSwag}: The \textbf{H}arder \textbf{E}ndings, \textbf{L}onger contexts, and \textbf{L}ow-shot \textbf{A}ctivities for \textbf{S}ituations \textbf{W}ith \textbf{A}dversarial \textbf{G}enerations dataset tests the robustness of AI models towards commonsense Natural Language Inference (NLI). It comprises questions formed from the ActivityNet captions dataset~\citep{activitynet} and the online WikiHow manuals with challenging, incorrect answer options obtained via Adversarial Filtering~\citep{hellaswag}.
    \item \textbf{OBQA}: The \textbf{O}pen\textbf{B}ook\textbf{QA} dataset tests the multi-hop reasoning ability of AI models in answering questions based on elementary-level science facts and commonsense knowledge. Note that the version of the OBQA dataset used by us -- which is created by~\citet{llm-adapters} -- does not include the (open book) scientific facts alongside the questions at all~\citep{obqa}.
    \item \textbf{PIQA}: The \textbf{P}hysical \textbf{I}nteraction: \textbf{Q}uestion \textbf{A}nswering dataset tests the physical commonsense of AI models by requiring knowledge of the physical properties of objects used in day-to-day life by humans to answer the `how-to' questions contained in it. Syntactically and topically similar semantic perturbations or alternative solutions have been introduced by annotators to counteract the possibility of spurious biases assisting AI models in achieving high performance~\citep{piqa}.
    \item \textbf{SIQA}: The \textbf{S}ocial \textbf{I}ntelligence \textbf{Q}uestion \textbf{A}nswering dataset tests the social and emotional intelligence of AI models with regards to everyday situations. Some incorrect answer options are obtained via question-switching around the same context so as to minimize the occurrence of stylistic artifacts arising from the cognitive biases of human annotators which could otherwise be exploited by an AI model for obtaining high performance~\citep{siqa}.
    \item \textbf{WinoGrande}: This large-scale fill-in-the-blank dataset tests pronoun resolution-based commonsense reasoning capabilities of AI models and is unsolvable upon their complete reliance on embedding associations. Systematic algorithmic bias reduction is performed via AfLite, a lightweight and improved Adversarial Filtering algorithm~\citep{winogrande}.
\end{itemize}

The sizes of the training and test splits of each of these datasets as well as their response formats are provided in~\autoref{tab:datasets}.

\begin{table}[htbp]
\centering
\caption{Details of all the datasets~\citep{llm-adapters} used across all our experiments.}
\vskip 0.1in
\small
\begin{tabular}{cccc}
\toprule
\textbf{Dataset} & \textbf{Training Set Size} & \textbf{Test Set Size} & \textbf{Response Format} \\ 
\midrule
\multirow{2}{*}{ARC-c} & \multirow{2}{*}{1119} & \multirow{2}{*}{1172} & the correct answer is answer$\langle\text{ID}\rangle$ \\ 
&& & ID = \{1,2,3,4\} \\
\midrule
\multirow{2}{*}{ARC-e} & \multirow{2}{*}{2251} & \multirow{2}{*}{2376} & the correct answer is answer$\langle\text{ID}\rangle$\\
&& & ID = \{1,2,3,4\} \\
\midrule
\multirow{2}{*}{BoolQ} & \multirow{2}{*}{9427} & \multirow{2}{*}{3270} & the correct answer is $\langle\text{BOOL}\rangle$\\
&& & BOOL = \{true/false\} \\
\midrule
\multirow{2}{*}{HellaSwag} & \multirow{2}{*}{39905} & \multirow{2}{*}{10042} & the correct answer is ending$\langle\text{ID}\rangle$\\
&& & ID = \{1,2,3,4\} \\
\midrule
\multirow{2}{*}{OBQA} & \multirow{2}{*}{4957} & \multirow{2}{*}{500} & the correct answer is answer$\langle\text{ID}\rangle$ \\ 
&& & ID = \{1,2,3,4\} \\
\midrule
\multirow{2}{*}{PIQA} & \multirow{2}{*}{16113} & \multirow{2}{*}{1837} & the correct answer is solution$\langle\text{ID}\rangle$\\
&& & ID = \{1,2\} \\
\midrule
\multirow{2}{*}{SIQA} & \multirow{2}{*}{33410} & \multirow{2}{*}{1954} & the correct answer is answer$\langle\text{ID}\rangle$\\
&& & ID = \{1,2,3\} \\
\midrule
\multirow{2}{*}{WinoGrande} & \multirow{2}{*}{63238} & \multirow{2}{*}{1267} & the correct answer is option$\langle\text{ID}\rangle$\\
&& & ID = \{1,2\} \\
\bottomrule
\end{tabular}
\label{tab:datasets}
\vskip -0.1in
\end{table}

\section{Hyperparameters for Fine-Tuning Algorithms}
\label{AppendixB}

We fix the maximum input sequence length to 512 in all our experiments and utilize the chunked cross-entropy loss for fine-tuning, so as to save memory by upcasting only a single chunk (token) at a time from BF16 to FP32 while computing the loss. We perform early stopping in order to select the best fine-tuned model checkpoint across the three epochs, based on the ID accuracy on the test-dev set (here, the ID test set serves as the test-dev set for a given experimental run). We set $\alpha = 0.5$ for WiSE-FT and $\alpha = \beta = 0.5$ for LiNeS following the overall recommendations of their authors~\citep{wise_ft,lines}. The specific hyperparameters are as follows:

\subsection{GaLLoP and SAFT}

\autoref{tab:gallop_saft_hyperparameters} shows the hyperparameter configurations employed upon performing fine-tuning with GaLLoP and SAFT. We use a low learning rate of 2e-5 in order to prevent divergences during fine-tuning.

\begin{table}[htbp]
\centering
\caption{Hyperparameter configurations of GaLLoP and SAFT used while fine-tuning LLaMA3 8B and Gemma 2B models. * indicates that Effective Batch Size = (Per-GPU Batch Size $\times$ Gradient Accumulation Steps $\times$ No. of GPUs).}
\vskip 0.1in
\small
\begin{tabular}{ccc}
\toprule
\textbf{Hyperparameters} & \textbf{LLaMA3 8B} & \textbf{Gemma 2B} \\ \midrule
Dataset Sample Proportion $d_s$ & \multicolumn{2}{c}{0.5}  \\ 
Score Sample Proportion $s$ & \multicolumn{2}{c}{0.1} \\ 
Optimizer & \multicolumn{2}{c}{Fused AdamW ($\beta_1$ = 0.9; $\beta_1$ = 0.999)}  \\ 
Weight Decay & \multicolumn{2}{c}{0} \\
Learning Rate & \multicolumn{2}{c}{2e-5}  \\
Learning Rate Scheduler & \multicolumn{2}{c}{Cosine} \\
Warmup Steps & \multicolumn{2}{c}{100} \\
Effective Batch Size* & \multicolumn{2}{c}{16} \\
Per-GPU Batch Size & \multicolumn{2}{c}{4} \\
Gradient Accumulation Steps & 1 & 2 \\
No. of GPUs & 4 & 2 \\
Epochs & \multicolumn{2}{c}{3} \\
\bottomrule
\end{tabular}
\label{tab:gallop_saft_hyperparameters}
\vskip -0.1in
\end{table}

\subsection{SpIEL}

\autoref{tab:spiel_hyperparameters} shows the hyperparameter configurations employed upon performing fine-tuning with SpIEL. Most of these are in line with those mentioned in~\citep{spiel}. However, unlike~\citet{spiel} and just like with GaLLoP and SAFT, we find that a lower learning rate works better since it prevents divergences during fine-tuning. Moreover, during the course of our experiments, we find that adding weight decay upon fine-tuning with SpIEL does not help. Upon exploring the addition of weight decay from amongst the following four values: \{0, 3, 10, 30\} (same as those examined by~\citet{spiel}) while fine-tuning LLaMA3 8B, we find that adding any amount of weight decay leads to a comparable/slightly-lowered performance with/than a zero weight decay. Hence, we do not use weight decay while fine-tuning models using SpIEL.

\begin{table}[htbp]
\centering
\caption{Hyperparameter configurations of SpIEL used while fine-tuning LLaMA3 8B and Gemma 2B models. * indicates that Effective Batch Size = (Per-GPU Batch Size $\times$ Gradient Accumulation Steps $\times$ No. of GPUs).}
\vskip 0.1in
\small
\begin{tabular}{ccc}
\toprule
\textbf{Hyperparameters} & \textbf{LLaMA3 8B} & \textbf{Gemma 2B} \\ \midrule
Optimizer & \multicolumn{2}{c}{AdamW ($\beta_1$ = 0.9; $\beta_1$ = 0.999)}  \\ 
Weight Decay & \multicolumn{2}{c}{0} \\
Learning Rate & \multicolumn{2}{c}{2e-5}  \\
Learning Rate Scheduler & \multicolumn{2}{c}{Linear} \\
Warmup Ratio & \multicolumn{2}{c}{0.03} \\
Effective Batch Size & \multicolumn{2}{c}{16} \\
Per-GPU Batch Size & \multicolumn{2}{c}{16} \\
Gradient Accumulation Steps & \multicolumn{2}{c}{1} \\
No. of GPUs & \multicolumn{2}{c}{1} \\
Epochs & \multicolumn{2}{c}{3} \\
\bottomrule
\end{tabular}
\label{tab:spiel_hyperparameters}
\vskip -0.1in
\end{table}

\subsection{FFT}

\autoref{tab:fft_hyperparameters} shows the hyperparameter configurations employed upon performing fine-tuning with FFT. Just like with GaLLoP and SAFT, we use a low learning rate of 2e-5 in order to prevent divergences during fine-tuning.

\begin{table}[htbp]
\centering
\caption{Hyperparameter configurations of FFT used while fine-tuning LLaMA3 8B and Gemma 2B models. * indicates that Effective Batch Size = (Per-GPU Batch Size $\times$ Gradient Accumulation Steps $\times$ No. of GPUs).}
\vskip 0.1in
\small
\begin{tabular}{ccc}
\toprule
\textbf{Hyperparameters} & \textbf{LLaMA3 8B} & \textbf{Gemma 2B} \\ \midrule
Optimizer & \multicolumn{2}{c}{Fused AdamW ($\beta_1$ = 0.9; $\beta_1$ = 0.999)}  \\ 
Weight Decay & \multicolumn{2}{c}{0} \\
Learning Rate & \multicolumn{2}{c}{2e-5}  \\
Learning Rate Scheduler & \multicolumn{2}{c}{Cosine} \\
Warmup Steps & \multicolumn{2}{c}{100} \\
Effective Batch Size & \multicolumn{2}{c}{16} \\
Per-GPU Batch Size & \multicolumn{2}{c}{4} \\
Gradient Accumulation Steps & \multicolumn{2}{c}{1} \\
No. of GPUs & \multicolumn{2}{c}{4} \\
Epochs & \multicolumn{2}{c}{3} \\
\bottomrule
\end{tabular}
\label{tab:fft_hyperparameters}
\vskip -0.1in
\end{table}

\subsection{LoRA and DoRA}

\autoref{tab:lora_hyperparameters} and~\autoref{tab:dora_hyperparameters} show the hyperparameter configurations employed upon performing fine-tuning with LoRA and DoRA (respectively). Most of these are in line with those mentioned in~\citep{dora}. Unlike GaLLoP and SAFT, we observe that a low learning rate of 2e-5 slows down learning and results in decreased ID and OOD performance for both LoRA and DoRA. Hence, we use the learning rates of 3e-4 and 1e-4 suggested by~\citet{dora} for faster (and divergence-free) learning and better performance for LoRA and DoRA (respectively). Moreover, during the course of our experiments, we find that adding dropout upon fine-tuning with LoRA and DoRA does not help. Upon exploring the addition of dropout from amongst the following four values: \{0, 0.05, 0.10, 0.20\} while fine-tuning Gemma 2B, we find that adding any amount of dropout leads to a comparable/slightly-lowered performance with/than the case when no dropout is used. Hence, we do not use dropout while fine-tuning models using LoRA and DoRA.

\begin{table}[htbp]
\centering
\vskip -0.1in
\caption{Hyperparameter configurations of LoRA used while fine-tuning LLaMA3 8B and Gemma 2B models. * indicates that Effective Batch Size = (Per-GPU Batch Size $\times$ Gradient Accumulation Steps $\times$ No. of GPUs).}
\small
\begin{tabular}{ccc}
\toprule
\textbf{Hyperparameters} & \textbf{LLaMA3 8B} & \textbf{Gemma 2B} \\ \midrule
Optimizer & \multicolumn{2}{c}{Fused AdamW ($\beta_1$ = 0.9; $\beta_1$ = 0.999)}  \\ 
Weight Decay & \multicolumn{2}{c}{0.01} \\
Learning Rate & \multicolumn{2}{c}{3e-4}  \\
Learning Rate Scheduler & \multicolumn{2}{c}{Cosine} \\
Warmup Steps & \multicolumn{2}{c}{100} \\
Effective Batch Size* & \multicolumn{2}{c}{16} \\
Per-GPU Batch Size & \multicolumn{2}{c}{4} \\
Gradient Accumulation Steps & 1 & 2 \\
No. of GPUs & 4 & 2 \\
Epochs & \multicolumn{2}{c}{3} \\
Target Modules & \multicolumn{2}{c}{q\_proj, k\_proj, v\_proj, up\_proj, down\_proj}\\
\bottomrule
\end{tabular}
\label{tab:lora_hyperparameters}
\vskip -0.1in
\end{table}

\begin{table}[htbp]
\centering
\vskip -0.1in
\caption{Hyperparameter configurations of DoRA used while fine-tuning LLaMA3 8B and Gemma 2B models. * indicates that Effective Batch Size = (Per-GPU Batch Size $\times$ Gradient Accumulation Steps $\times$ No. of GPUs).}
\small
\begin{tabular}{ccc}
\toprule
\textbf{Hyperparameters} & \textbf{LLaMA3 8B} & \textbf{Gemma 2B} \\ \midrule
Optimizer & \multicolumn{2}{c}{Fused AdamW ($\beta_1$ = 0.9; $\beta_1$ = 0.999)}  \\ 
Weight Decay & \multicolumn{2}{c}{0.01} \\
Learning Rate & \multicolumn{2}{c}{1e-4} \\
Learning Rate Scheduler & \multicolumn{2}{c}{Cosine} \\
Warmup Steps & \multicolumn{2}{c}{100} \\
Effective Batch Size* & \multicolumn{2}{c}{16} \\
Per-GPU Batch Size & \multicolumn{2}{c}{4} \\
Gradient Accumulation Steps & 1 & 2 \\
No. of GPUs & 4 & 2 \\
Epochs & \multicolumn{2}{c}{3} \\
Target Modules & \multicolumn{2}{c}{q\_proj, k\_proj, v\_proj, up\_proj, down\_proj}\\
\bottomrule
\end{tabular}
\label{tab:dora_hyperparameters}
\vskip -0.1in
\end{table}

\section{Hyperparameters for Decoding}
\label{AppendixC}

The values of the hyperparameters used for decoding the LLM-generated outputs are provided in~\autoref{tab:decoding_hyperparameters} and have been taken \emph{as-is} from~\citep{llm-adapters} so as to maintain consistency with prior work.

\begin{table}[!htb]
\centering
\vskip -0.1in
\caption{Hyperparameters for decoding the output generated by the fine-tuned models into text. Values have been taken as-is from~\citep{llm-adapters}.}
\small
\begin{tabular}{ccc}
\toprule
\textbf{Hyperparameters} & \textbf{Values} \\ \midrule
Temperature $T$ & 0.1  \\ 
$k$ (as in top-$k$) & 40 \\ 
$p$ (as in top-$p$) & 0.75  \\ 
Number of Beams (as in Beam Search) & 4 \\
\bottomrule
\end{tabular}
\label{tab:decoding_hyperparameters}
\vskip -0.1in
\end{table}

\section{Per-Run Results of our Experiments}
\label{AppendixD}

\subsection{ID and OOD Accuracy}
\label{app:id_ood}

\begin{figure*}[htbp]
    \begin{center}
        \subfigure[]{
            \label{fig_CR_llama_arc_c}
            \includegraphics[scale = 0.25]{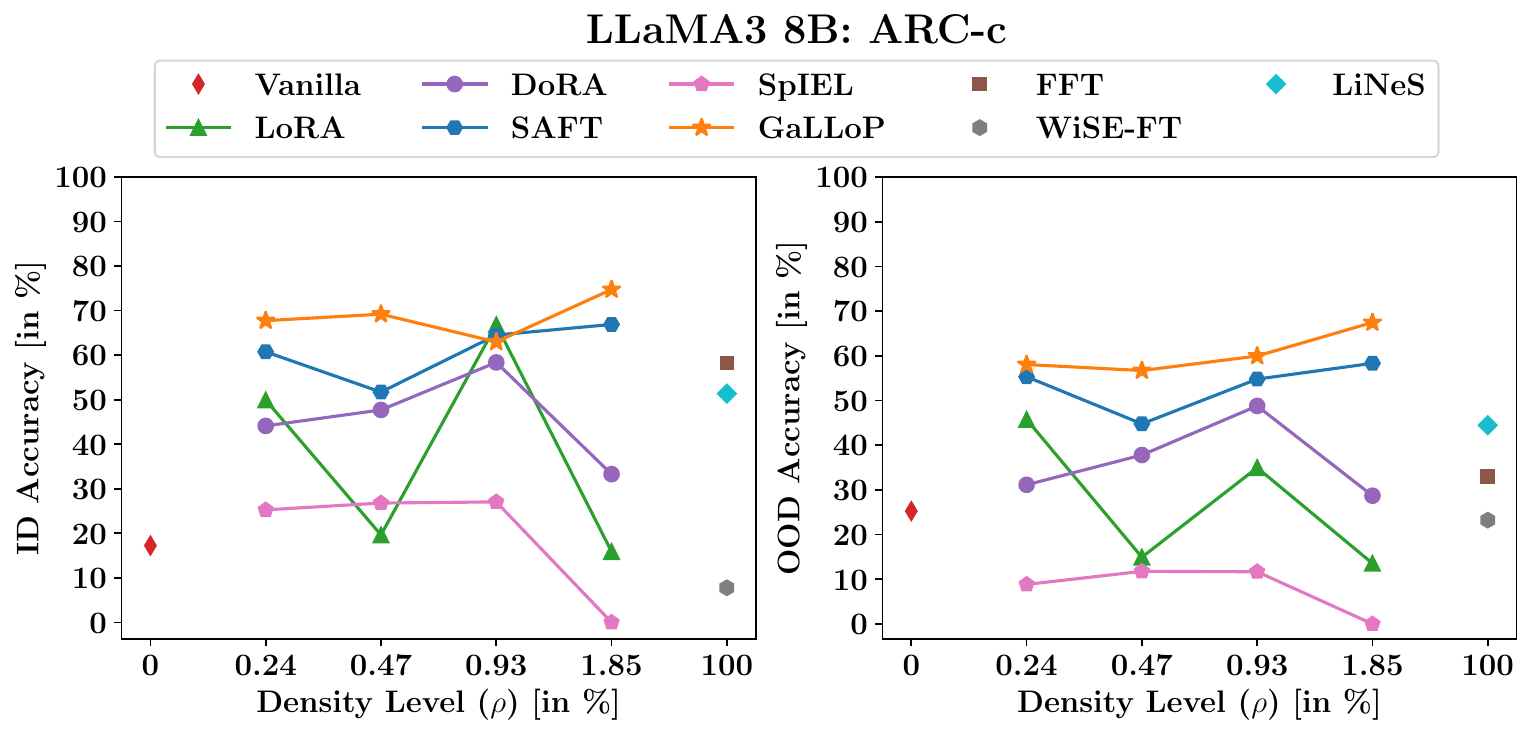}
        }
        \subfigure[]{
            \label{fig_CR_llama_arc_e}
            \includegraphics[scale = 0.25]{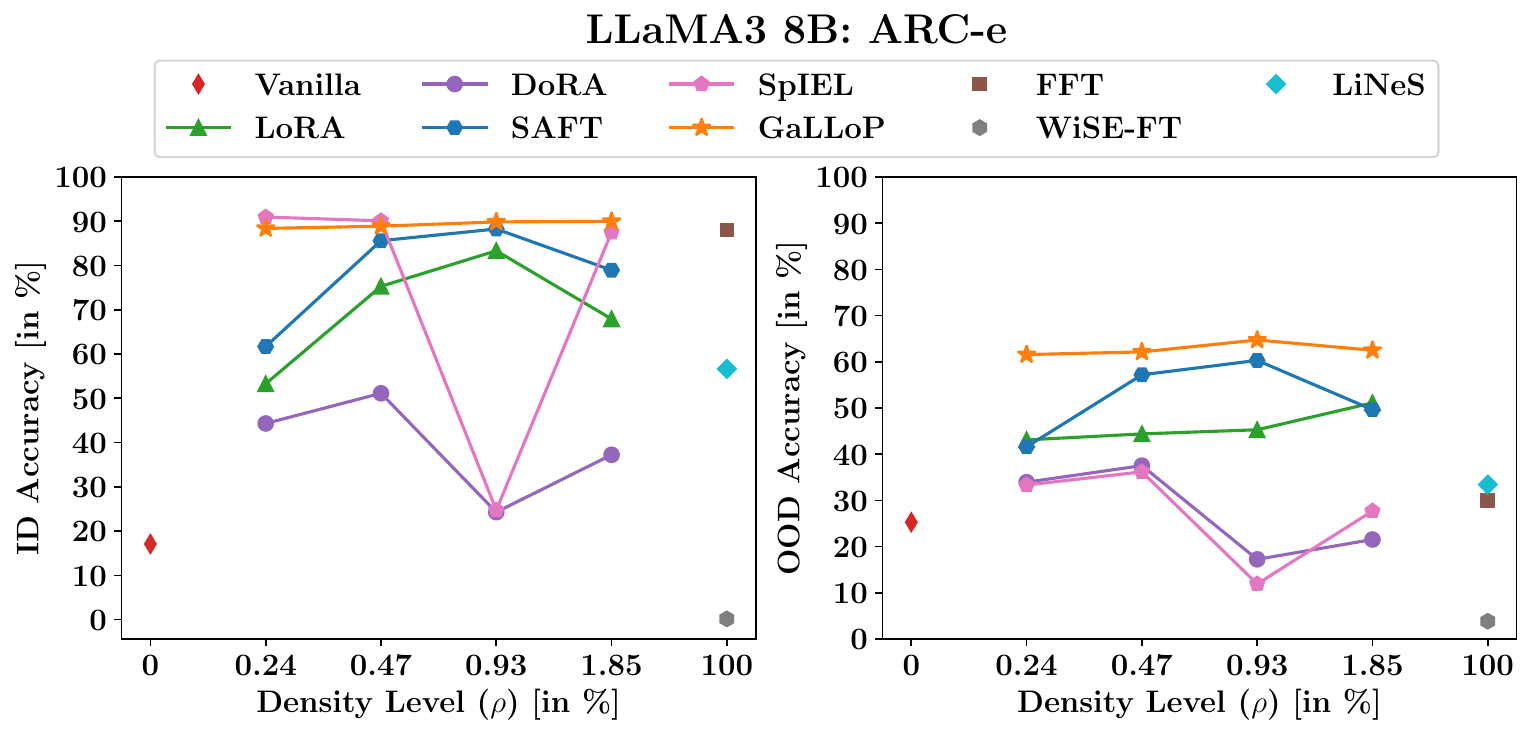}
        }
        \subfigure[]{
            \label{fig_CR_llama_hellaswag}
            \includegraphics[scale = 0.25]{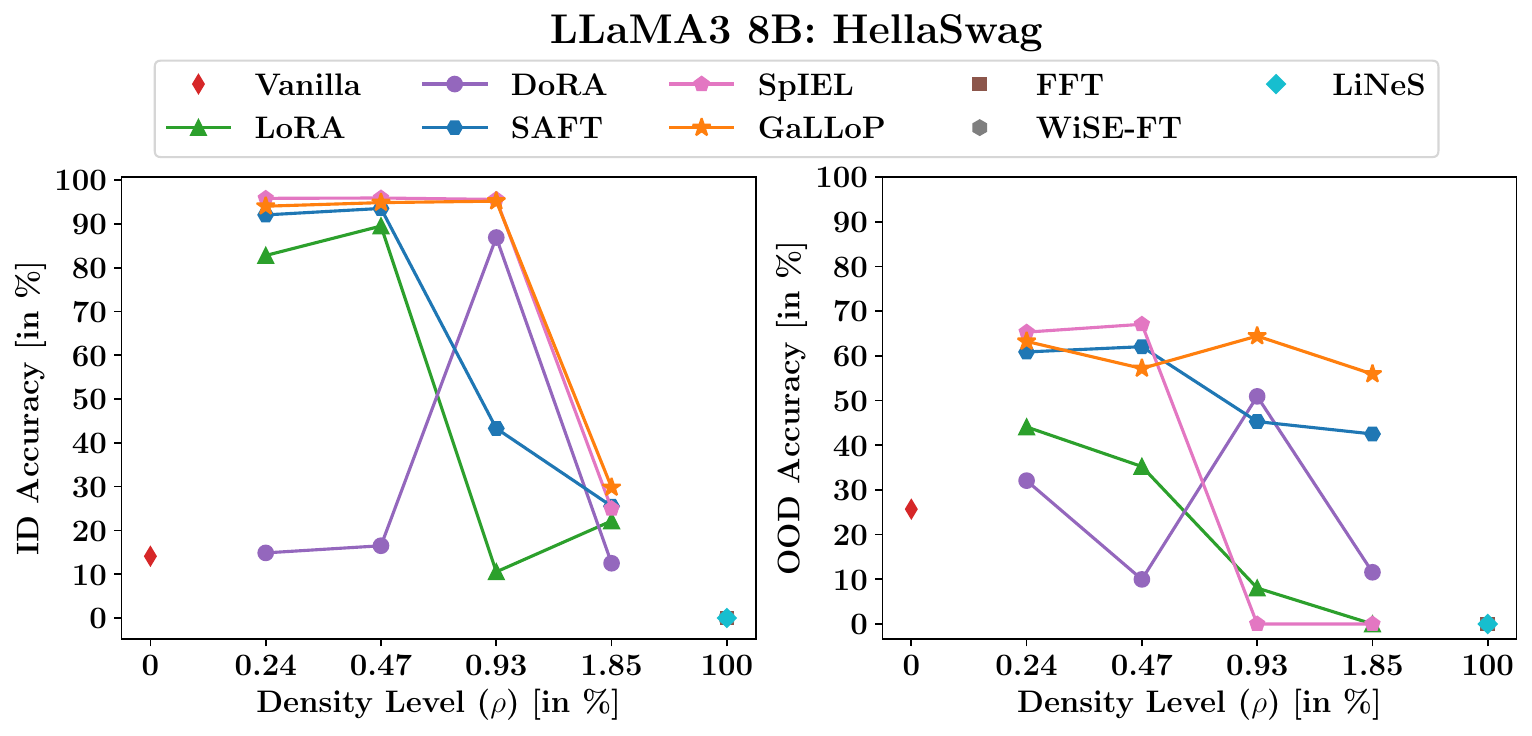}
        }
        \subfigure[]{
            \label{fig_CR_llama_obqa}
            \includegraphics[scale = 0.25]{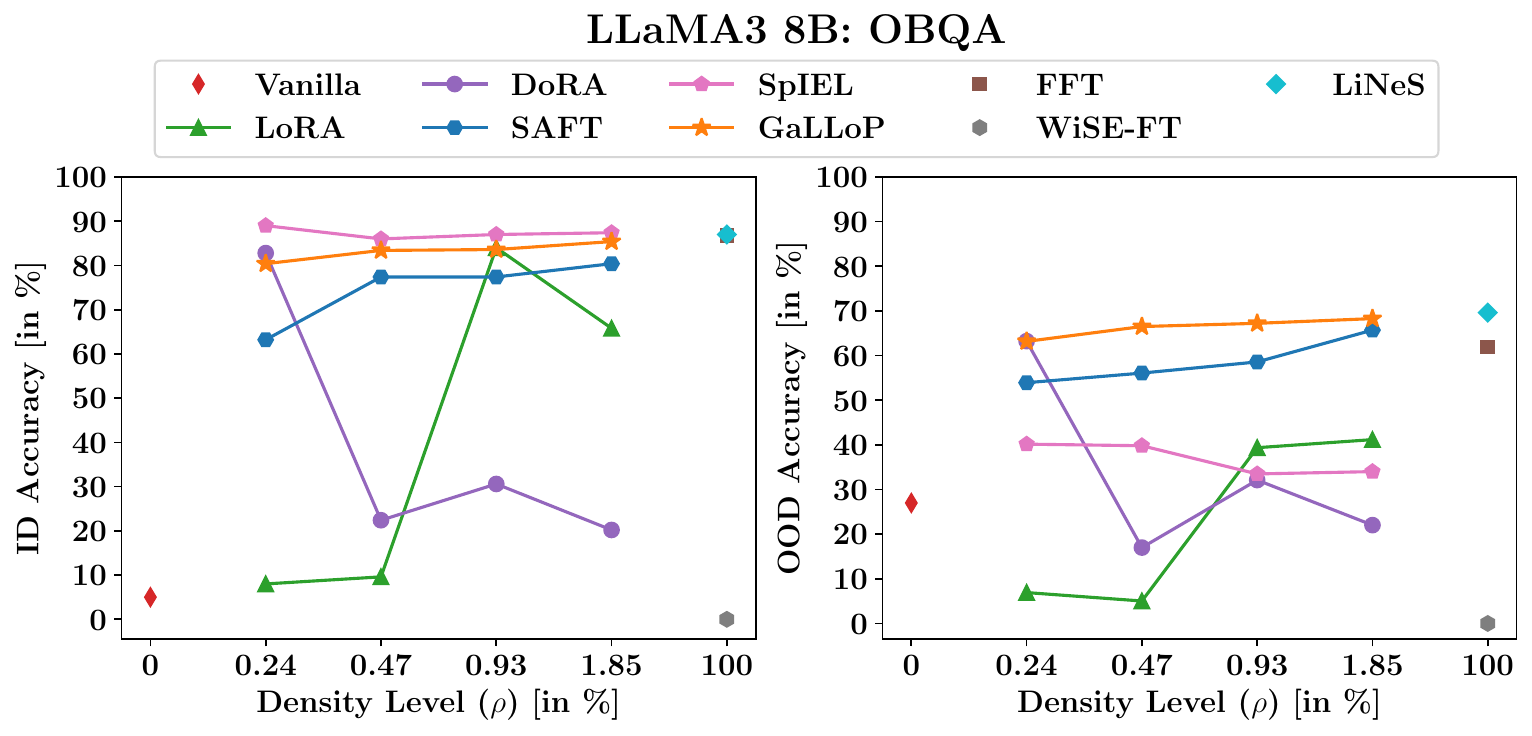}
        }
        \subfigure[]{
            \label{fig_CR_llama_piqa}
            \includegraphics[scale = 0.25]{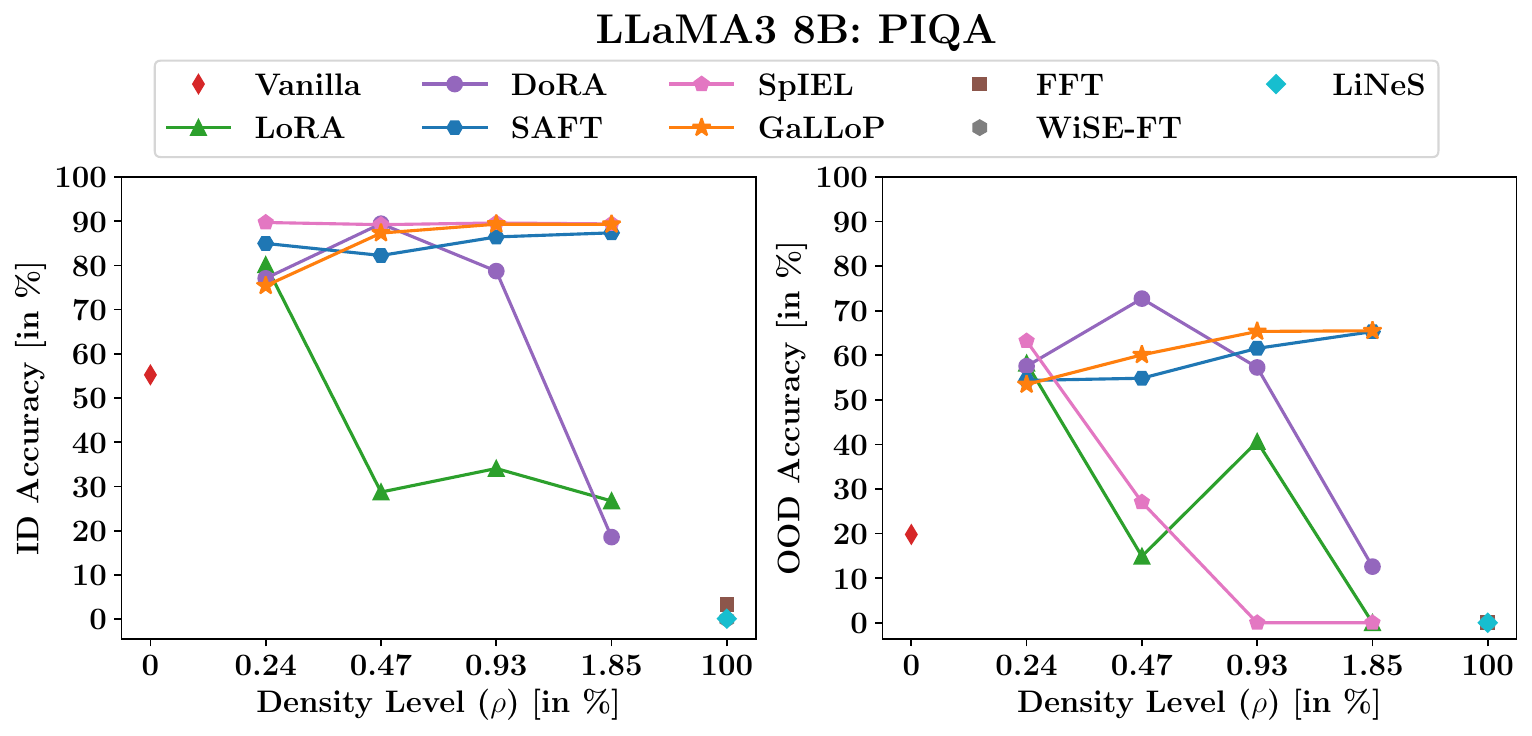}
        }
        \subfigure[]{
            \label{fig_CR_llama_winogrande}
            \includegraphics[scale = 0.25]{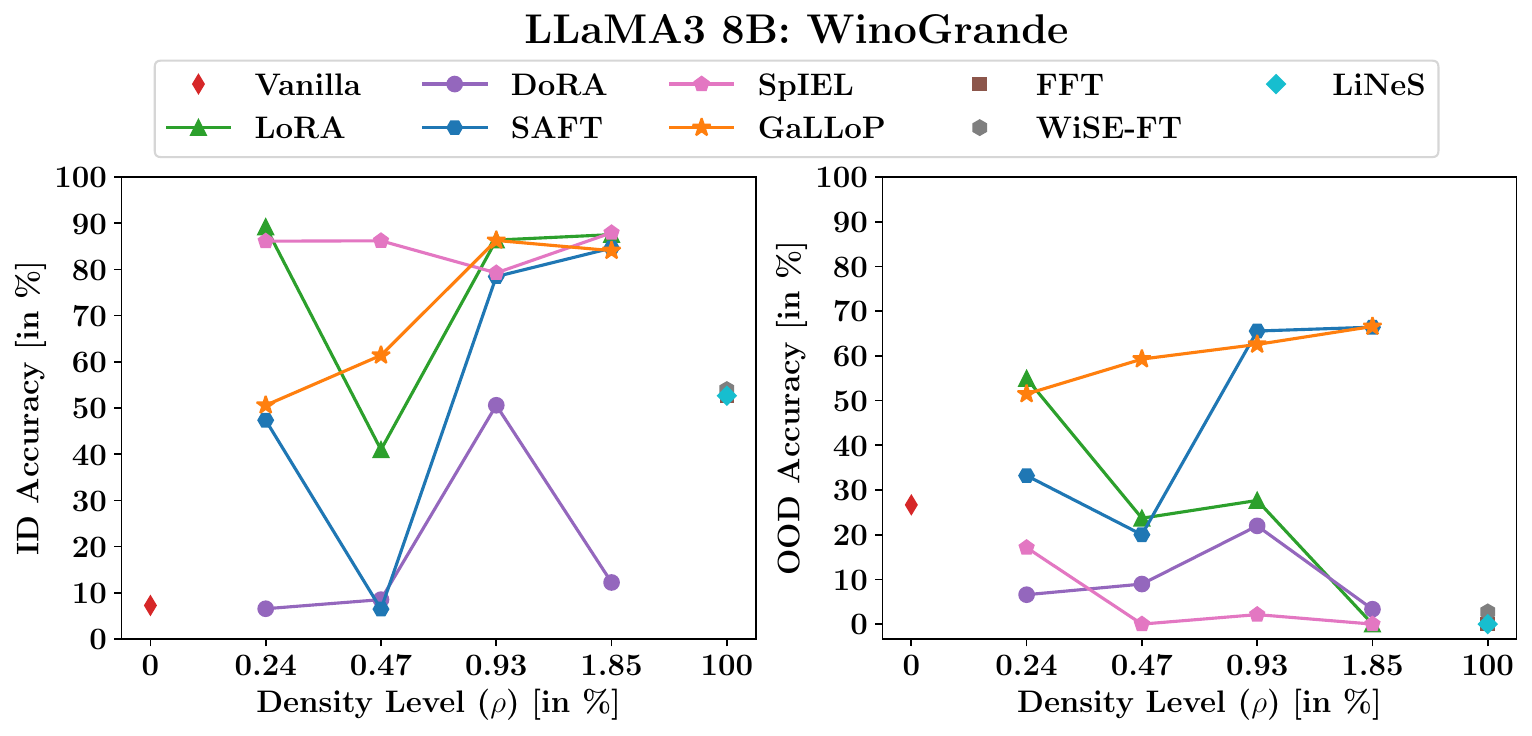}
        }
        \subfigure[]{
            \label{fig_CR_llama_siqa}
            \includegraphics[scale = 0.25]{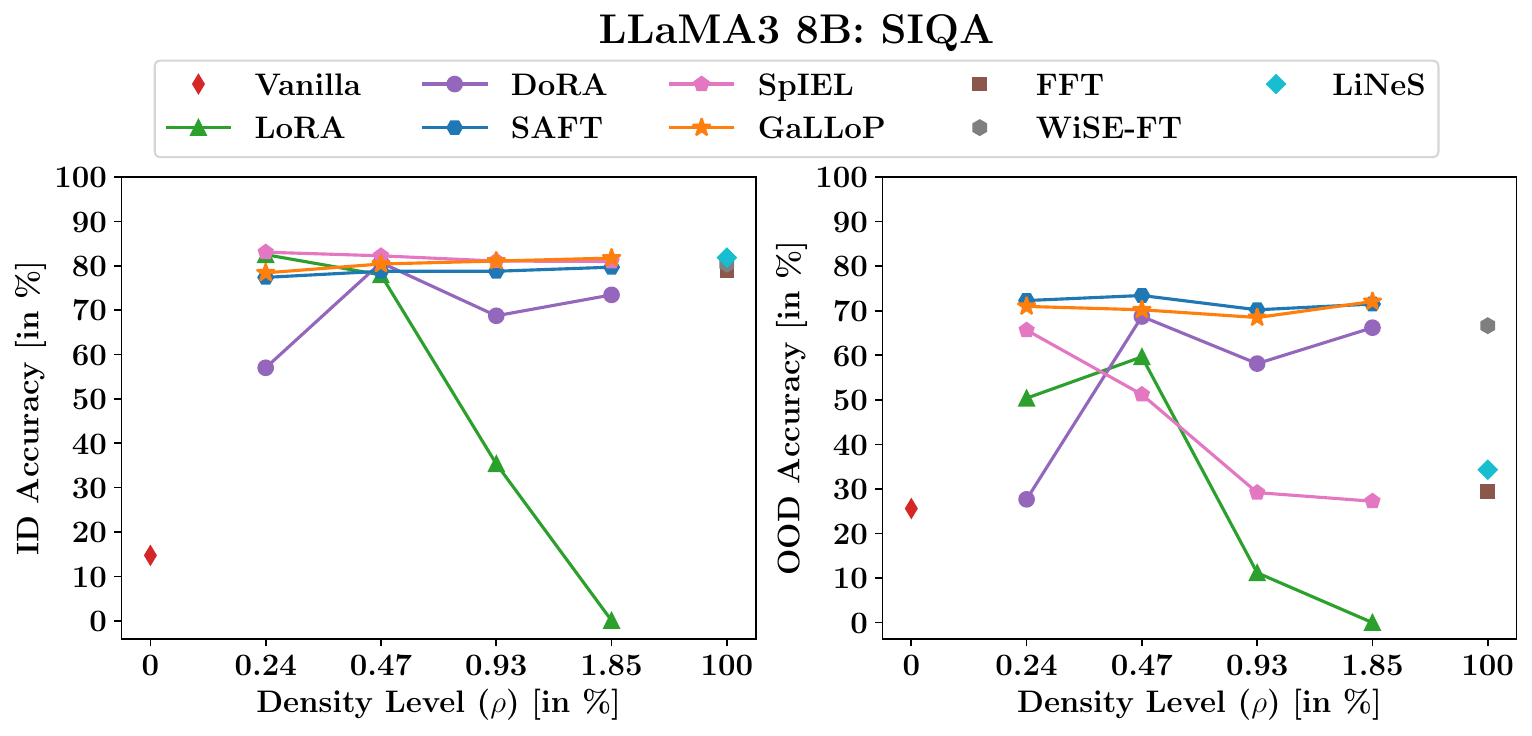}
        }
        \subfigure[]{
            \label{fig_CR_llama_boolq}
            \includegraphics[scale = 0.25]{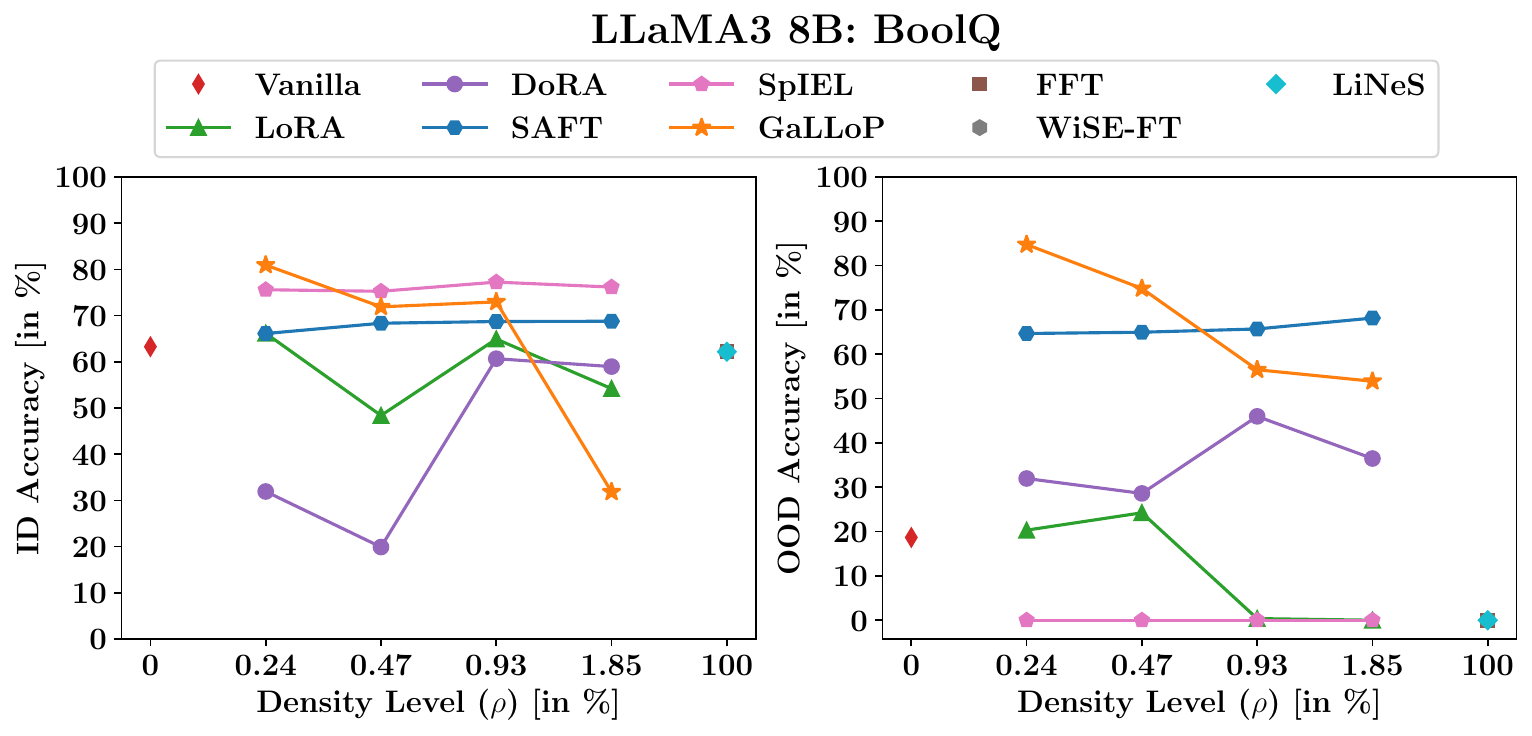}
        }
    \end{center}
    \caption{LLaMA3 8B models fine-tuned with GaLLoP attain the most stable and highest/high levels of ID and OOD performance across all density levels when fine-tuning is performed on a) ARC-c, b) ARC-e, c) HellaSwag, d) OBQA, e) PIQA, f) WinoGrande, g) SIQA, and h) BoolQ (except OOD).}
    \label{fig_CR_llama_per_run}
\end{figure*}

For LLaMA3 8B, models fine-tuned on the ARC-c and ARC-e datasets using GaLLoP form a dominant Pareto front for both ID as well as OOD performance over models fine-tuned and/or edited on the same datasets using the other algorithms (respectively; see~\autoref{fig_CR_llama_arc_c} and~\autoref{fig_CR_llama_arc_e}). While models fine-tuned on the HellaSwag, OBQA, PIQA, and WinoGrande datasets using SpIEL outperform those fine-tuned on the same datasets using GaLLoP by a slight margin on the ID test sets, models fine-tuned with GaLLoP form a dominant Pareto front on the corresponding OOD test sets on which the models fine-tuned with SpIEL perform quite poorly and with a performance drop that widens as we approach the higher density levels (respectively; see~\autoref{fig_CR_llama_hellaswag},~\autoref{fig_CR_llama_obqa},~\autoref{fig_CR_llama_piqa}, and~\autoref{fig_CR_llama_winogrande}). On the SIQA dataset, the model fine-tuned with SAFT is quite competitive and has an almost similar/comparable level of performance with the model fine-tuned with GaLLoP (see~\autoref{fig_CR_llama_siqa}). 

\begin{figure*}[htbp]
    \begin{center}
        \subfigure[]{
            \label{fig_CR_gemma_arc_c}
            \includegraphics[scale = 0.25]{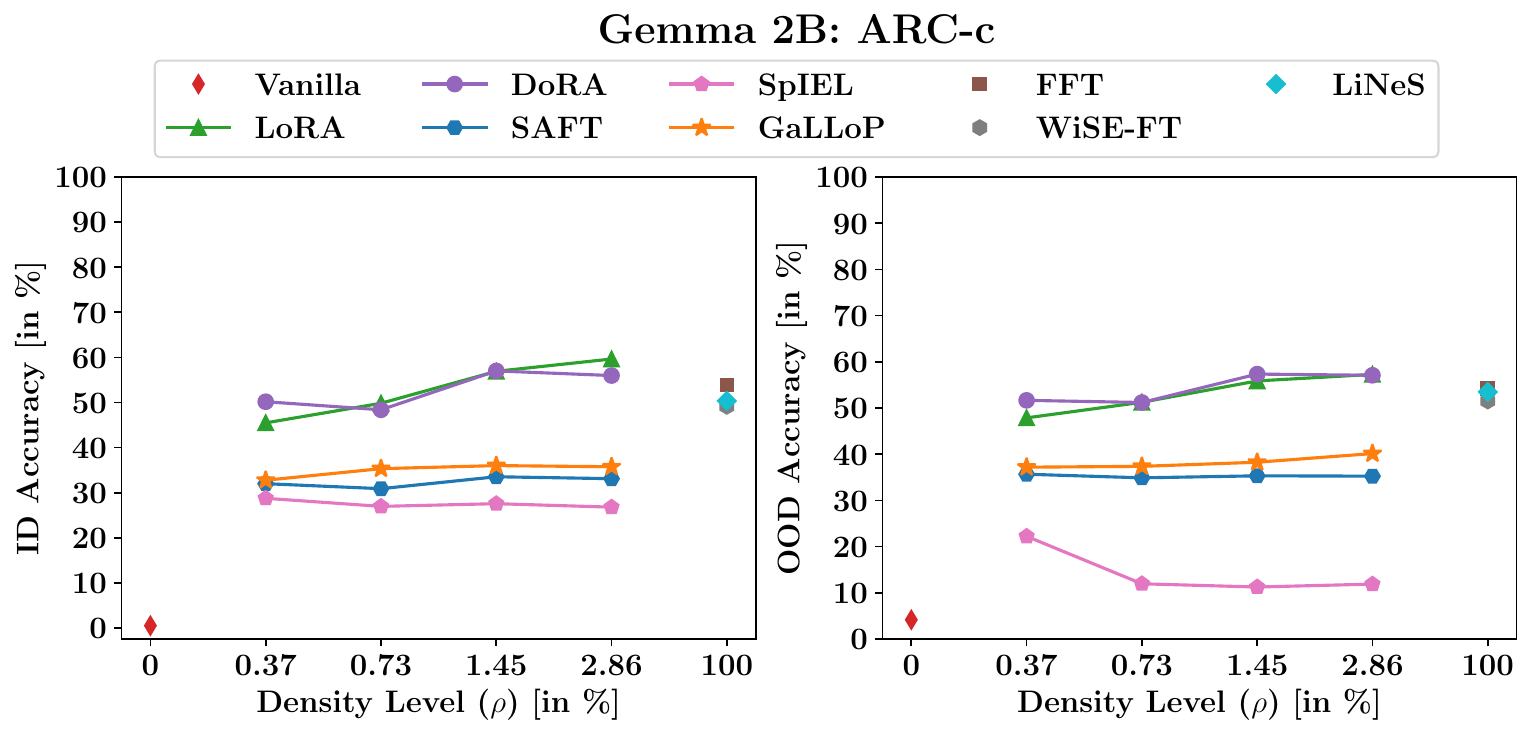}
        }
        \subfigure[]{
            \label{fig_CR_gemma_arc_e}
            \includegraphics[scale = 0.25]{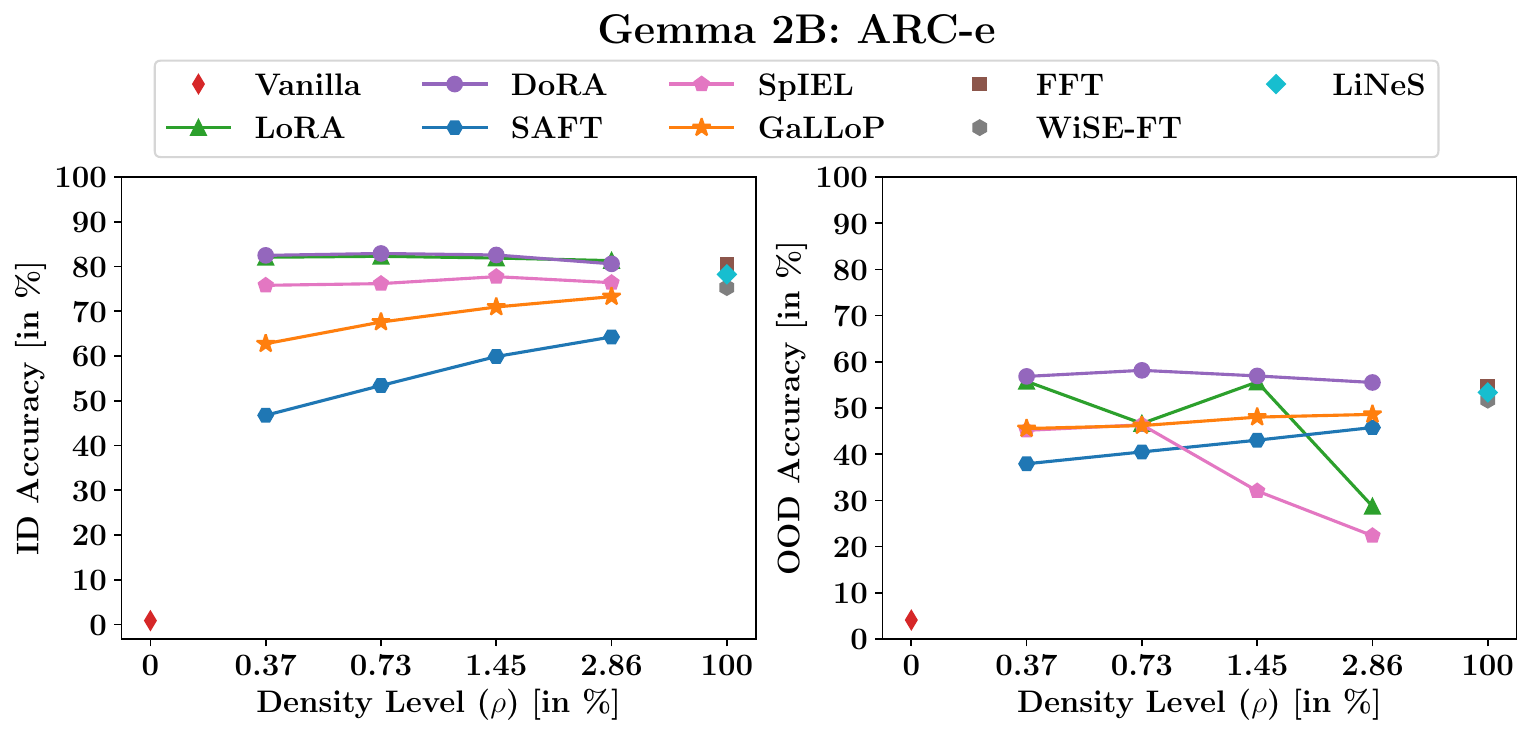}
        }
    \end{center}
    \caption{While Gemma 2B models fine-tuned on a) ARC-c and b) ARC-e with GaLLoP attain stable and high levels of ID and OOD performance across all density levels, Gemma 2B models fine-tuned on the same datasets with DoRA attain the highest levels of ID and OOD performance across all density levels owing to the two training sets possessing the smallest sizes.}
    \label{fig_CR_gemma_arc_c_arc_e}
\end{figure*}

\begin{figure*}[htbp]
    \begin{center}
        \subfigure[]{
            \label{fig_CR_gemma_obqa}
            \includegraphics[scale = 0.25]{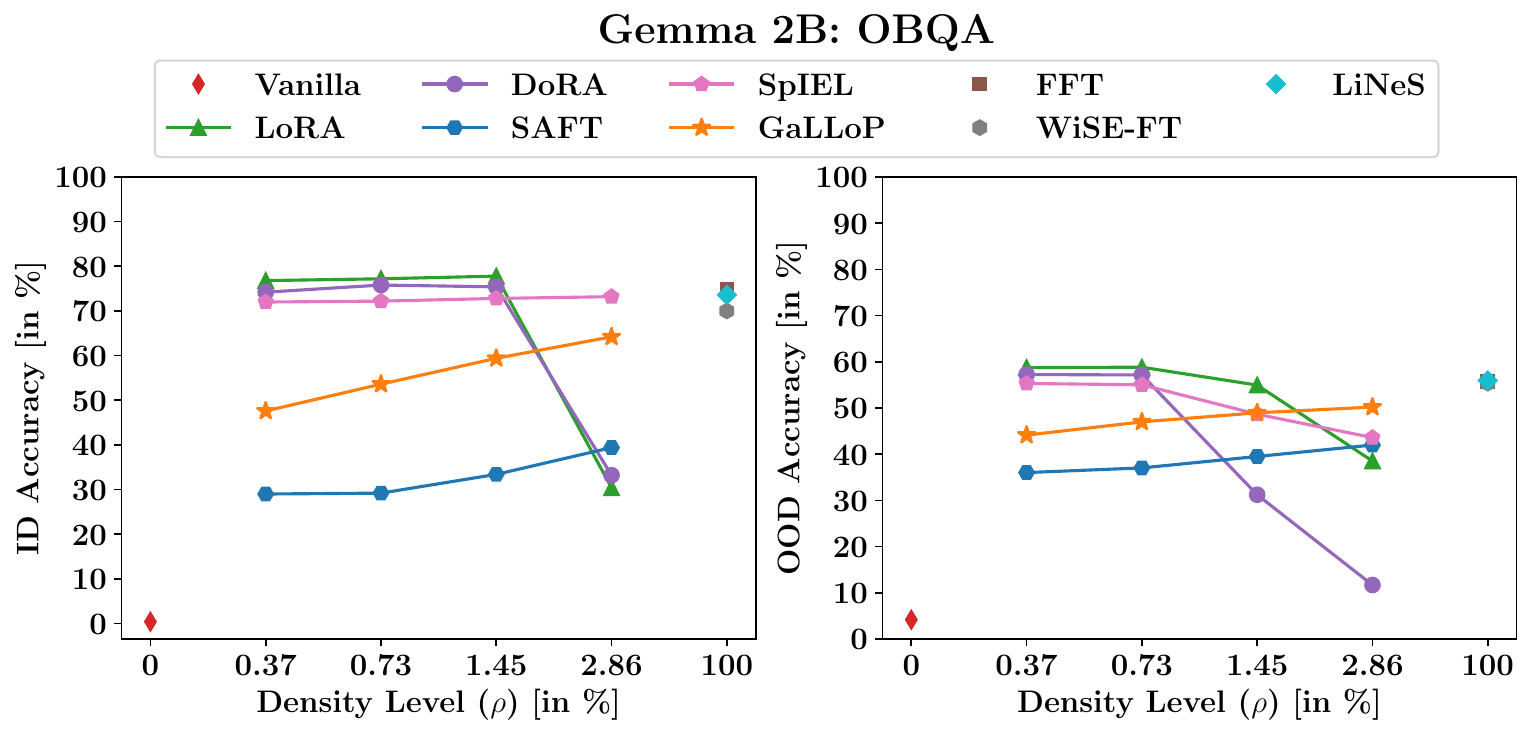}
        }
        \subfigure[]{
            \label{fig_CR_gemma_boolq}
            \includegraphics[scale = 0.25]{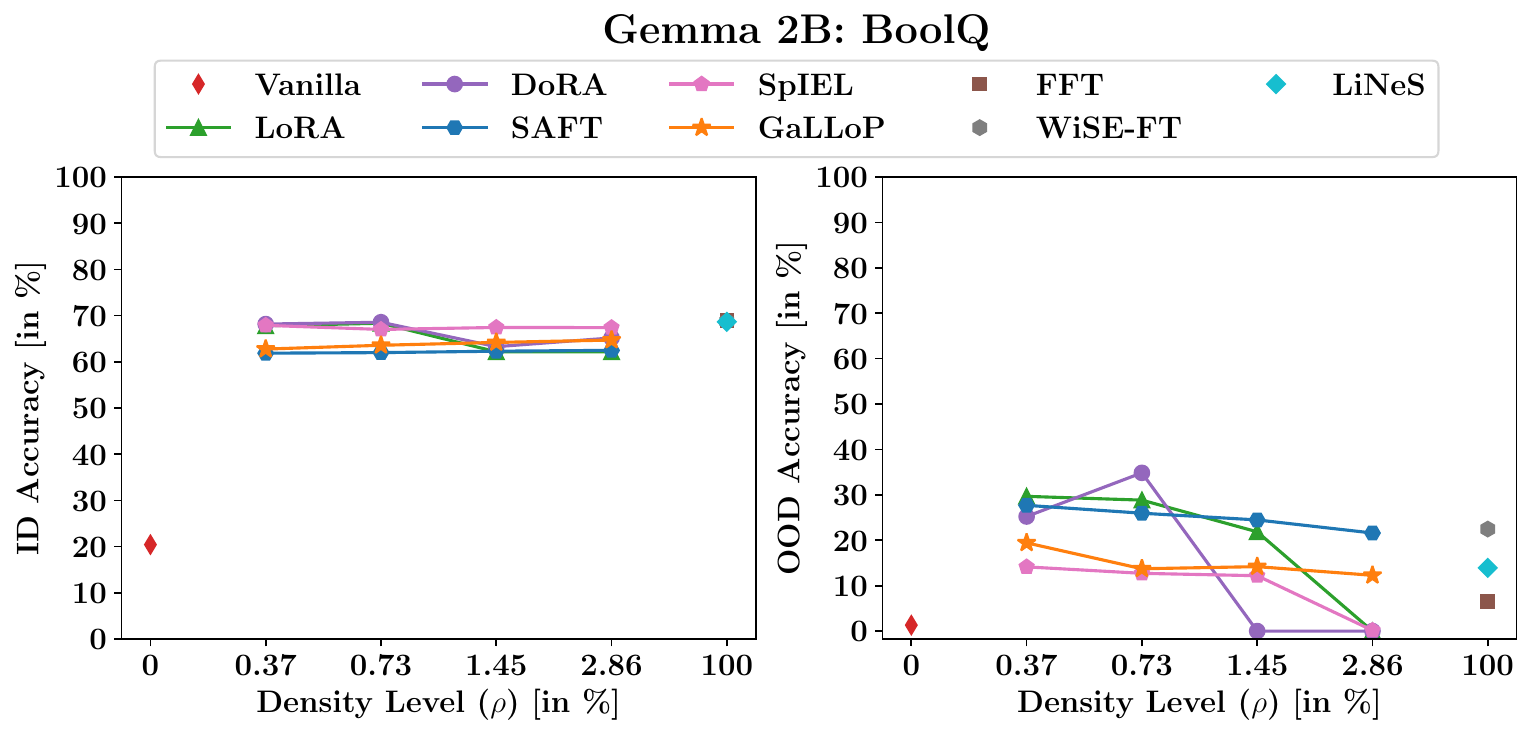}
        }
        \subfigure[]{
            \label{fig_CR_gemma_piqa}
            \includegraphics[scale = 0.25]{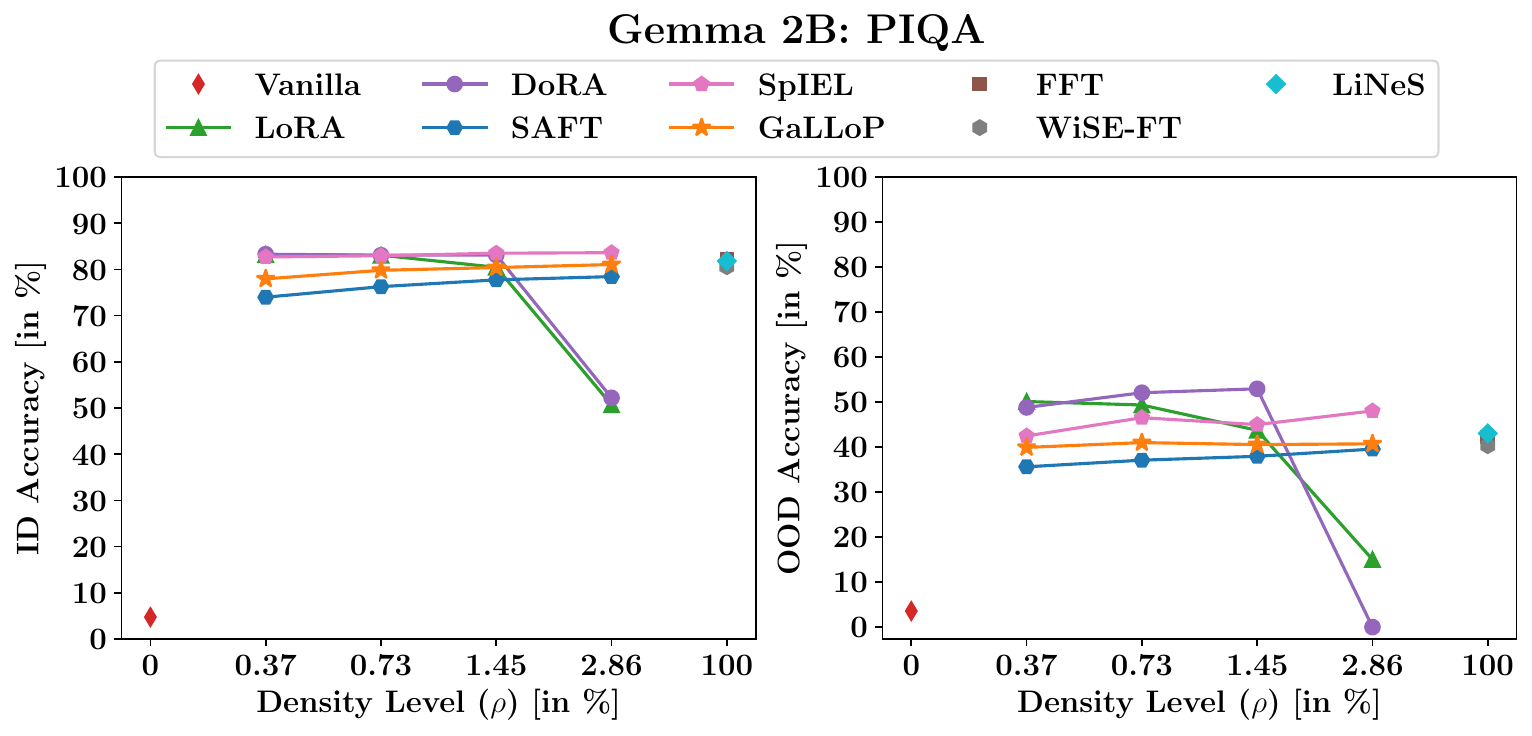}
        }
    \end{center}
    \caption{While Gemma 2B models fine-tuned on a) OBQA, b) BoolQ, and c) PIQA with GaLLoP attain stable and high levels of ID and OOD performance across all density levels, Gemma 2B models fine-tuned with LoRA and DoRA tend to overfit on their moderately-large sized training sets and show drastic ID and/or OOD performance drops.}
    \label{fig_CR_gemma_obqa_boolq_piqa}
\end{figure*}

\begin{figure*}[htbp]
    \begin{center}
        \subfigure[]{
            \label{fig_CR_gemma_siqa}
            \includegraphics[scale = 0.25]{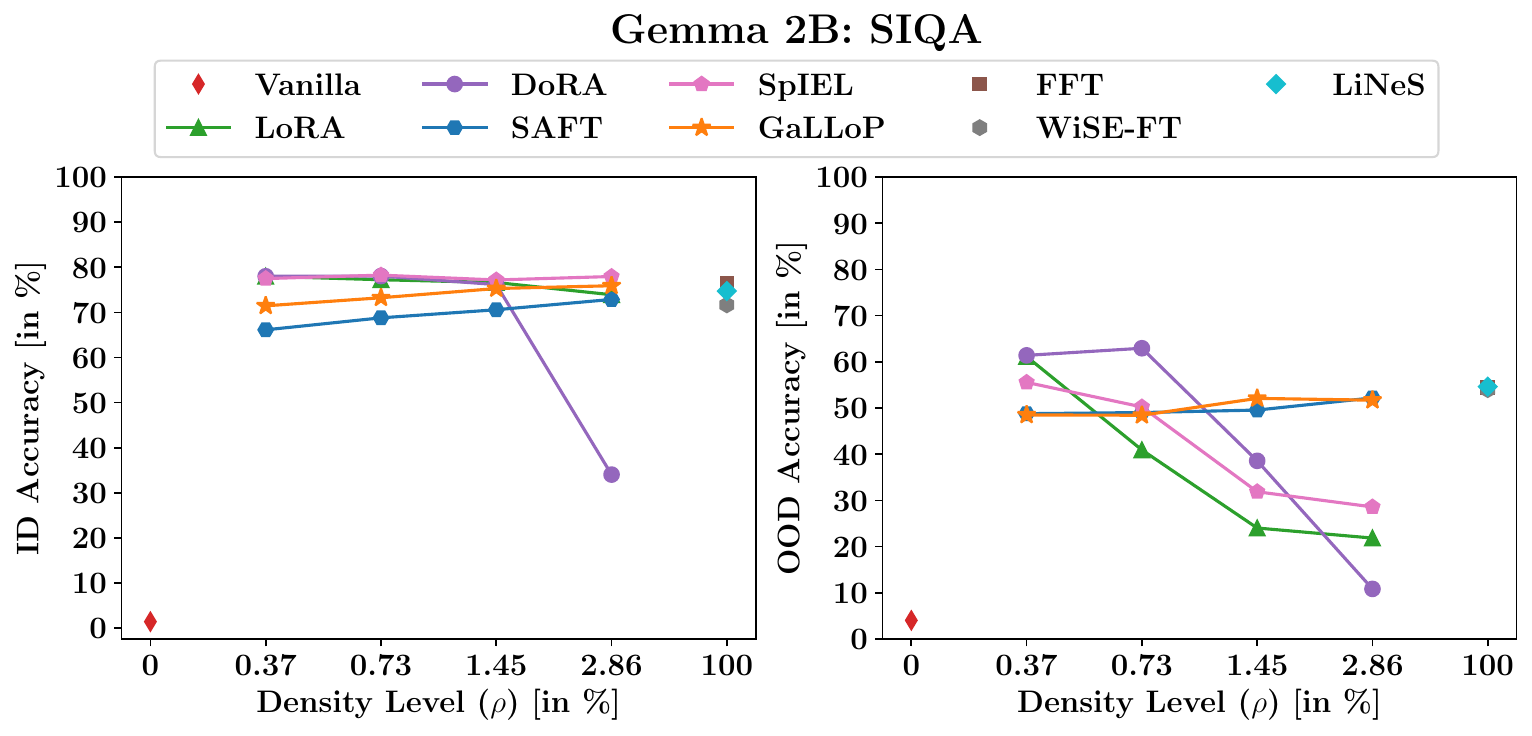}
        }
        \subfigure[]{
            \label{fig_CR_gemma_hellaswag}
            \includegraphics[scale = 0.25]{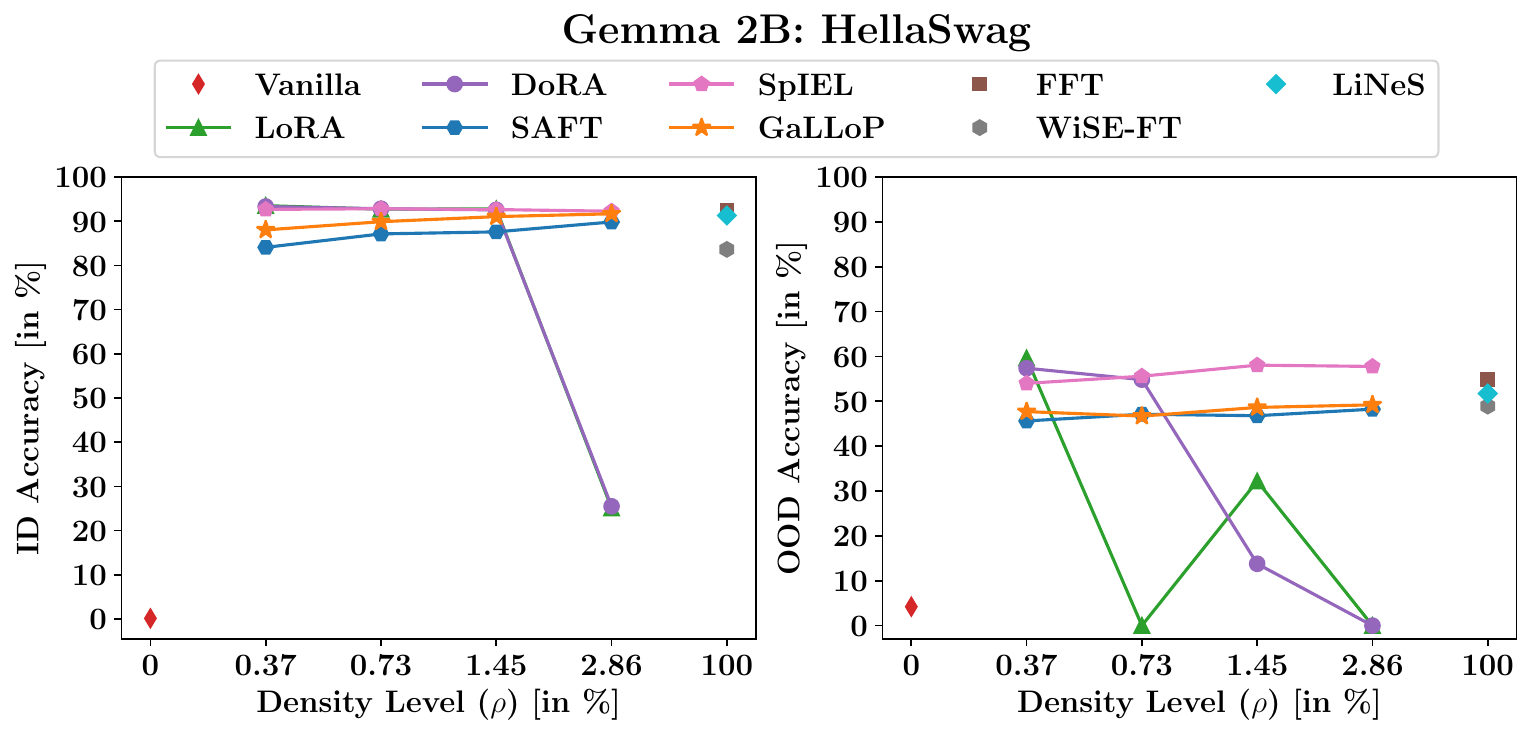}
        }
        \subfigure[]{
            \label{fig_CR_gemma_winogrande}
            \includegraphics[scale = 0.25]{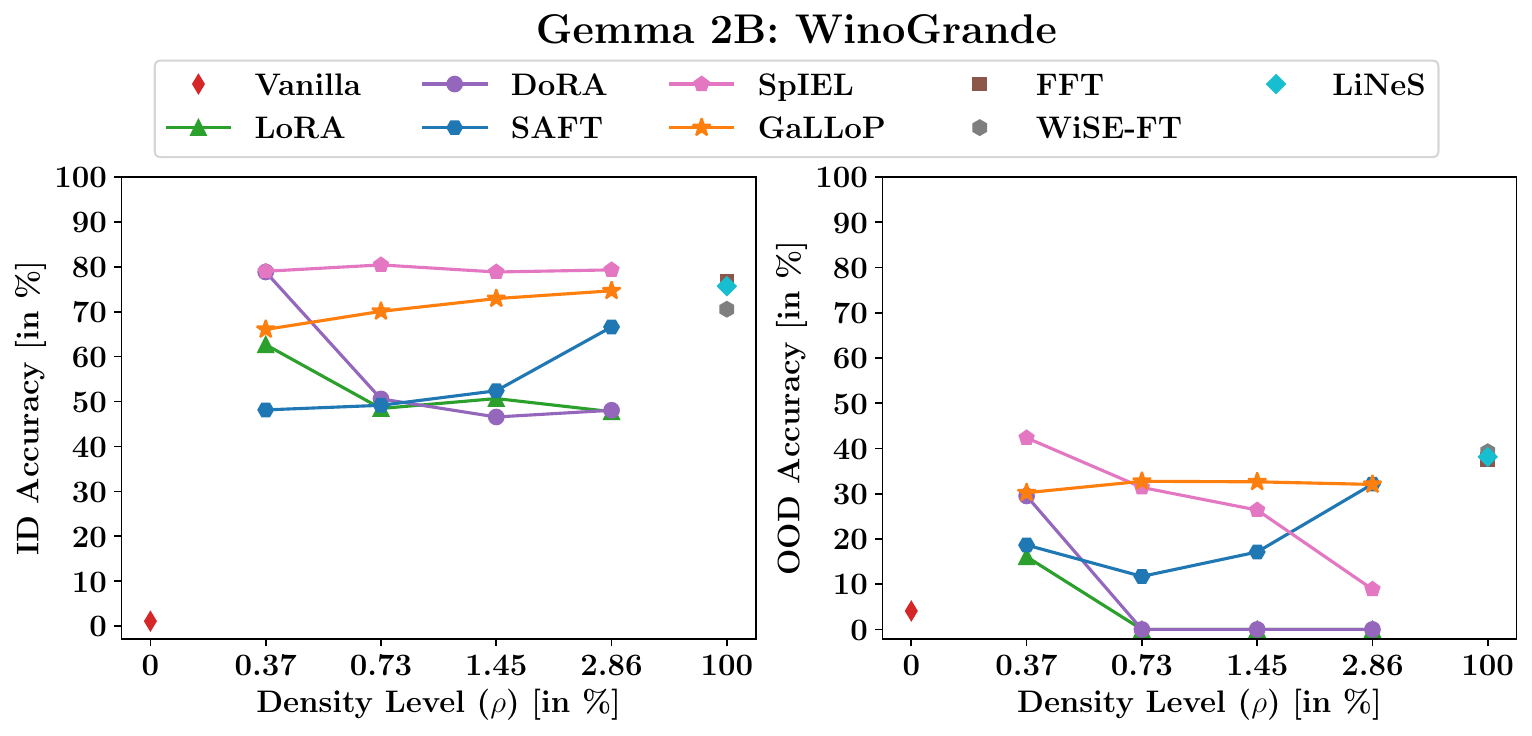}
        }
    \end{center}
    \caption{While Gemma 2B models fine-tuned on a) SIQA, b) HellaSwag, and c) WinoGrande with GaLLoP attain the most stable and the highest/high levels of ID and OOD performance across all density levels, Gemma 2B models fine-tuned with LoRA or DoRA tend to overfit on their moderately-large/large sized training sets and show drastic ID and/or OOD performance drops.}
    \label{fig_CR_gemma_siqa_hellaswag_winogrande}
\end{figure*}

For Gemma 2B, since pre-training has been performed on a much less number of tokens (= 6T) as compared to LLaMA3 8B (= 15T), there is a significantly lower risk of fine-tuning sensitivity due to overtraining~\citep{overtraining}. Hence, Gemma 2B models fine-tuned with some of the competing PEFT algorithms (LoRA, DoRA, SpIEL) and even FFT (and hence, WiSE-FT and LiNeS) perform much better than their LLaMA3 8B counterparts. However, these models still perform quite poorly when it comes to OOD tasks and display high performance drops as we approach the higher density levels. These drops in performance are not only present for the OOD tasks (LoRA, DoRA, and SpIEL) but are also present for the ID tasks (LoRA and DoRA). Moreover, an important trend is noticeable across all these datasets: overfitting across all the density levels with the increase in the size of the dataset used for fine-tuning. While models fine-tuned with LoRA and DoRA attain higher accuracies as compared to models fine-tuned with GaLLoP on both ID and OOD tasks upon utilizing small datasets (ARC-c and ARC-e; models fine-tuned with LoRA already start exhibiting large OOD performance drops upon increasing the density level for fine-tuning on ARC-e) for fine-tuning (see~\autoref{fig_CR_gemma_arc_c_arc_e}), they tend to overfit and break down with drastic OOD performance drops along with even moderately-high/high ID performance drops for higher density levels as move towards utilizing moderately-large sized datasets (OBQA, BoolQ, PIQA, SIQA, HellaSwag; see~\autoref{fig_CR_gemma_obqa_boolq_piqa},~\autoref{fig_CR_gemma_siqa}, and~\autoref{fig_CR_gemma_hellaswag}) and completely break down on both the ID and OOD tasks upon fine-tuning on WinoGrande, the largest dataset (see~\autoref{fig_CR_gemma_winogrande}). In contrast, models fine-tuned with GaLLoP demonstrate high and stable performance across all density levels and upon fine-tuning on all the datasets (\emph{individually}) with sizes spread across the spectrum, and form a dominant Pareto front over all the other fine-tuned models when fine-tuning is performed on WinoGrande, the largest dataset (see~\autoref{fig_CR_gemma_arc_c_arc_e},~\autoref{fig_CR_gemma_obqa_boolq_piqa}, and~\autoref{fig_CR_gemma_siqa_hellaswag_winogrande}). This shows that fine-tuning with GaLLoP makes models the most robust to overfitting. Finally, on shifting our focus to models fine-tuned with SAFT, we can clearly see that on average (across all datasets: see~\autoref{fig_CR_full_avg_gemma}) as well as individually, across seven out of the eight datasets (see~\autoref{fig_CR_gemma_arc_c_arc_e},~\autoref{fig_CR_gemma_obqa},~\autoref{fig_CR_gemma_piqa}, and~\autoref{fig_CR_gemma_siqa_hellaswag_winogrande}), models fine-tuned with GaLLoP form a dominant Pareto front over models fine-tuned with SAFT.  

\begin{figure*}[htbp]
\centering
\includegraphics[scale=0.25]{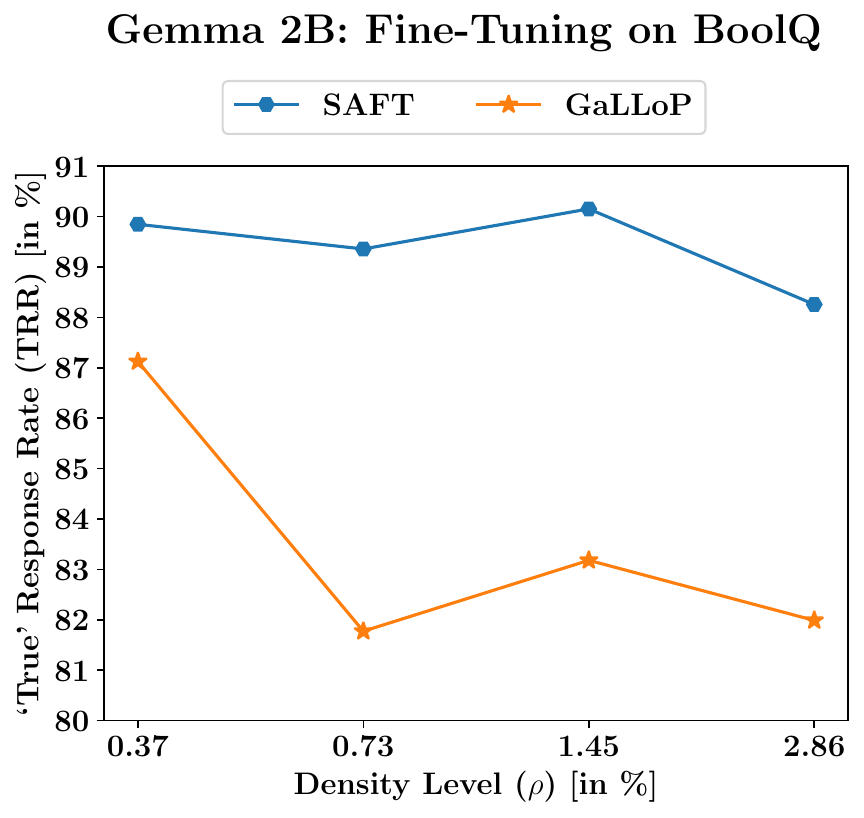}
\caption{Gemma 2B models fine-tuned with GaLLoP on the highly skewed BoolQ dataset increasingly learn what characterizes a `False' response while those fine-tuned with SAFT frequently generate `True' as the response potentially due to the memorization of the dominant correct response in BoolQ.}
\label{fig_boolq_trr}
\end{figure*}

Finally, for both LLaMA3 8B and Gemma 2B, an important case to note is that of fine-tuning on the BoolQ dataset. While the LLaMA3 8B (except for the highest density level) and Gemma 2B models fine-tuned with GaLLoP consistently outperform the models fine-tuned with SAFT on the ID (BoolQ) test set, the models fine-tuned with SAFT consistently outperform (except for the first two density levels for LLaMA3 8B) the models fine-tuned with GaLLoP on the corresponding OOD test sets (see~\autoref{fig_CR_llama_boolq} and~\autoref{fig_CR_gemma_boolq}). This might potentially be due to the heavily skewed distribution of the correct responses for BoolQ in its own test set (`True': 62.171\%; `False': 37.829\%) which is in consonance with the correct response distribution in its training set (`True': 62.304\%; `False': 37.686\%); `True' is roughly about twice as frequent as `False' for BoolQ. On analyzing the generated responses, we find that models fine-tuned with GaLLoP consistently try to learn what characterizes a `False' response whereas models fine-tuned with SAFT \emph{undesirably} generate `True' as the most frequent response, \emph{albeit incorrectly} (their ID accuracies are generally lower than those of the models fine-tuned with GaLLoP), possibly due to the near-perfect memorization of `True' as the correct response. For LLaMA3 8B, fine-tuning with GaLLoP leads to a huge drop in the prediction rate of `True' (from 66.391\% for $\rho$ = 0.93\% to 47.073\% for $\rho$ = 1.85\%) while fine-tuning with SAFT leads to a fairly constant and high prediction rate of `True' (80.764\% for $\rho$ = 0.93\% and 80.367\% for $\rho$ = 1.85\%). For Gemma 2B, such a trend can be seen across all density levels in~\autoref{fig_boolq_trr}. 

While memorizing and frequently generating `True' allows models fine-tuned with SAFT to easily attain a constant, moderate/moderately-low level OOD accuracy, learning what makes a response `False' seems to require harder effort on the part of models fine-tuned with GaLLoP and they increasingly start to respond with something different than the response format for the OOD tasks: either they generate the answer in words rather than the corresponding option number, or generate an explanation to the answer, or even generate the question itself, or etc. We leave the further examination of this phenomenon for future work.

\clearpage

\subsection{Catastrophic Forgetting}
\label{app:cfr}

\begin{figure*}[htbp]
    \begin{center}
        \subfigure[]{
            \label{fig_CRFR_gemma_boolq}
            \includegraphics[scale = 0.25]{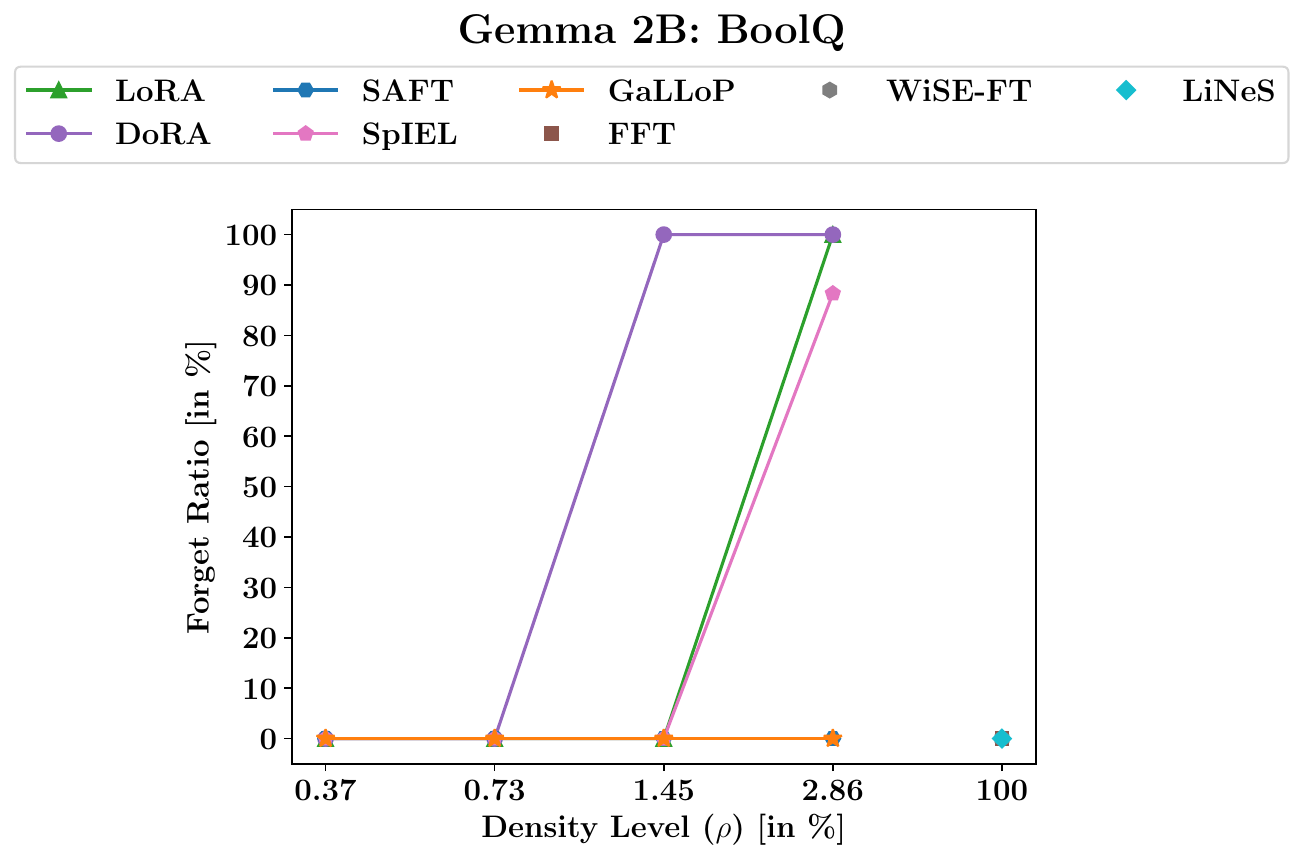}
        }
        \subfigure[]{
            \label{fig_CRFR_gemma_hellaswag}
            \includegraphics[scale = 0.25]{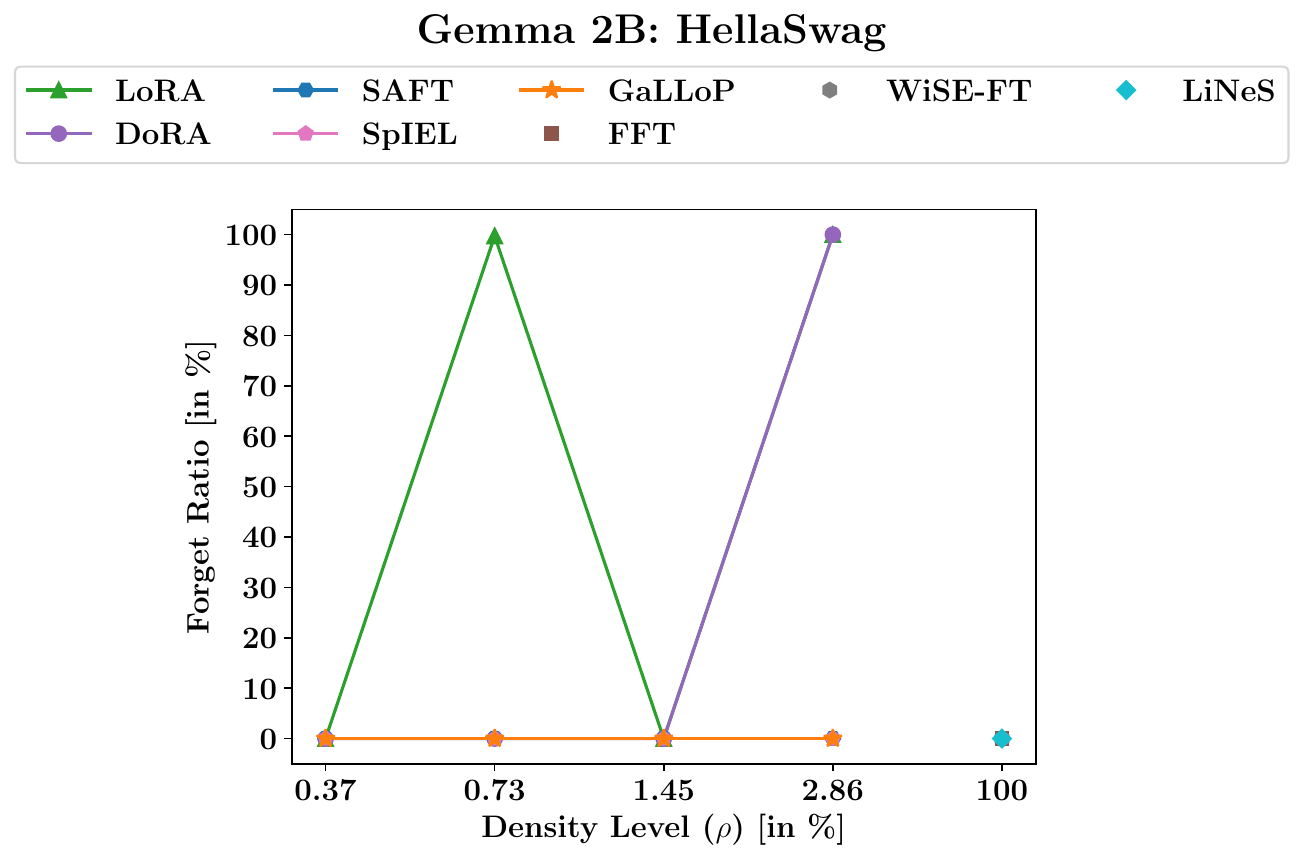}
        }
        \subfigure[]{
            \label{fig_CRFR_gemma_piqa}
            \includegraphics[scale = 0.25]{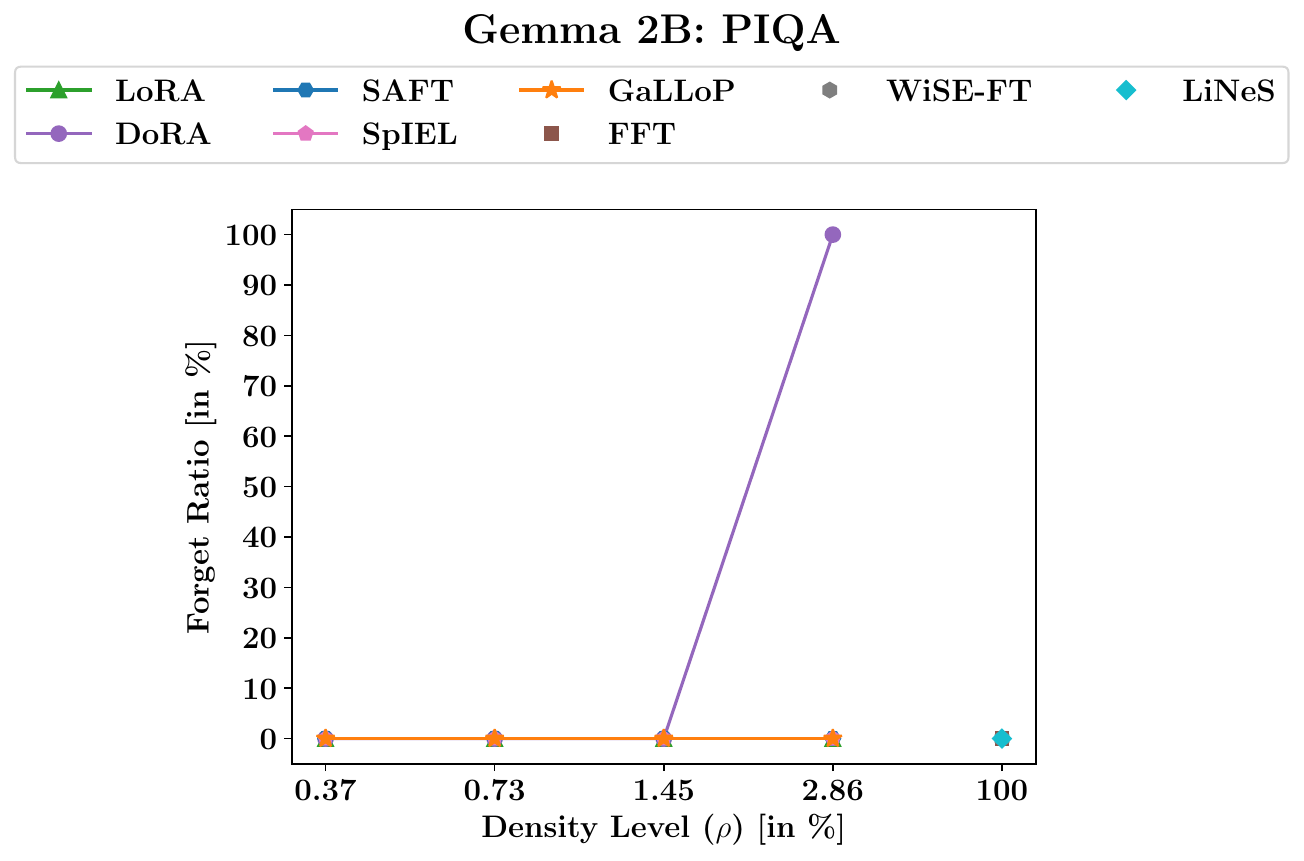}
        }
        \subfigure[]{
            \label{fig_CRFR_gemma_winogrande}
            \includegraphics[scale = 0.25]{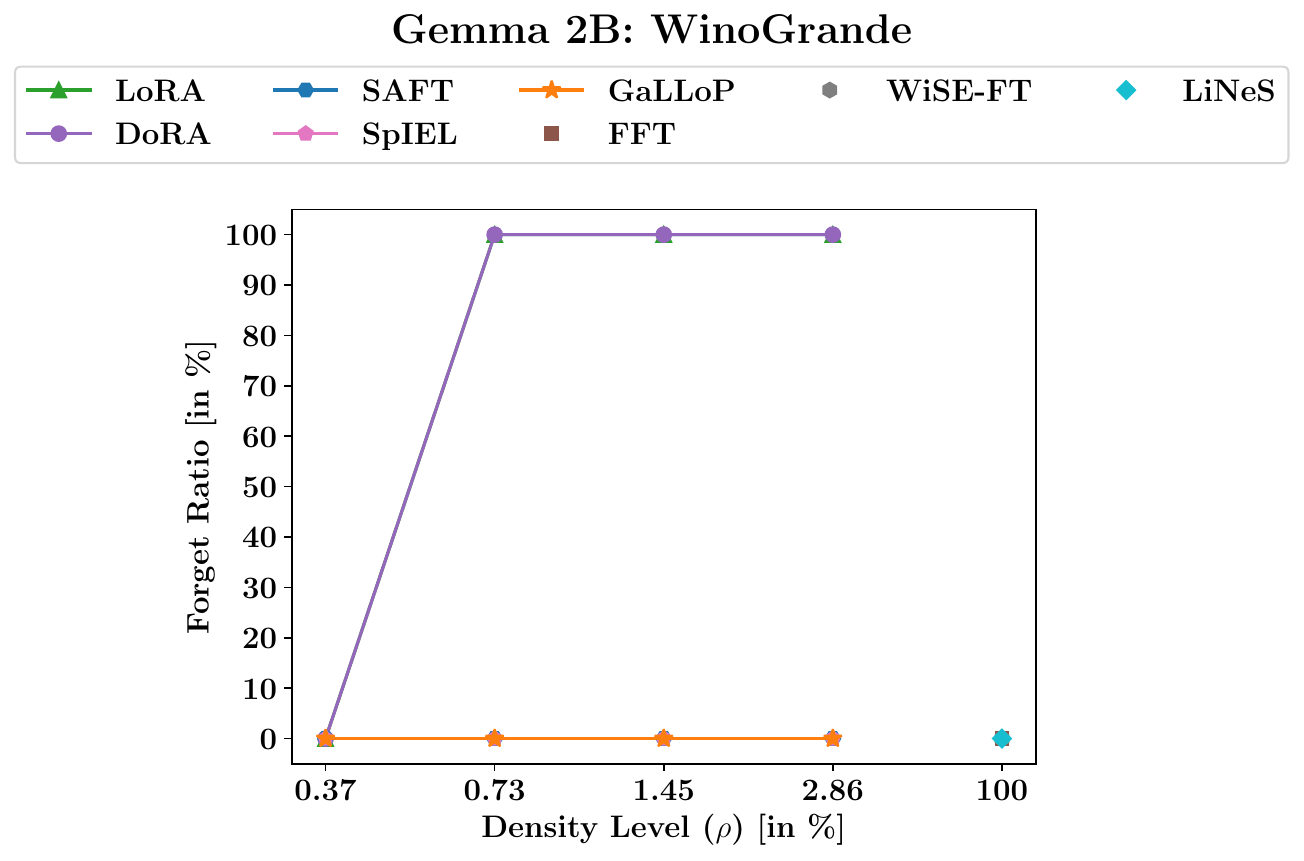}
        }
    \end{center}
    \caption{Gemma 2B models fine-tuned with GaLLoP show a 0\% forget ratio across all density levels when fine-tuning is performed on a) BoolQ, b) HellaSwag, c) PIQA, and d) WinoGrande.}
    \label{fig_CR_gemma_per_run_cfr}
\end{figure*}

\begin{figure*}[htbp]
    \begin{center}
        \subfigure[]{
            \label{fig_CRFR_llama_arc_c}
            \includegraphics[scale = 0.25]{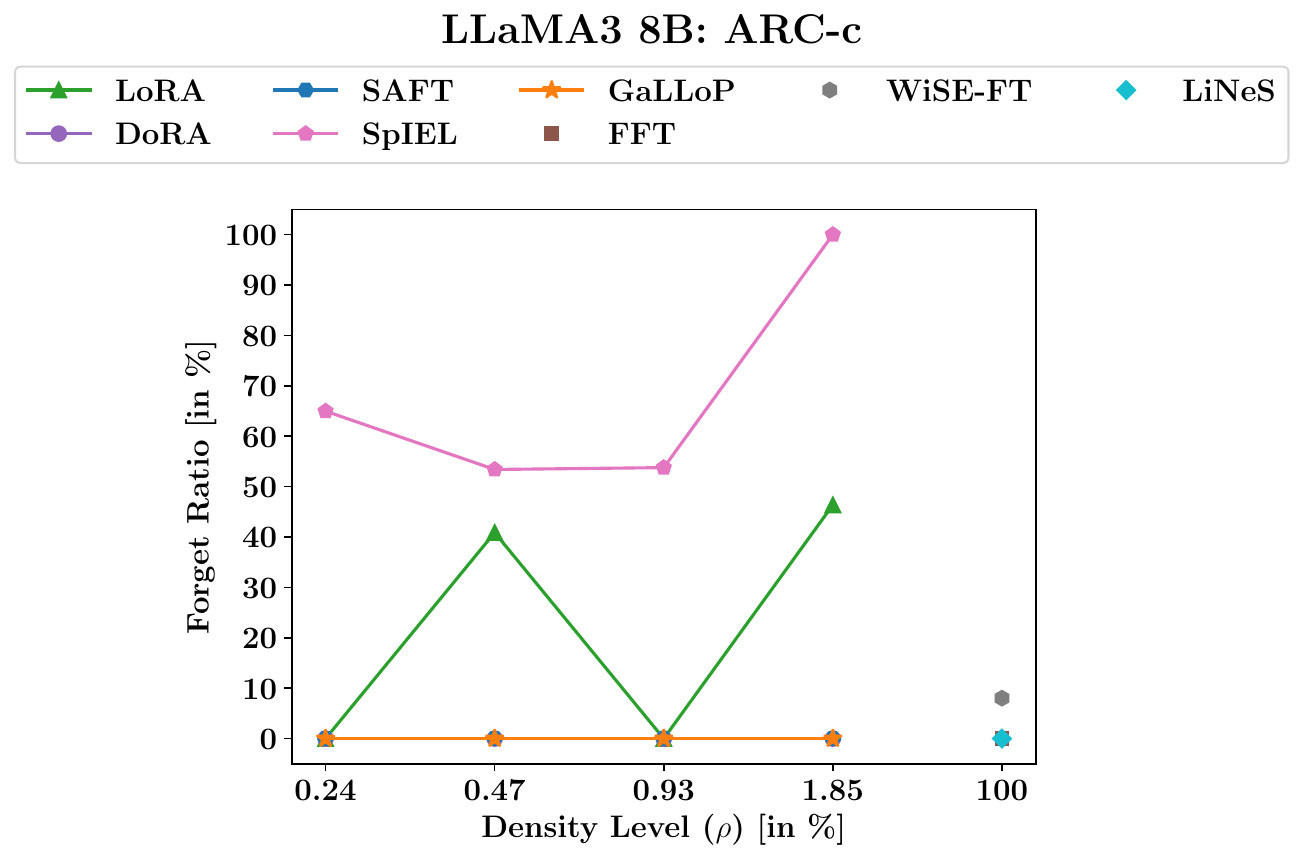}
        }
        \subfigure[]{
            \label{fig_CRFR_llama_arc_e}
            \includegraphics[scale = 0.25]{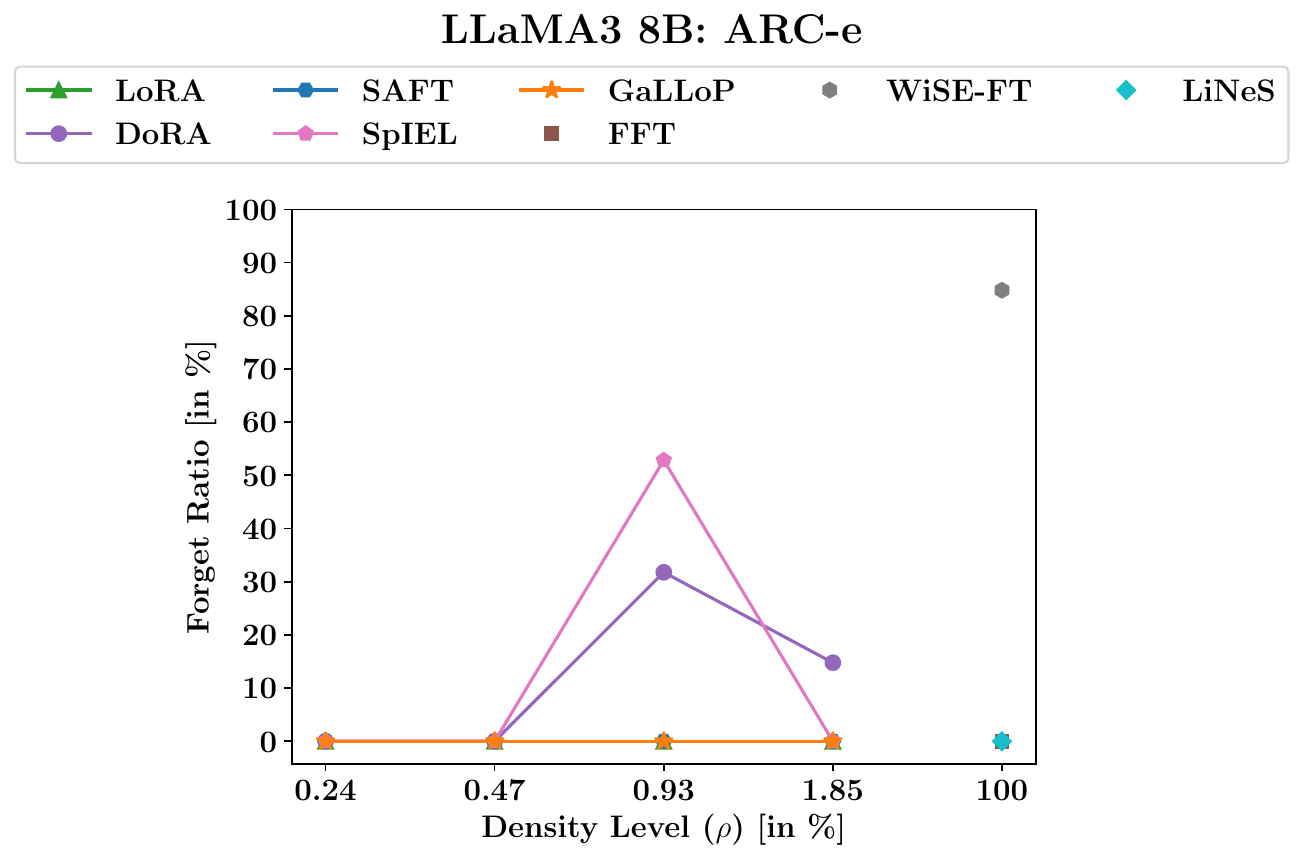}
        }
        \subfigure[]{
            \label{fig_CRFR_llama_hellaswag}
            \includegraphics[scale = 0.25]{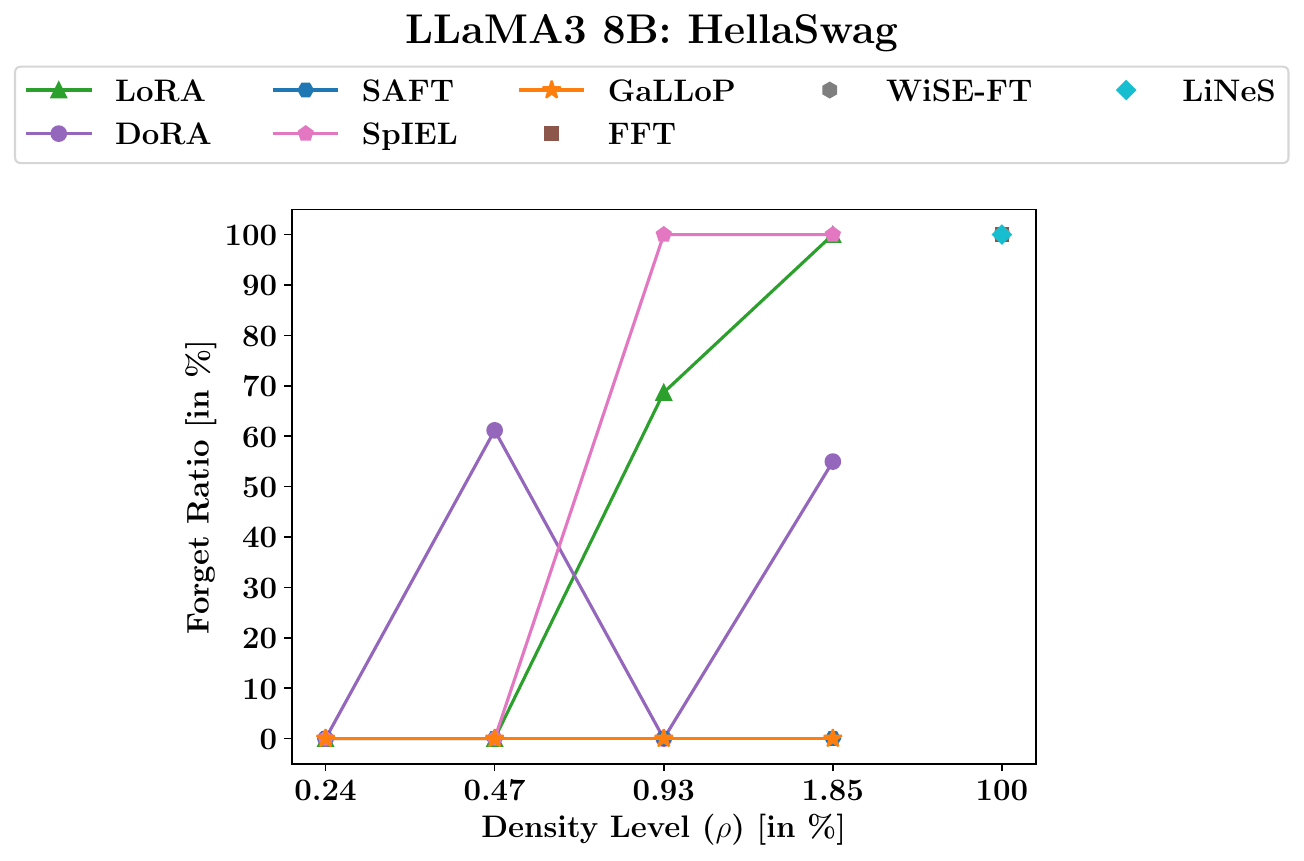}
        }
        \subfigure[]{
            \label{fig_CRFR_llama_obqa}
            \includegraphics[scale = 0.25]{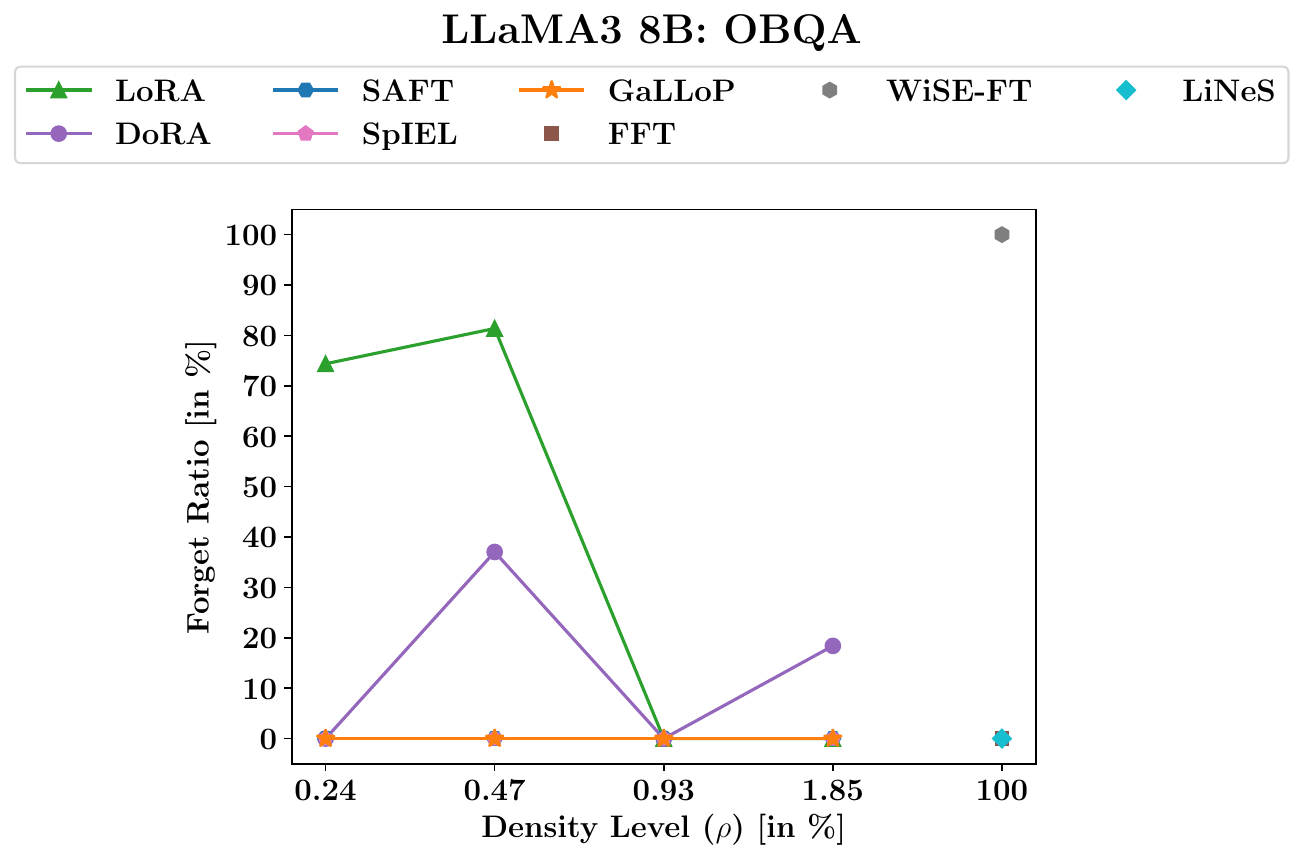}
        }
        \subfigure[]{
            \label{fig_CRFR_llama_piqa}
            \includegraphics[scale = 0.25]{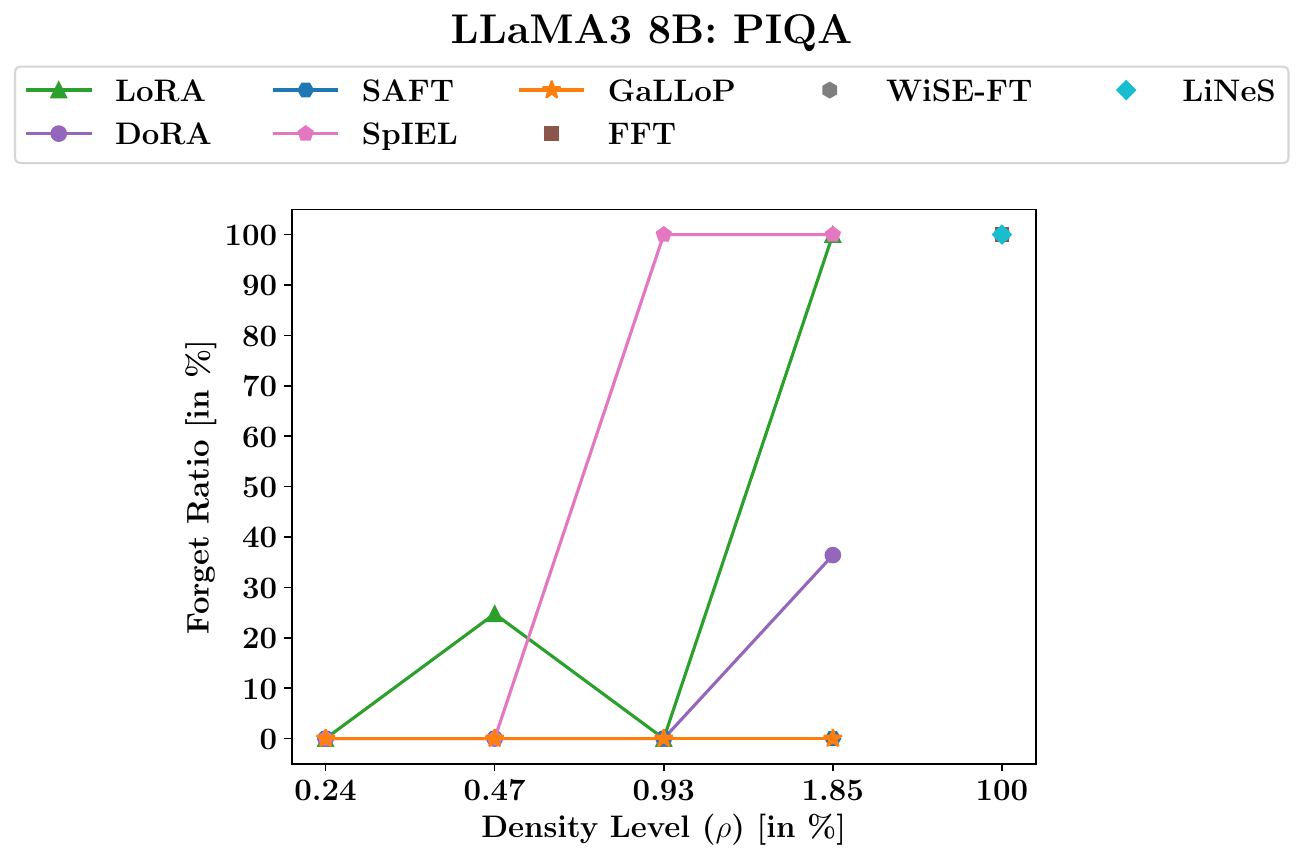}
        }
        \subfigure[]{
            \label{fig_CRFR_llama_winogrande}
            \includegraphics[scale = 0.25]{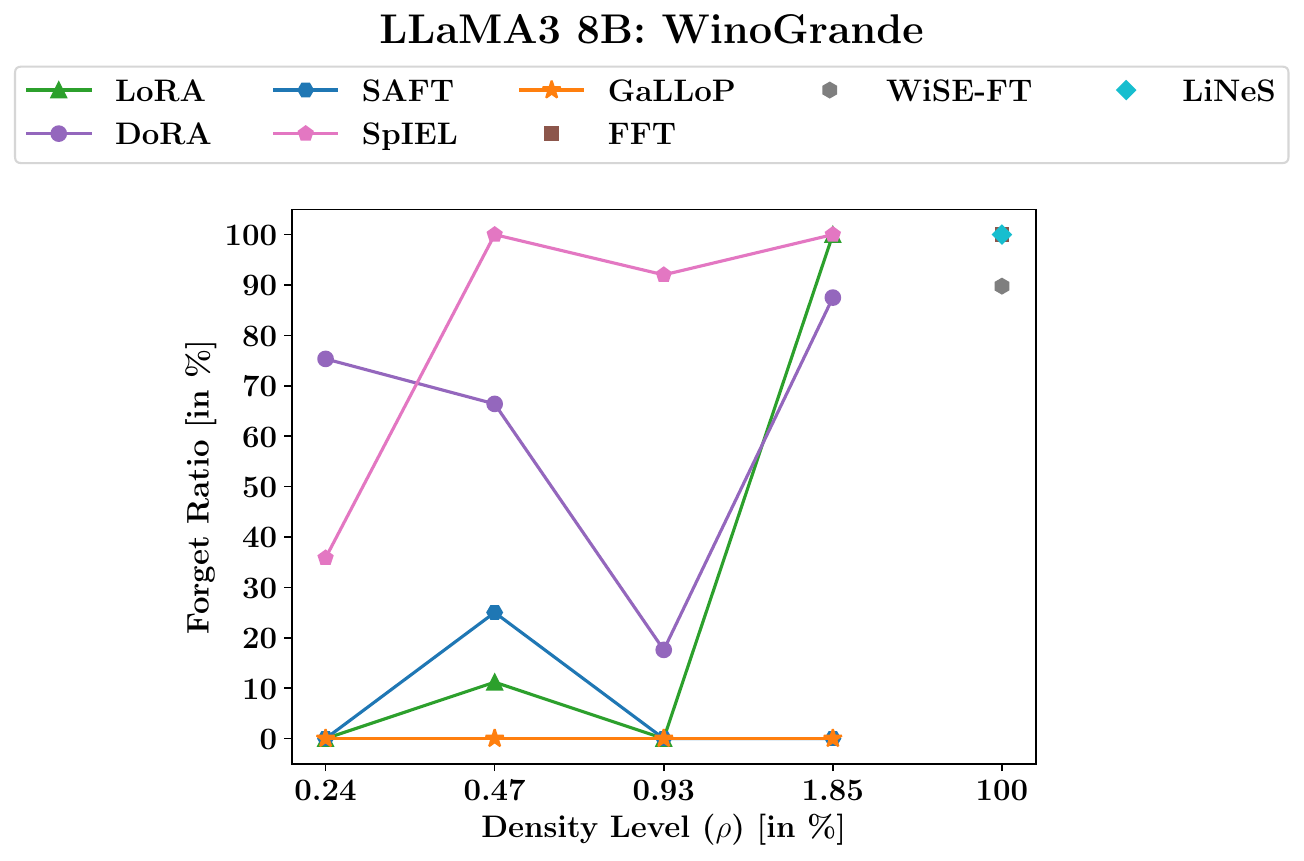}
        }
        \subfigure[]{
            \label{fig_CRFR_llama_siqa}
            \includegraphics[scale = 0.25]{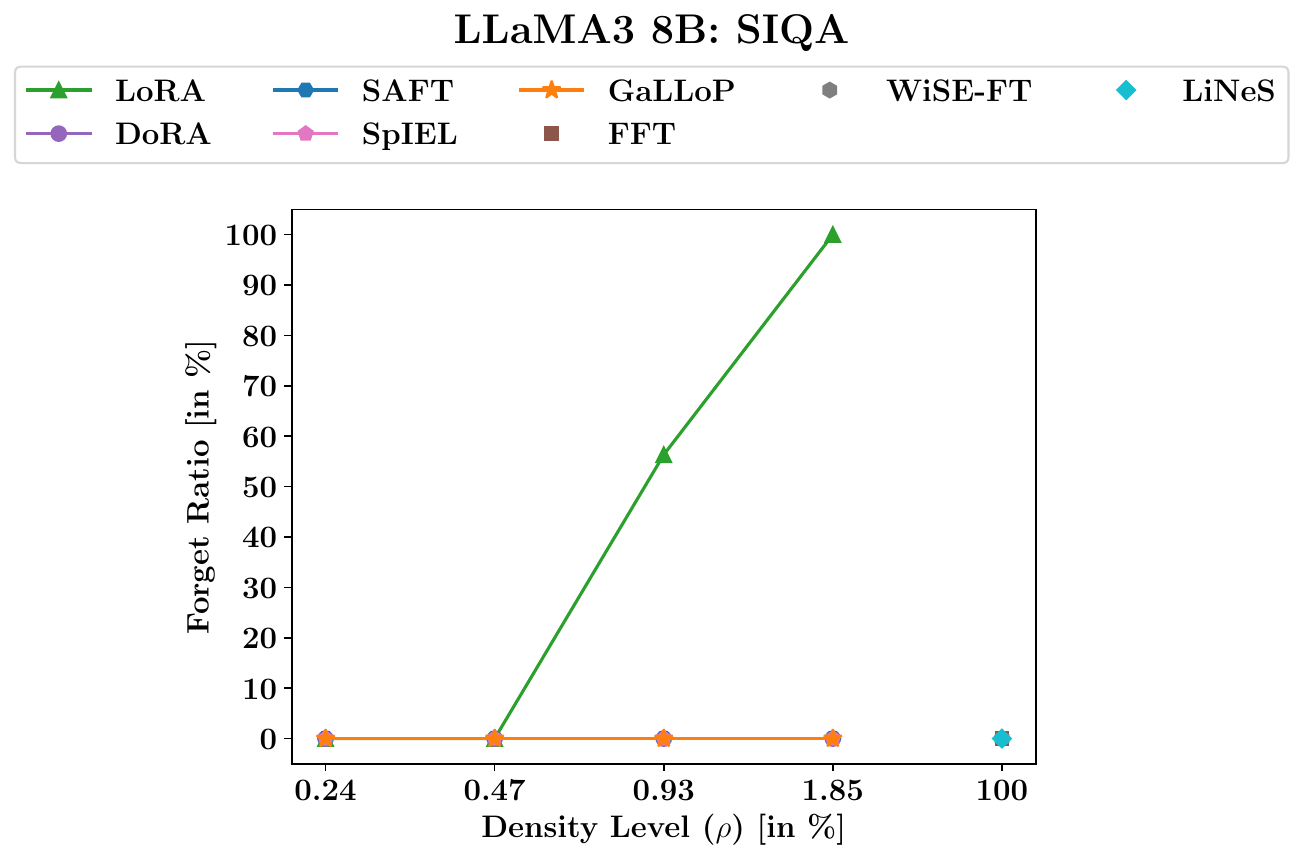}
        }
        \subfigure[]{
            \label{fig_CRFR_llama_boolq}
            \includegraphics[scale = 0.25]{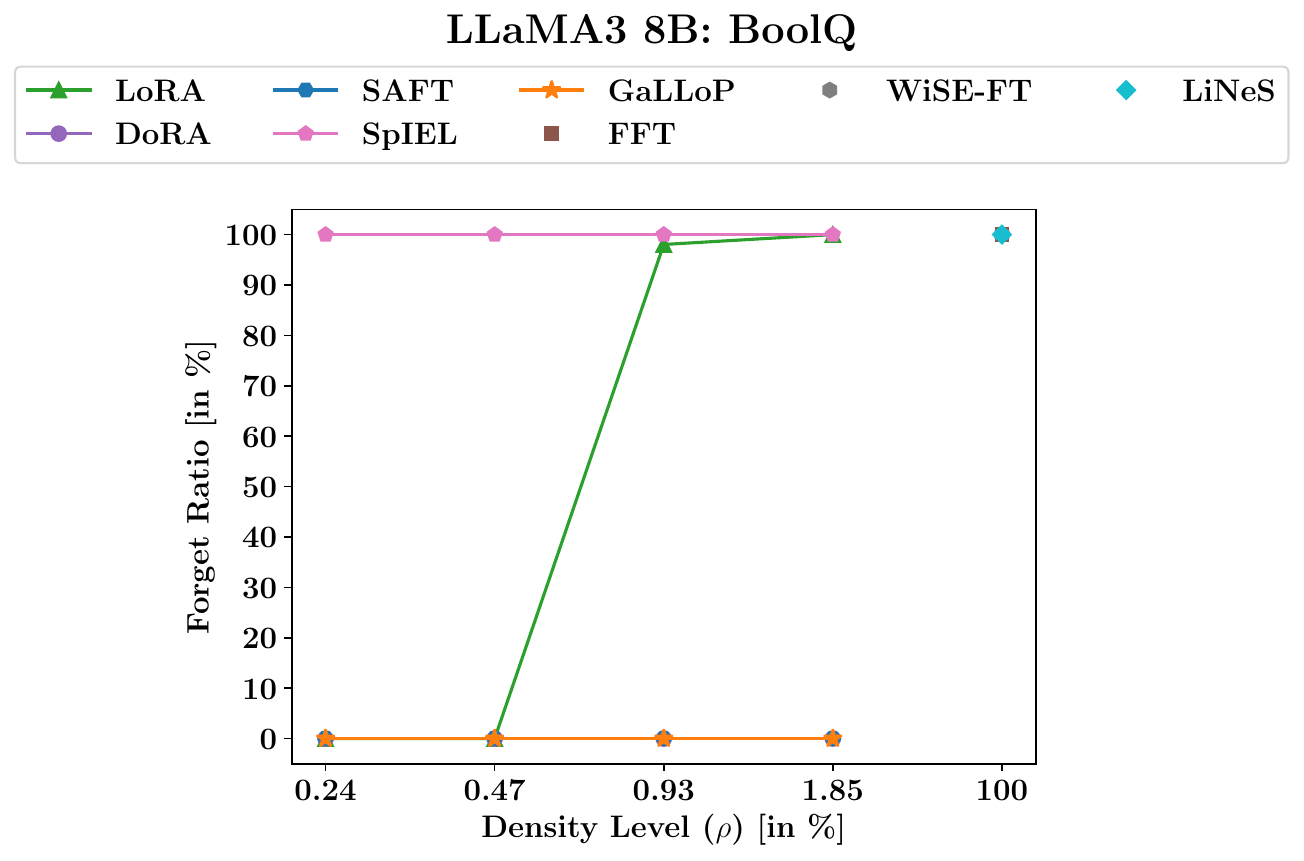}
        }
    \end{center}
    \caption{LLaMA3 8B models fine-tuned with GaLLoP show a 0\% forget ratio across all density levels when fine-tuning is performed on a) ARC-c, b) ARC-e, c) HellaSwag, d) OBQA, e) PIQA, f) WinoGrande, g) SIQA, and h) BoolQ.}
    \label{fig_CR_llama_per_run_cfr}
\end{figure*}

The per-run forget ratios are shown in~\autoref{fig_CR_llama_per_run_cfr} for fine-tuned and/or edited LLaMA3 8B models and in~\autoref{fig_CR_gemma_per_run_cfr} for fine-tuned and/or edited Gemma 2B models. 

The fact that fine-tuning LLaMA3 8B models with the runners-up algorithm SAFT results in instability with sudden and high rises/falls in ID and/or OOD performance on all datasets (except BoolQ; see~\autoref{fig_CR_llama_per_run}) and moderately-high catastrophic forgetting for $\rho = 0.47\%$ on the largest dataset (WinoGrande) (see~\autoref{fig_CRFR_llama_winogrande}) while fine-tuning them with GaLLoP does not --- reveals the benefits of GaLLoP's dual parameter selection criterion. Restricting the selection of high gradient parameters to those with the smallest pre-trained magnitudes leads to increased stability, relatively balanced ID as well as OOD performance, and prevents catastrophic forgetting across all datasets even in the overtrained regime.

Interestingly, upon fine-tuning Gemma 2B models individually on four datasets (ARC-c, ARC-e, OBQA, and SIQA), no catastrophic forgetting occurs (and hence those plots are not shown here). While this follows, \emph{in part}, from the explanation provided in~\autoref{sec:cfr}, why this specifically happens for only these four datasets is discussed in detail in the next appendix on memorization (\autoref{app:mem}).

Finally, as a follow-up to the discussion on the near-zero, vanilla (zero-shot) performance of Gemma 2B as compared to the relatively higher vanilla (zero-shot) performance of LLaMA3 8B (see~\autoref{sec:cfr}), we analyze the responses of the vanilla Gemma 2B and LLaMA3 8B models. Consequently, we observe that while the vanilla LLaMA3 8B model does respond with an answer/answers adhering to the response format (on most datasets), the vanilla Gemma 2B model, in most cases, does not respect the response format while responding or simply repeats the entire question or repeats all the answer choices verbatim from the question used for prompting it. A representative example of the responses generated by both the vanilla models is shown in~\autoref{vanilla_case}.

\begin{figure*}
\small
\begin{tcolorbox}[title=\textbf{Zero-Shot (Vanilla) Models}, halign title=flush center]
        
\textbf{Prompt (Example from PIQA):}\\Below is an instruction that describes a task. Write a response that appropriately completes the request.\\\\
\#\#\# Instruction:
\\
Please choose the correct solution to the question: How can I raise and lower a picavet with a camera attached so it can be flown with a kite?
\\\\
Solution1: You would implement a pulley system with wheels and strong string.
\\\\
Solution2: You would implement a pulley system with carabiners and strong string.
\\\\
Answer format: solution1/solution2
\\\\
\#\#\# Response:\\
\tcblower
\textbf{Reference Response:} the correct answer is solution2
\tcbsubtitle{\textbf{LLaMA3 8B}}
\textbf{Response:} solution1
\tcbsubtitle{\textbf{Gemma 2B}}
\textbf{Response:} I would implement a pulley system with wheels and strong string.\\\\I would implement a pulley system with carabiners and strong string.

\end{tcolorbox}
\caption{The vanilla LLaMA3 8B model generates an answer when prompted with a question from the PIQA dataset while the vanilla Gemma 2B model generates both answers without respecting the answer format when prompted with the same question.}
\label{vanilla_case}
\end{figure*}

\clearpage

\section{Memorization}
\label{app:mem}

Collapse rates and forget ratios do not always reveal the same information; rather, they actually reveal \emph{complementary} information. For instance, even though the LLaMA3 8B models fine-tuned with DoRA possess high forget ratios varying between 10-20\% (see~\autoref{fig_CR_cfr_llama}), they do not undergo extremely high memorization and hence, showcase low collapse rates (see ~\autoref{fig_CR_fr_llama}). Additionally, while Gemma 2B models fine-tuned with SpIEL (for the first three density levels) and FFT possess 0\% forget ratios (see~\autoref{fig_CR_cfr_gemma}), they do undergo memorization (see~\autoref{fig_CR_fr_gemma}). Decrease in the extent of memorization on fine-tuning LLaMA3 8B models with DoRA as compared to fine-tuning them with LoRA is likely due to the former's fine-grained (magnitude and directional) reparametrized update of the overtrained parameter matrix which imparts higher learning capabilities than the latter. Nevertheless, since DoRA is still an RFT technique, all the trainable parameters introduced by it are fine-tuned on the training set which then leads to catastrophic forgetting and moderate levels of memorization. 0\% forget ratios versus high collapse rates of Gemma 2B models fine-tuned with SpIEL and FFT follows from our previous discussion on the performance of the vanilla Gemma 2B model (see~\autoref{sec:cfr}). Near-zero performance of the vanilla Gemma 2B model along with the near-zero performance of the same model fine-tuned with SpIEL and FFT yields a 0\% forget ratio in line with~\autoref{per_run_cfr}. However, an analysis of the responses generated by both of them (vanilla and fine-tuned) reveal striking differences between them with the latter consistently responding with an EOS token (as discussed in~\autoref{sec:mem}).

When models memorize the response formats of the dataset used for fine-tuning (see~\autoref{tab:datasets} in~\autoref{AppendixA} for response formats), it means that while a model fine-tuned on the training set of any one of the following datasets: ARC-c, ARC-e, OBQA, and SIQA (which share the same response format), is increasingly likely to perform well on the test sets of the remaining three OOD datasets (note that this is also the reason why forget ratios equal zero on these datasets for the fine-tuned Gemma 2B models: see~\autoref{app:cfr}), models fine-tuned on any one of the following datasets: BoolQ, HellaSwag, PIQA, and WinoGrande (which possess response formats completely orthogonal to the other datasets), increasingly fail on the test sets of all the corresponding seven OOD datasets. 

While repetition of the most frequently occurring words/phrases in the fine-tuning dataset is partly due to the maximum likelihood-based fine-tuning objective~\citep{likelihood_max_causes_repetition}, and might also arise due to an unfortunate set of parameters selected for combined deterministic beam-search~\citep{beam_search} and random (top-$k$~\citep{top_k} and top-$p$~\citep{top_p}) sampling~\citep{beam_search_top_p_combined, decoding_inconsistency, top_p_repetitive} which makes the repetition-encouraging nature dominant during decoding (this is solely because of the parameter values defined by~\cite{llm-adapters} and enlisted in~\autoref{tab:decoding_hyperparameters} in~\autoref{AppendixC}), its occurrence seems to also depend upon the type of fine-tuning employed since it only occurs on performing RFT and not SpFT. Following the discussion in~\autoref{sec:mem}, \emph{concerted overfitting}, along with the detrimental effects of maximum likelihood-based fine-tuning, only gets exacerberated in the overtrained regime~\citep{overtraining} which is why this is only observed for LLaMA3 8B (and not Gemma 2B) models. 

While the generation of the EOS token might partly be due to the dominant effect of beam search (over random sampling) in the overtrained regime which consequently might cause high probablities to be assigned to the shortest hypotheses with just the EOS token~\citep{beam_search_eos}, the fact that it only occurs upon fine-tuning LLaMA3 8B models with SpIEL and FFT must also be noted and hence, we attribute it to the tendency of these algorithms to cause instability in the learning dynamics in the overtrained regime.

Representative samples of responses generated by such models, showcasing all the aforementioned forms of memorization, against those generated by models fine-tuned with GaLLoP and SAFT can be seen in~\autoref{resp3},~\autoref{resp4},~\autoref{resp1}, and~\autoref{resp2}.

\begin{figure*}[hbp]
\small
\begin{tcolorbox}[title=\textbf{Gemma 2B Models Fine-Tuned on WinoGrande with {$\rho=0.73\%$}}, halign title=flush center]
\textbf{Prompt (Example from ARC-e):}\\Below is an instruction that describes a task. Write a response that appropriately completes the request.\\\\
\#\#\# Instruction:
\\
Please choose the correct answer to the question: A student throws a ball into the air. While the ball travels up, the speed of the ball decreases. What force causes the ball to slow while traveling up?
\\\\
Answer1: electricity Answer2: gravity Answer3: magnetism Answer4: tension
\\\\
Answer format: answer1/answer2/answer3/answer4
\\\\
\#\#\# Response:\\
\tcblower
\textbf{Reference Response:} the correct answer is answer2
\tcbsubtitle{\textbf{LoRA}}
\textbf{Response:} the correct answer is option1
\tcbsubtitle{\textbf{DoRA}}
\textbf{Response:} the correct answer is option1
\tcbsubtitle{\textbf{SAFT}}
\textbf{Response:} the correct answer is `answer2'
\tcbsubtitle{\textbf{SpIEL}}
\textbf{Response:} the correct answer is answer2
\tcbsubtitle{\textbf{GaLLoP}}
\textbf{Response:} the correct answer is answer2
\tcbsubtitle{\textbf{FFT}}
\textbf{Response:} the correct answer is answer2
\tcbsubtitle{\textbf{WiSE-FT}}
\textbf{Response:} the correct answer is answer2
\tcbsubtitle{\textbf{LiNeS}}
\textbf{Response:} the correct answer is answer2

\end{tcolorbox}
\caption{Gemma 2B models fine-tuned on WinoGrande with GaLLoP, SAFT, SpIEL, FFT and/or edited with WiSE-FT and LiNeS respect ARC-e's response format whereas those fine-tuned with LoRA and DoRA do not.}
\label{resp3}
\end{figure*}

\begin{figure*}
\small
\begin{tcolorbox}[title=\textbf{Gemma 2B Models Fine-Tuned on SIQA with {$\rho=1.45\%$}}, halign title=flush center]
\textbf{Prompt (Example from HellaSwag):}\\Below is an instruction that describes a task. Write a response that appropriately completes the request.\\\\
\#\#\# Instruction:
\\
Please choose the correct ending to complete the given sentence: Having an ice cream: Children bring desert out for their family member. the family
\\\\
Ending1: floats in a river. Ending2: member stands looking into a hut and then handing people photographs. Ending3: member cuts a piece of sunscreen. Ending4: enjoys eating the desert together.
\\\\
Answer format: ending1/ending2/ending3/ending4
\\\\
\#\#\# Response:\\
\tcblower
\textbf{Reference Response:} the correct answer is ending4
\tcbsubtitle{\textbf{LoRA}}
\textbf{Response:} the correct answer is answer4
\tcbsubtitle{\textbf{DoRA}}
\textbf{Response:} the correct answer is answer4
\tcbsubtitle{\textbf{SAFT}}
\textbf{Response:} the correct answer is ending4
\tcbsubtitle{\textbf{SpIEL}}
\textbf{Response:} the correct answer is answer4
\tcbsubtitle{\textbf{GaLLoP}}
\textbf{Response:} the correct answer is ending4
\tcbsubtitle{\textbf{FFT}}
\textbf{Response:} the correct answer is answer3
\tcbsubtitle{\textbf{WiSE-FT}}
\textbf{Response:} the correct answer is answer4
\tcbsubtitle{\textbf{LiNeS}}
\textbf{Response:} the correct answer is answer4

\end{tcolorbox}
\caption{Gemma 2B models fine-tuned on SIQA with GaLLoP and SAFT respect HellaSwag's response format whereas those fine-tuned and/or edited with the other competing algorithms do not.}
\label{resp4}
\end{figure*}

\begin{figure*}
\small
\begin{tcolorbox}[title=\textbf{LLaMA3 8B Models Fine-Tuned on ARC-c with {$\rho=1.85\%$}}, halign title=flush center]
\textbf{Prompt (Example from OBQA):}\\Below is an instruction that describes a task. Write a response that appropriately completes the request.\\\\
\#\#\# Instruction:
\\
Please choose the correct answer to the question: As the rain forest is deforested the atmosphere will increase with
\\\\
Answer1: oxygen Answer2: nitrogen Answer3: carbon Answer4: rain
\\\\
Answer format: answer1/answer2/answer3/answer4
\\\\
\#\#\# Response:\\
\tcblower
\textbf{Reference Response:} the correct answer is answer3
\tcbsubtitle{\textbf{LoRA}}
\textbf{Response:} the correct answer is answer3assistantassistant\\\\the correct answer is answer3\\\\assistant\\\\the correct answer is answer
\tcbsubtitle{\textbf{DoRA}}
\textbf{Response:} the correct answer is answer3assistant the correct answer is answer3assistant the correct answer is
\tcbsubtitle{\textbf{SAFT}}
\textbf{Response:} the correct answer is answer3
\tcbsubtitle{\textbf{SpIEL}}
\textbf{Response:} $<\text{EOS}>$
\tcbsubtitle{\textbf{GaLLoP}}
\textbf{Response:} the correct answer is answer3
\tcbsubtitle{\textbf{FFT}}
\textbf{Response:} the correct answer is answer3
\tcbsubtitle{\textbf{WiSE-FT}}
\textbf{Response:} the correct answer is answer4
\tcbsubtitle{\textbf{LiNeS}}
\textbf{Response:} the correct answer is answer3

\end{tcolorbox}
\caption{LLaMA3 8B models fine-tuned on ARC-c with GaLLoP, SAFT, FFT and/or edited with WiSE-FT and LiNeS respect OBQA's response format whereas those fine-tuned with LoRA, DoRA, and SpIEL either generate repetitive text or do not respond (except with the EOS token; $<\text{EOS}>$).}
\label{resp1}
\end{figure*}

\begin{figure*}
\small
\begin{tcolorbox}[title=\textbf{LLaMA3 8B Models Fine-Tuned on PIQA with {$\rho=1.85\%$}}, halign title=flush center]
\textbf{Prompt (Example from BoolQ):}\\Below is an instruction that describes a task. Write a response that appropriately completes the request.\\\\
\#\#\# Instruction:
\\
Please answer the following question with true or false, question: is northern ireland part of the great britain?
\\\\
Answer format: true/false
\\\\
\#\#\# Response:\\
\tcblower
\textbf{Reference Response:} the correct answer is false
\tcbsubtitle{\textbf{LoRA}}
\textbf{Response:} the correct answer is solution2assistant\\\\the correct answer is solution1assistant\\\\the correct answer is solution2assistant\\\\the correct
\tcbsubtitle{\textbf{DoRA}}
\textbf{Response:} the correct answer is trueassistant\\\\the correct answer is trueassistant\\\\the correct answer is true
\tcbsubtitle{\textbf{SAFT}}
\textbf{Response:} the correct answer is true, northern ireland is part of the great britain.\\\\the correct answer is true, northern ireland is part of the great
\tcbsubtitle{\textbf{SpIEL}}
\textbf{Response:} the correct answer is solution
\tcbsubtitle{\textbf{GaLLoP}}
\textbf{Response:} the correct answer is false
\tcbsubtitle{\textbf{FFT}}
\textbf{Response:} $<\text{EOS}>$
\tcbsubtitle{\textbf{WiSE-FT}}
\textbf{Response:} $<\text{EOS}>$
\tcbsubtitle{\textbf{LiNeS}}
\textbf{Response:} $<\text{EOS}>$

\end{tcolorbox}
\caption{LLaMA3 8B models fine-tuned on PIQA with GaLLoP and SAFT respect BoolQ's response format whereas those fine-tuned and/or edited with the other competing algorithms either do not respect the response format and/or generate repetitive text or do not respond (except with the EOS token; $<\text{EOS}>$).}
\label{resp2}
\end{figure*}

\clearpage

\section{Variability in Forget Ratios and Collapse Rates}
\label{app:var_cfr_mem}

\begin{figure*}[htbp]
    \begin{center}
        \subfigure[]{
            \label{fig_cfr_variance}
            \includegraphics[scale = 0.30]{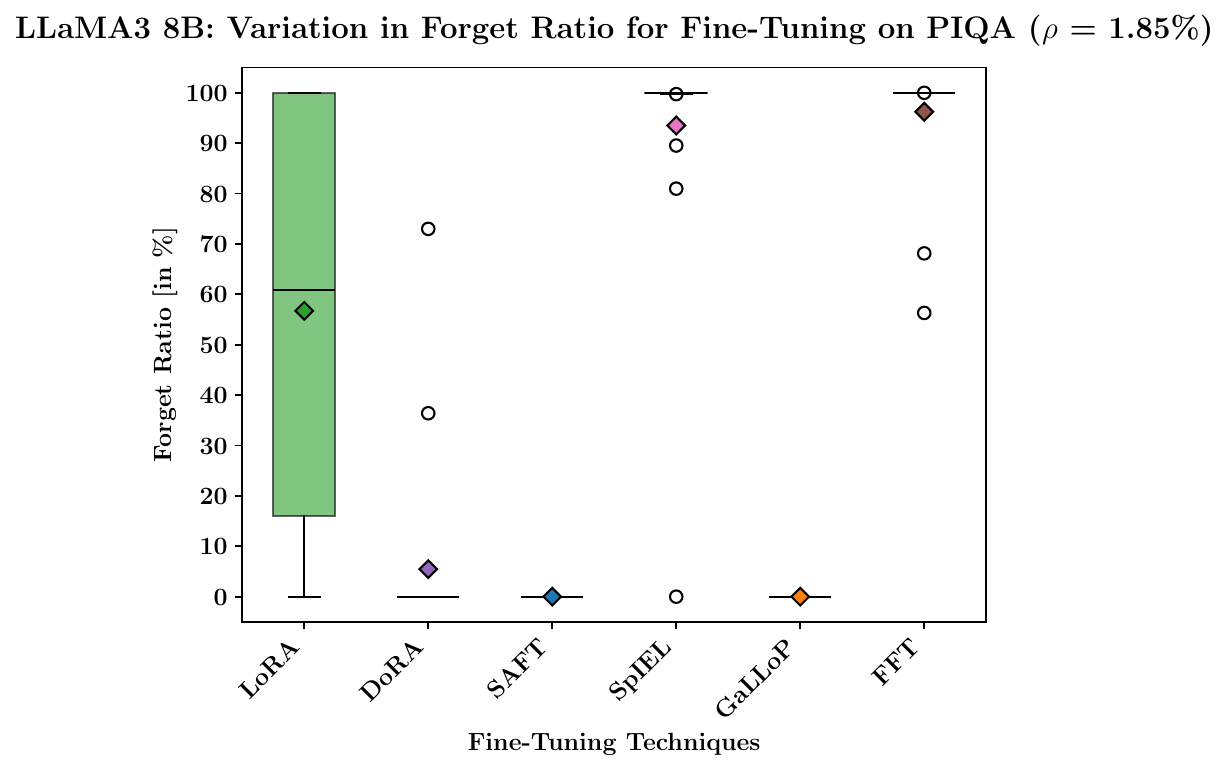}
        }
        \subfigure[]{
            \label{fig_fr_variance}
            \includegraphics[scale = 0.30]{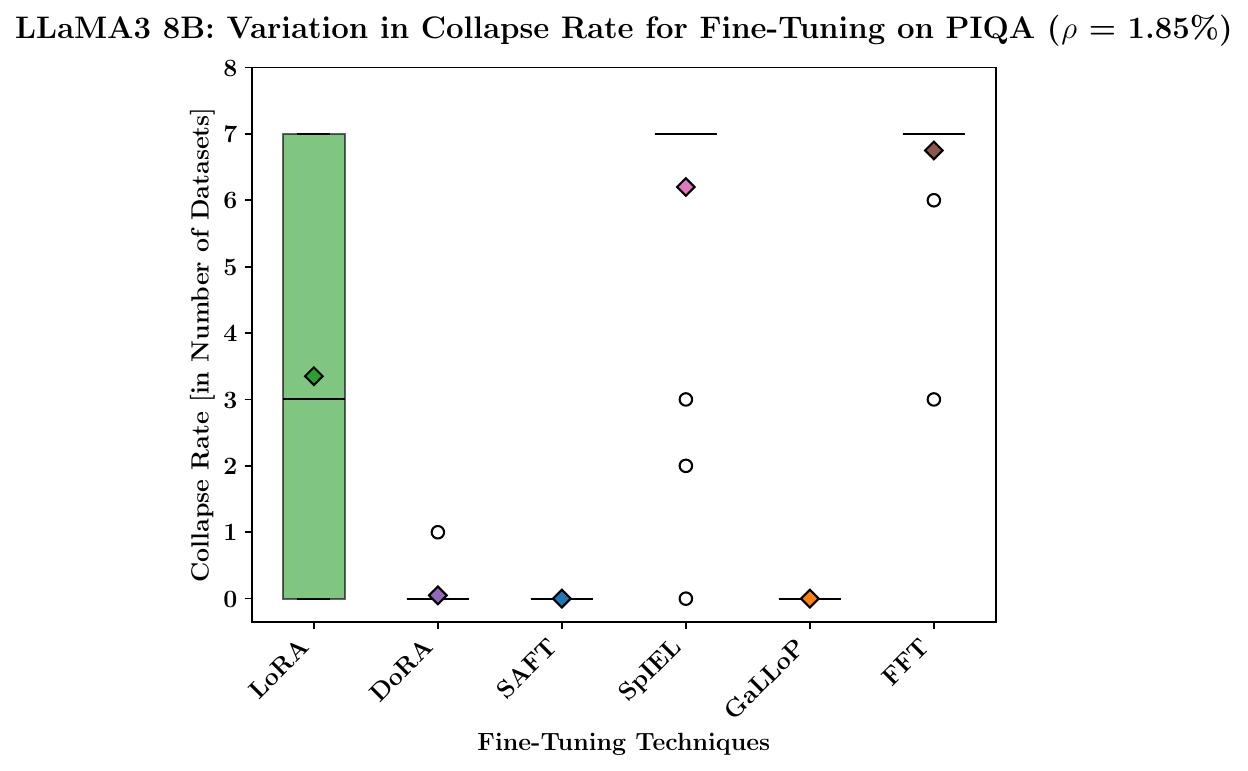}
        }
    \end{center}
    \caption{LLaMA3 8B models fine-tuned with GaLLoP consistently attain a) 0\% forget ratios and b) 0\% collapse rates across 20 different random seeds upon fine-tuning on the PIQA dataset with the highest density level ($\rho = 1.85\%$).}
    \label{fig_variance}
\end{figure*}

\autoref{fig_cfr_variance} and~\autoref{fig_fr_variance} show the boxplots of forget ratios and collapse rates (respectively) for LLaMA3 8B models fine-tuned with GaLLoP against models fine-tuned with the competing algorithms across 20 different random seeds. 

Models fine-tuned with GaLLoP and SAFT consistently attain 0\% forget ratios and 0\% collapse rates. In contrast, models fine-tuned with SpIEL consistently attain high forget ratios (80 - 100\%) and high collapse rates mostly equivalent to a complete failure on all the seven OOD datasets. The extent of variability in their forget ratios and collapse rates is also quite high and never allows them to perform on par/even close to models fine-tuned with GaLLoP on OOD tasks. These observations can possibly be owed to SpIEL's dynamic parameter selection: variability in random seeds can lead to changes in the candidate parameter set and hence, different sets of parameters being considered for updates at each reselection stage. Models fine-tuned with the RFT techniques also possess high forget ratios and high collapse rates (LoRA and FFT) and/or with high variance in their values (LoRA, DoRA, and FFT) owing to an absence of regularization during fine-tuning.

\section{Ablations on Parameter Selection Criteria}
\label{Appendix_abl}

\begin{figure*}[htbp]
\centering
\includegraphics[scale=0.30]{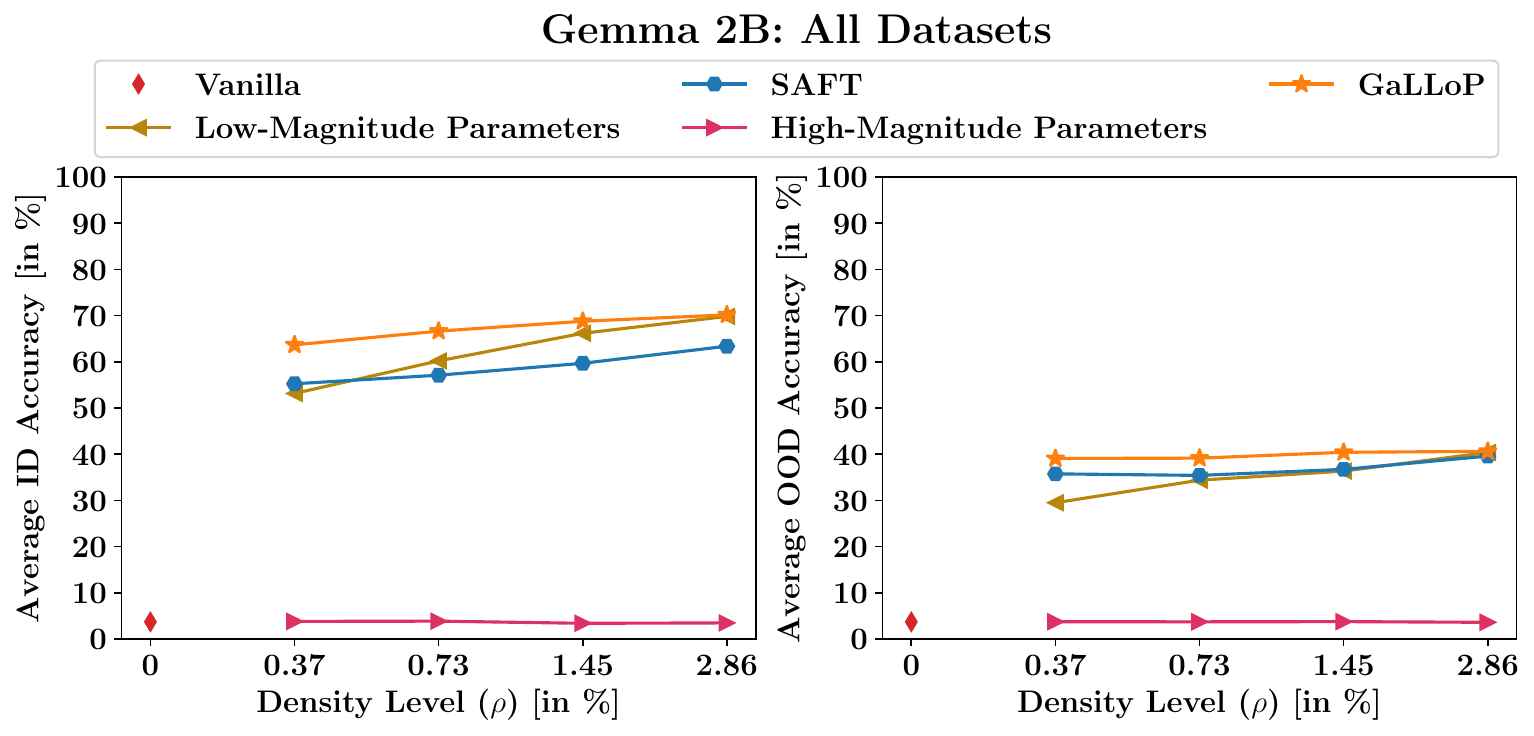}
\caption{Gemma 2B models fine-tuned with GaLLoP form a dominant Pareto front for both ID and OOD performance (on average) over models fine-tuned by selecting parameters on the basis of either parameter or gradient magnitudes.}
\label{fig_ablations_full_avg}
\end{figure*}

\begin{figure*}[htbp]
    \begin{center}
        \subfigure[]{
            \label{fig_CR_gemma_ablations_arc_c}
            \includegraphics[scale = 0.25]{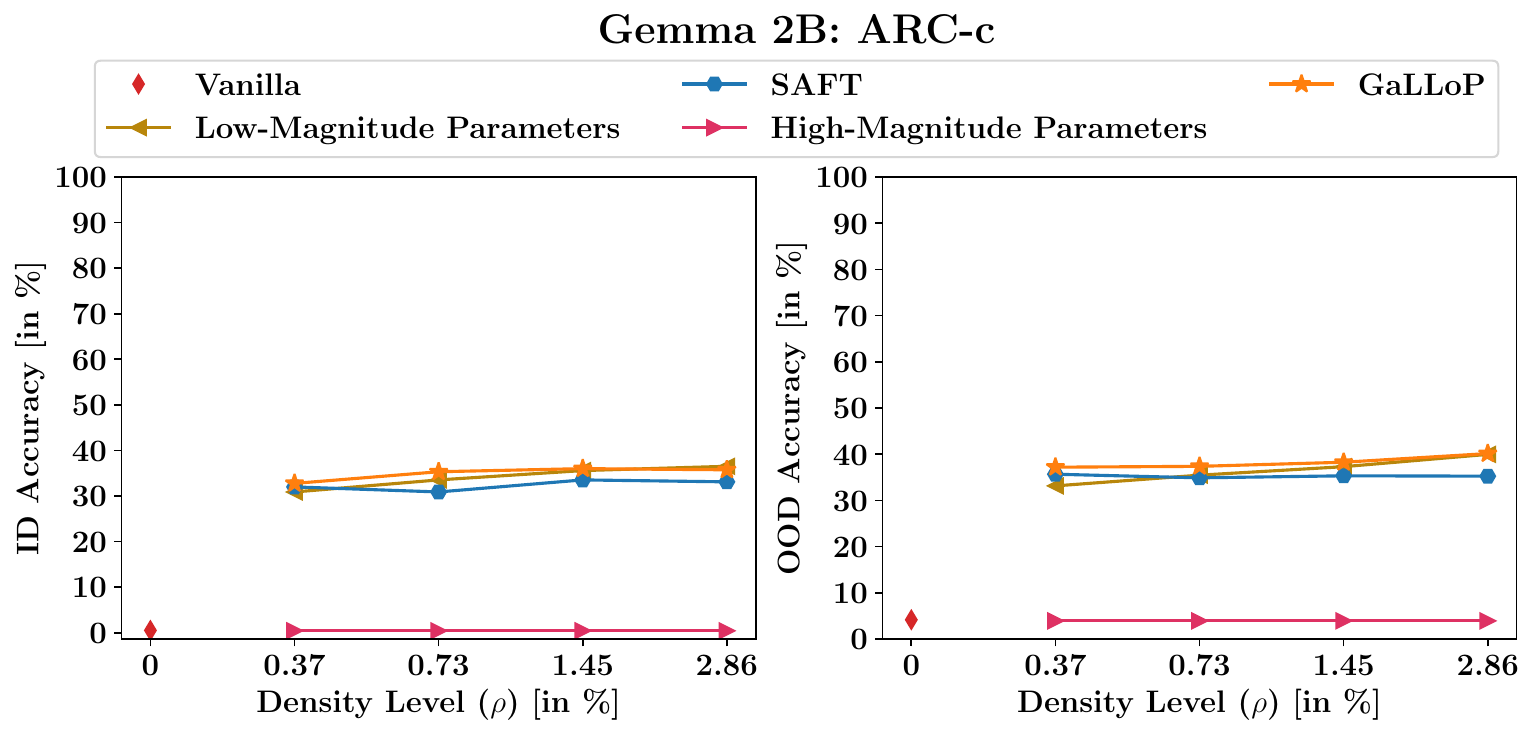}
        }
        \subfigure[]{
            \label{fig_CR_gemma_ablations_arc_e}
            \includegraphics[scale = 0.25]{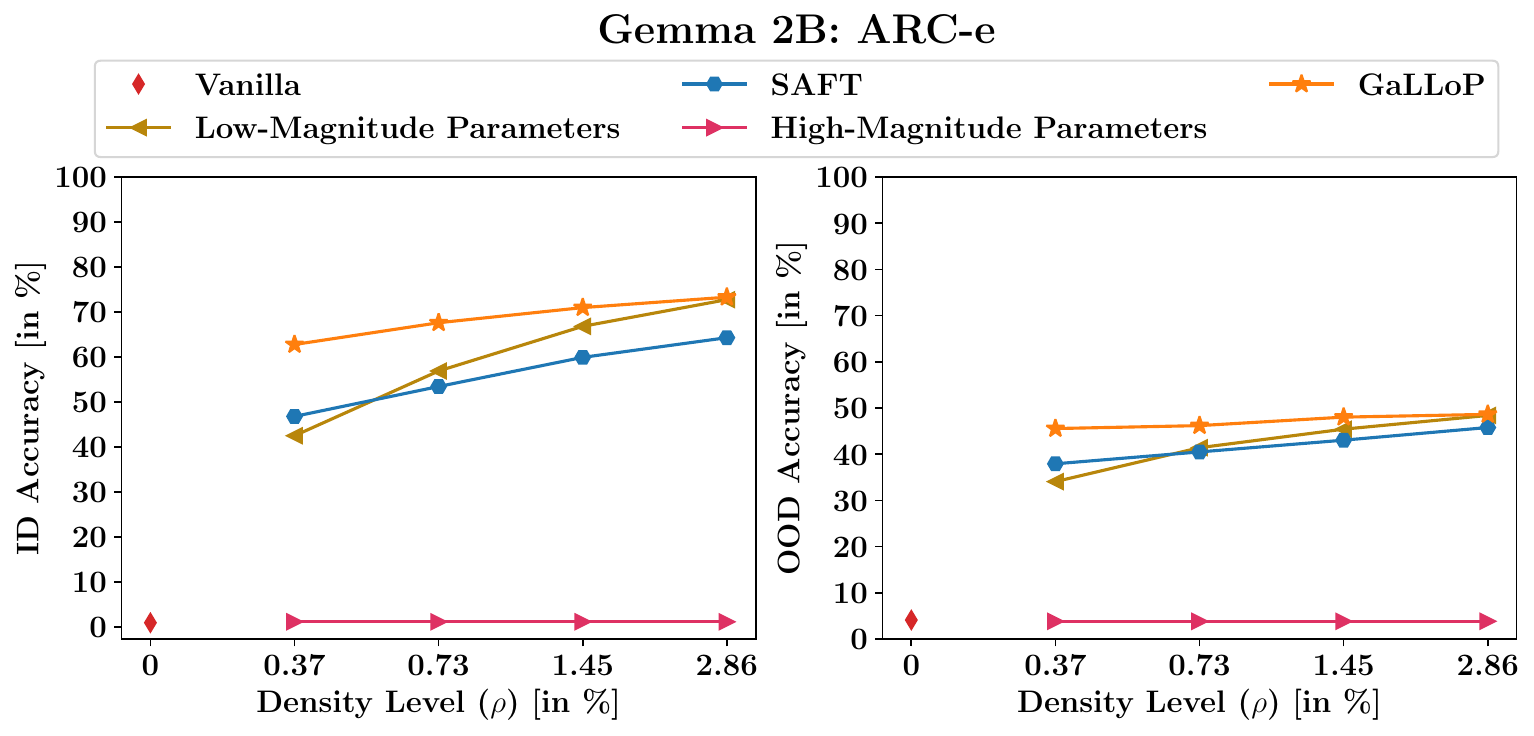}
        }
        \subfigure[]{
            \label{fig_CR_gemma_ablations_piqa}
            \includegraphics[scale = 0.25]{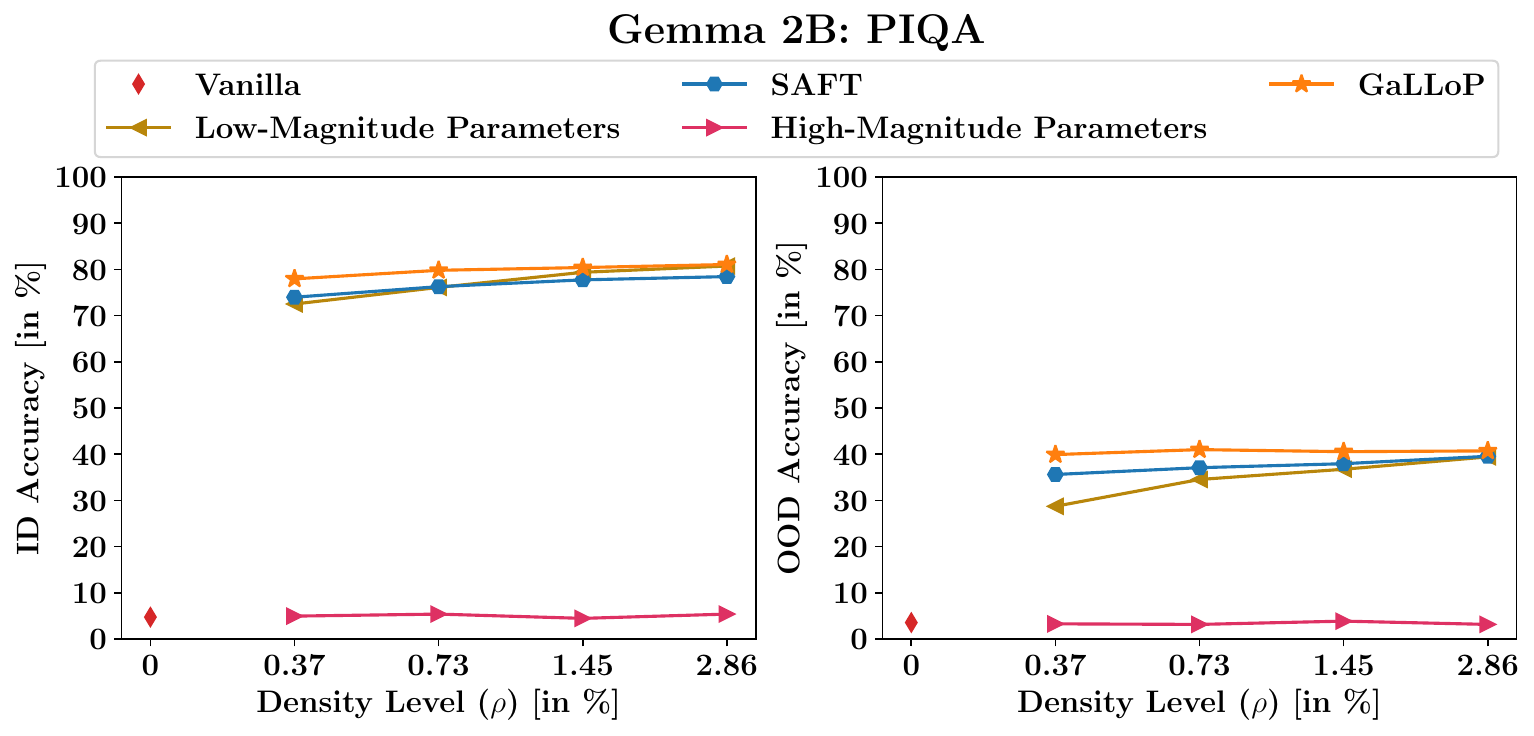}
        }
        \subfigure[]{
            \label{fig_CR_gemma_ablations_siqa}
            \includegraphics[scale = 0.25]{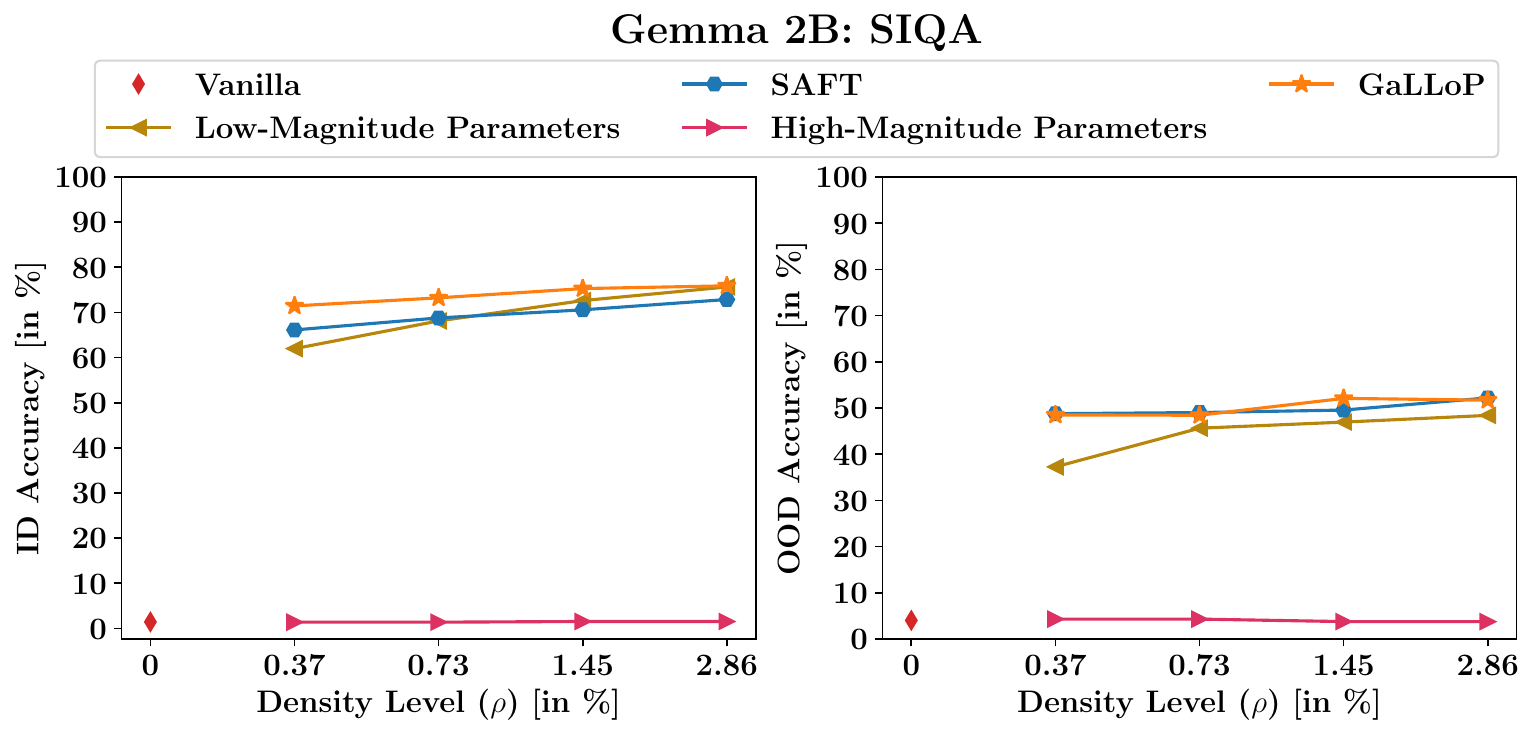}
        }
        \subfigure[]{
            \label{fig_CR_gemma_ablations_hellaswag}
            \includegraphics[scale = 0.25]{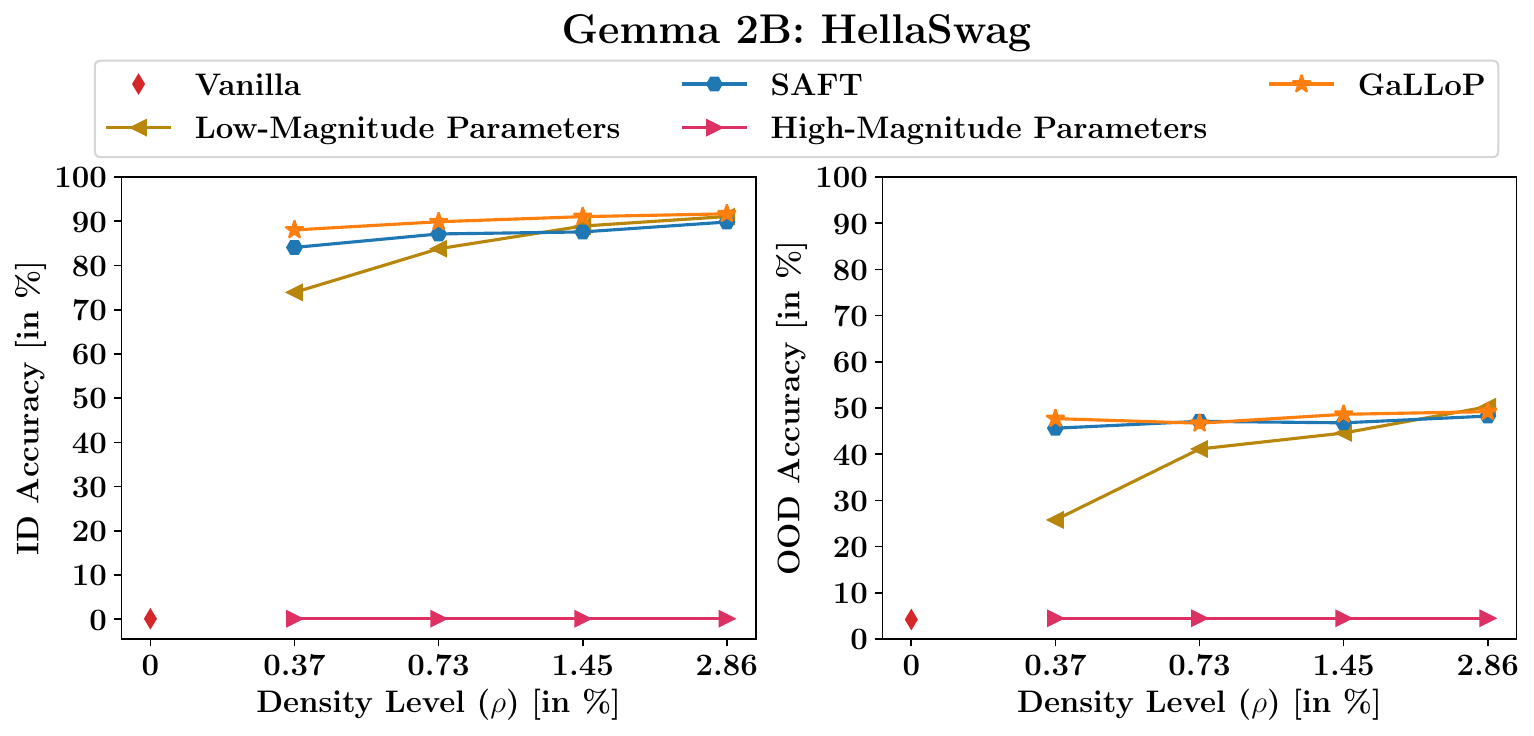}
        }
        \subfigure[]{
            \label{fig_CR_gemma_ablations_winogrande}
            \includegraphics[scale = 0.25]{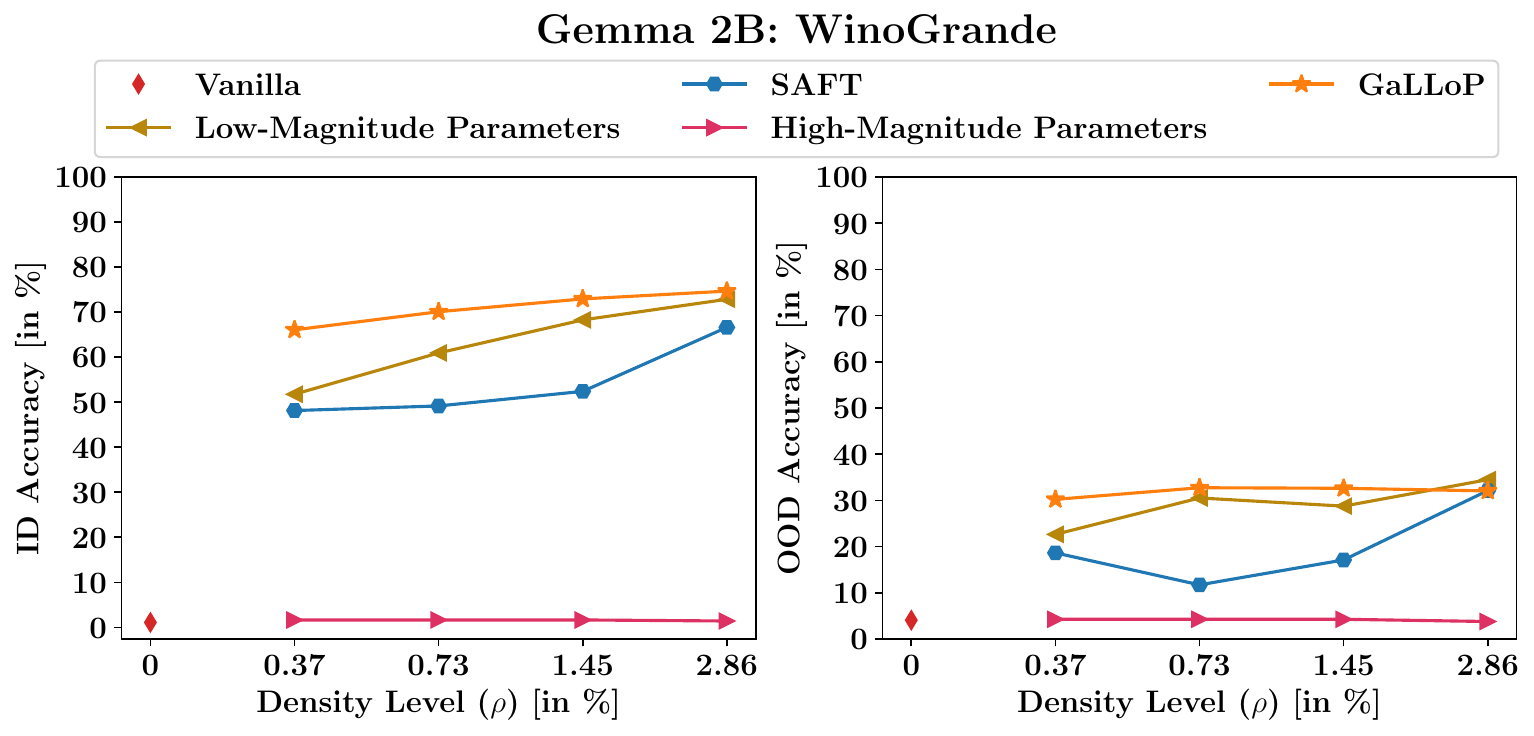}
        }
        \subfigure[]{
            \label{fig_CR_gemma_ablations_obqa}
            \includegraphics[scale = 0.25]{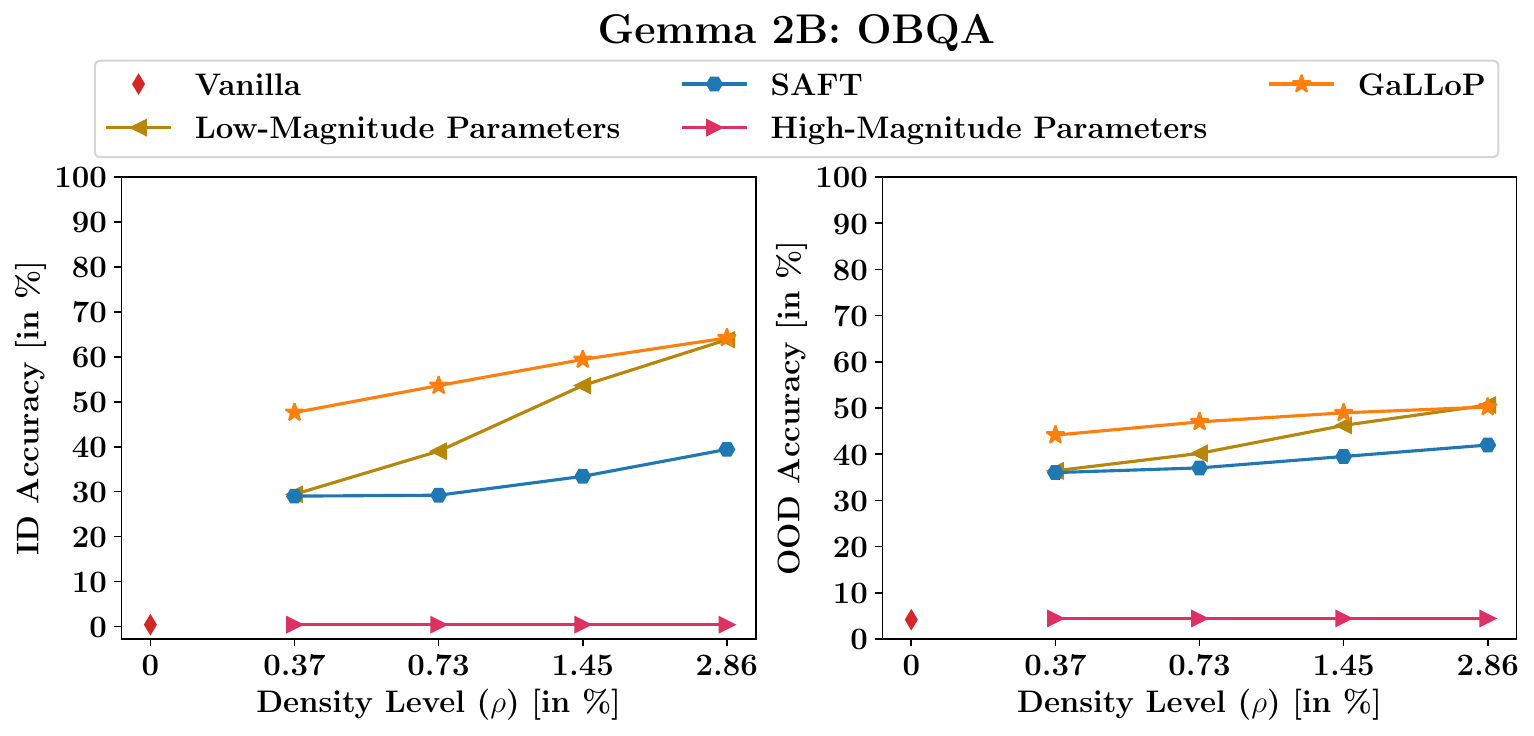}
        }
        \subfigure[]{
            \label{fig_CR_gemma_ablations_boolq}
            \includegraphics[scale = 0.25]{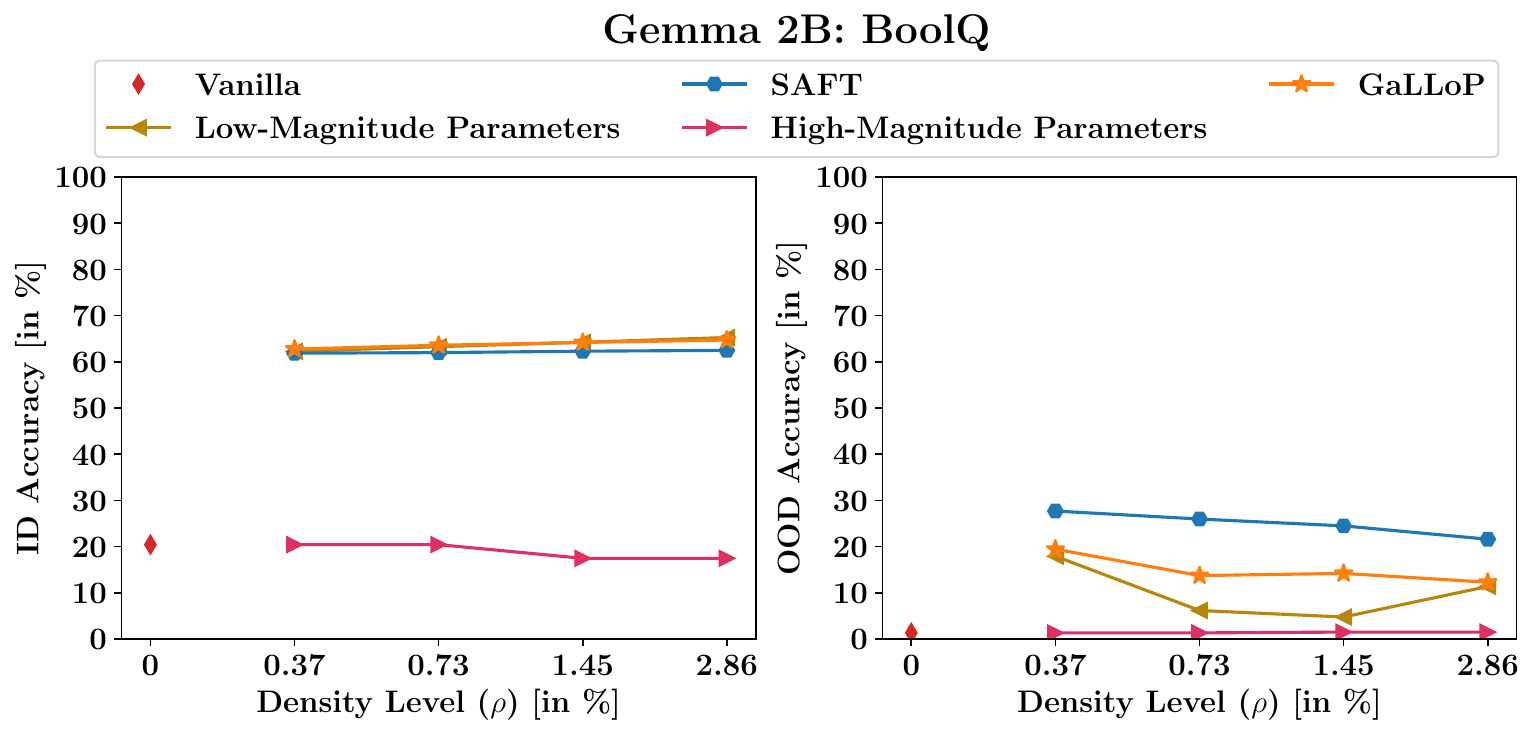}
        }
    \end{center}
    \caption{Gemma 2B models fine-tuned with GaLLoP form a dominant Pareto front for both ID and OOD performance over models fine-tuned by selecting parameters on the basis of either parameter or gradient magnitudes when fine-tuning is performed on a) ARC-c, b) ARC-e, c) PIQA, d) SIQA, e) HellaSwag, f) WinoGrande, g) OBQA, and h) BoolQ (except OOD).}
    \label{fig_CR_gemma_ablations_per_run}
\end{figure*}

To investigate whether GaLLoP's dual parameter selection criterion (parameters with high gradients and low-magnitudes) potentially leads to performance benefits over selecting parameters satisfying only one of these two criteria, we conduct an ablation study on Gemma 2B models.~\autoref{fig_ablations_full_avg} shows the results of this study with the corresponding averaged ID and OOD accuracies which are reported for all the four density levels. The per-run performance results are shown in~\autoref{fig_CR_gemma_ablations_per_run}. 

Fine-tuning with GaLLoP allows for the attainment of the highest ID and OOD accuracies: both on average and across all the datasets (except OOD accuracies obtained upon fine-tuning on BoolQ: see~\autoref{fig_CR_gemma_ablations_boolq} and~\autoref{app:id_ood} in~\autoref{AppendixD} for an explanation). Further, as can also be seen from~\autoref{fig_ablations_full_avg} and as discussed in~\autoref{sec:motivation}, fine-tuning high-magnitude parameters leads to no improvements over the performance of the non fine-tuned, vanilla counterpart. In contrast to this, fine-tuning low-magnitude parameters allows models to perform quite well on both ID and OOD tasks. In fact, except for the OOD performance obtained upon fine-tuning with BoolQ (see~\autoref{fig_CR_gemma_ablations_boolq}; the explanation from~\autoref{app:id_ood} in~\autoref{AppendixD} applies here as well, \emph{albeit} with lesser OOD performance metrics than the model fine-tuned with GaLLoP), ID and OOD performance levels obtained on fine-tuning low-magnitude parameters interpolate between the corresponding performance levels obtained on fine-tuning parameters with high gradients (as in SAFT; generally forms the lower bound) and fine-tuning low-magnitude parameters with high gradients (as in GaLLoP; forms the upper bound): see~\autoref{fig_CR_gemma_ablations_per_run}.

Moreover, with increasing density levels, performance obtained on fine-tuning low magnitude parameters generally overtakes that obtained on fine-tuning parameters with high gradients (as in SAFT) and finally, saturates at a performance level comparable to that obtained on fine-tuning with GaLLoP. This illustrates how GaLLoP potentially operates in different density regimes to yield the most superior performance: while the high gradient parameter selection criterion might dominate at lower density levels, the low-magnitude parameter selection criterion becomes increasingly dominant as the density level increases. 

Further, as mentioned earlier (see~\autoref{sec:ID_OOD_acc}), \emph{low-magnitude dilution} kicks in as we move closer towards the highest density level because of which we observe a somewhat saturation and convergence in the performance attained by models fine-tuned with GaLLoP and those fine-tuned by selecting parameters with low-magnitudes. This trend of saturation is also reflected in models fine-tuned with SAFT which suggests that \emph{high gradient dilution} starts to set in at that point as well. Effectively, in the asymptotic limit ($\rho \rightarrow 100\%$), as what can also be expected theoretically, the performance of models fine-tuned with static SpFT algorithms such as GaLLoP and SAFT, and models fine-tuned by selecting parameters with low-magnitudes, will converge to the performance of models fine-tuned with the dense FFT algorithm. This hypothesis is also supported by several previous studies wherein it was found that FFT predominantly updates a small fraction of low magnitude parameters for downstream task(s) and utilization of SpFT should be able to recover FFT performance~\citep{lth_bert_sparsity, lth_bert_sparsity_attn, sparsity_emergence, blockllm}. Nevertheless, as can be seen from our experiments, by combining the strengths of high gradients and low-magnitudes, fine-tuning with GaLLoP is expected to continue forming a performance upper bound over other static SpFT algorithms. 

\end{document}

%% file: math_commands.tex
\usepackage{amsmath,amsfonts,bm}

\def\eqref#1{equation~\ref{#1}}

\def\1{\bm{1}}

\DeclareMathAlphabet{\mathsfit}{\encodingdefault}{\sfdefault}{m}{sl}
\SetMathAlphabet{\mathsfit}{bold}{\encodingdefault}{\sfdefault}{bx}{n}













%% file: iclr2026_conference.bbl
\begin{thebibliography}{60}
\providecommand{\natexlab}[1]{#1}
\providecommand{\url}[1]{\texttt{#1}}
\expandafter\ifx\csname urlstyle\endcsname\relax
  \providecommand{\doi}[1]{doi: #1}\else
  \providecommand{\doi}{doi: \begingroup \urlstyle{rm}\Url}\fi

\bibitem[Aghajanyan et~al.(2021{\natexlab{a}})Aghajanyan, Gupta, and Zettlemoyer]{intrinsic_dim_2}
Armen Aghajanyan, Sonal Gupta, and Luke Zettlemoyer.
\newblock Intrinsic dimensionality explains the effectiveness of language model fine-tuning.
\newblock In Chengqing Zong, Fei Xia, Wenjie Li, and Roberto Navigli (eds.), \emph{Proceedings of the 59th Annual Meeting of the Association for Computational Linguistics and the 11th International Joint Conference on Natural Language Processing (Volume 1: Long Papers)}, pp.\  7319--7328, Online, August 2021{\natexlab{a}}. Association for Computational Linguistics.
\newblock \doi{10.18653/v1/2021.acl-long.568}.
\newblock URL \url{https://aclanthology.org/2021.acl-long.568/}.

\bibitem[Aghajanyan et~al.(2021{\natexlab{b}})Aghajanyan, Shrivastava, Gupta, Goyal, Zettlemoyer, and Gupta]{repr_collapse}
Armen Aghajanyan, Akshat Shrivastava, Anchit Gupta, Naman Goyal, Luke Zettlemoyer, and Sonal Gupta.
\newblock Better fine-tuning by reducing representational collapse.
\newblock In \emph{International Conference on Learning Representations}, 2021{\natexlab{b}}.
\newblock URL \url{https://openreview.net/forum?id=OQ08SN70M1V}.

\bibitem[Ansell et~al.(2022)Ansell, Ponti, Korhonen, and Vuli{\'c}]{lt_sft}
Alan Ansell, Edoardo Ponti, Anna Korhonen, and Ivan Vuli{\'c}.
\newblock Composable sparse fine-tuning for cross-lingual transfer.
\newblock In Smaranda Muresan, Preslav Nakov, and Aline Villavicencio (eds.), \emph{Proceedings of the 60th Annual Meeting of the Association for Computational Linguistics (Volume 1: Long Papers)}, pp.\  1778--1796, Dublin, Ireland, May 2022. Association for Computational Linguistics.
\newblock \doi{10.18653/v1/2022.acl-long.125}.
\newblock URL \url{https://aclanthology.org/2022.acl-long.125/}.

\bibitem[Ansell et~al.(2024)Ansell, Vulić, Sterz, Korhonen, and Ponti]{spiel}
Alan Ansell, Ivan Vulić, Hannah Sterz, Anna Korhonen, and Edoardo~M. Ponti.
\newblock Scaling sparse fine-tuning to large language models, 2024.
\newblock URL \url{https://arxiv.org/abs/2401.16405}.

\bibitem[Bisk et~al.(2020)Bisk, Zellers, Le~bras, Gao, and Choi]{piqa}
Yonatan Bisk, Rowan Zellers, Ronan Le~bras, Jianfeng Gao, and Yejin Choi.
\newblock Piqa: Reasoning about physical commonsense in natural language.
\newblock \emph{Proceedings of the AAAI Conference on Artificial Intelligence}, 34\penalty0 (05):\penalty0 7432--7439, Apr. 2020.
\newblock \doi{10.1609/aaai.v34i05.6239}.
\newblock URL \url{https://ojs.aaai.org/index.php/AAAI/article/view/6239}.

\bibitem[Borec et~al.(2024)Borec, Sadler, and Schlangen]{top_p_repetitive}
Luka Borec, Philipp Sadler, and David Schlangen.
\newblock The unreasonable ineffectiveness of nucleus sampling on mitigating text memorization.
\newblock In Saad Mahamood, Nguyen~Le Minh, and Daphne Ippolito (eds.), \emph{Proceedings of the 17th International Natural Language Generation Conference}, pp.\  358--370, Tokyo, Japan, September 2024. Association for Computational Linguistics.
\newblock \doi{10.18653/v1/2024.inlg-main.30}.
\newblock URL \url{https://aclanthology.org/2024.inlg-main.30/}.

\bibitem[Chen et~al.(2020)Chen, Frankle, Chang, Liu, Zhang, Wang, and Carbin]{lth_bert_sparsity}
Tianlong Chen, Jonathan Frankle, Shiyu Chang, Sijia Liu, Yang Zhang, Zhangyang Wang, and Michael Carbin.
\newblock The lottery ticket hypothesis for pre-trained bert networks.
\newblock In H.~Larochelle, M.~Ranzato, R.~Hadsell, M.F. Balcan, and H.~Lin (eds.), \emph{Advances in Neural Information Processing Systems}, volume~33, pp.\  15834--15846. Curran Associates, Inc., 2020.
\newblock URL \url{https://proceedings.neurips.cc/paper_files/paper/2020/file/b6af2c9703f203a2794be03d443af2e3-Paper.pdf}.

\bibitem[Clark et~al.(2019)Clark, Lee, Chang, Kwiatkowski, Collins, and Toutanova]{boolq}
Christopher Clark, Kenton Lee, Ming-Wei Chang, Tom Kwiatkowski, Michael Collins, and Kristina Toutanova.
\newblock {B}ool{Q}: Exploring the surprising difficulty of natural yes/no questions.
\newblock In Jill Burstein, Christy Doran, and Thamar Solorio (eds.), \emph{Proceedings of the 2019 Conference of the North {A}merican Chapter of the Association for Computational Linguistics: Human Language Technologies, Volume 1 (Long and Short Papers)}, pp.\  2924--2936, Minneapolis, Minnesota, June 2019. Association for Computational Linguistics.
\newblock \doi{10.18653/v1/N19-1300}.
\newblock URL \url{https://aclanthology.org/N19-1300/}.

\bibitem[Clark et~al.(2018)Clark, Cowhey, Etzioni, Khot, Sabharwal, Schoenick, and Tafjord]{arc}
Peter Clark, Isaac Cowhey, Oren Etzioni, Tushar Khot, Ashish Sabharwal, Carissa Schoenick, and Oyvind Tafjord.
\newblock Think you have solved question answering? try arc, the ai2 reasoning challenge, 2018.
\newblock URL \url{https://arxiv.org/abs/1803.05457}.

\bibitem[Dodge et~al.(2020)Dodge, Ilharco, Schwartz, Farhadi, Hajishirzi, and Smith]{fft_instability1}
Jesse Dodge, Gabriel Ilharco, Roy Schwartz, Ali Farhadi, Hannaneh Hajishirzi, and Noah~A. Smith.
\newblock Fine-tuning pretrained language models: Weight initializations, data orders, and early stopping.
\newblock \emph{CoRR}, abs/2002.06305, 2020.
\newblock URL \url{https://arxiv.org/abs/2002.06305}.

\bibitem[Fan et~al.(2018)Fan, Lewis, and Dauphin]{top_k}
Angela Fan, Mike Lewis, and Yann Dauphin.
\newblock Hierarchical neural story generation.
\newblock In Iryna Gurevych and Yusuke Miyao (eds.), \emph{Proceedings of the 56th Annual Meeting of the Association for Computational Linguistics (Volume 1: Long Papers)}, pp.\  889--898, Melbourne, Australia, July 2018. Association for Computational Linguistics.
\newblock \doi{10.18653/v1/P18-1082}.
\newblock URL \url{https://aclanthology.org/P18-1082/}.

\bibitem[Grattafiori et~al.(2024)Grattafiori, Dubey, Jauhri, Pandey, Kadian, Al-Dahle, Letman, Mathur, Schelten, Vaughan, Yang, Fan, Goyal, Hartshorn, Yang, Mitra, Sravankumar, Korenev, Hinsvark, Rao, Zhang, Rodriguez, Gregerson, Spataru, Roziere, Biron, Tang, Chern, Caucheteux, Nayak, Bi, Marra, McConnell, Keller, Touret, Wu, Wong, Ferrer, Nikolaidis, Allonsius, Song, Pintz, Livshits, Wyatt, Esiobu, Choudhary, Mahajan, Garcia-Olano, Perino, Hupkes, Lakomkin, AlBadawy, Lobanova, Dinan, Smith, Radenovic, Guzmán, Zhang, Synnaeve, Lee, Anderson, Thattai, Nail, Mialon, Pang, Cucurell, Nguyen, Korevaar, Xu, Touvron, Zarov, Ibarra, Kloumann, Misra, Evtimov, Zhang, Copet, Lee, Geffert, Vranes, Park, Mahadeokar, Shah, van~der Linde, Billock, Hong, Lee, Fu, Chi, Huang, Liu, Wang, Yu, Bitton, Spisak, Park, Rocca, Johnstun, Saxe, Jia, Alwala, Prasad, Upasani, Plawiak, Li, Heafield, Stone, El-Arini, Iyer, Malik, Chiu, Bhalla, Lakhotia, Rantala-Yeary, van~der Maaten, Chen, Tan, Jenkins, Martin, Madaan, Malo, Blecher,
  Landzaat, de~Oliveira, Muzzi, Pasupuleti, Singh, Paluri, Kardas, Tsimpoukelli, Oldham, Rita, Pavlova, Kambadur, Lewis, Si, Singh, Hassan, Goyal, Torabi, Bashlykov, Bogoychev, Chatterji, Zhang, Duchenne, Çelebi, Alrassy, Zhang, Li, Vasic, Weng, Bhargava, Dubal, Krishnan, Koura, Xu, He, Dong, Srinivasan, Ganapathy, Calderer, Cabral, Stojnic, Raileanu, Maheswari, Girdhar, Patel, Sauvestre, Polidoro, Sumbaly, Taylor, Silva, Hou, Wang, Hosseini, Chennabasappa, Singh, Bell, Kim, Edunov, Nie, Narang, Raparthy, Shen, Wan, Bhosale, Zhang, Vandenhende, Batra, Whitman, Sootla, Collot, Gururangan, Borodinsky, Herman, Fowler, Sheasha, Georgiou, Scialom, Speckbacher, Mihaylov, Xiao, Karn, Goswami, Gupta, Ramanathan, Kerkez, Gonguet, Do, Vogeti, Albiero, Petrovic, Chu, Xiong, Fu, Meers, Martinet, Wang, Wang, Tan, Xia, Xie, Jia, Wang, Goldschlag, Gaur, Babaei, Wen, Song, Zhang, Li, Mao, Coudert, Yan, Chen, Papakipos, Singh, Srivastava, Jain, Kelsey, Shajnfeld, Gangidi, Victoria, Goldstand, Menon, Sharma, Boesenberg,
  Baevski, Feinstein, Kallet, Sangani, Teo, Yunus, Lupu, Alvarado, Caples, Gu, Ho, Poulton, Ryan, Ramchandani, Dong, Franco, Goyal, Saraf, Chowdhury, Gabriel, Bharambe, Eisenman, Yazdan, James, Maurer, Leonhardi, Huang, Loyd, Paola, Paranjape, Liu, Wu, Ni, Hancock, Wasti, Spence, Stojkovic, Gamido, Montalvo, Parker, Burton, Mejia, Liu, Wang, Kim, Zhou, Hu, Chu, Cai, Tindal, Feichtenhofer, Gao, Civin, Beaty, Kreymer, Li, Adkins, Xu, Testuggine, David, Parikh, Liskovich, Foss, Wang, Le, Holland, Dowling, Jamil, Montgomery, Presani, Hahn, Wood, Le, Brinkman, Arcaute, Dunbar, Smothers, Sun, Kreuk, Tian, Kokkinos, Ozgenel, Caggioni, Kanayet, Seide, Florez, Schwarz, Badeer, Swee, Halpern, Herman, Sizov, Guangyi, Zhang, Lakshminarayanan, Inan, Shojanazeri, Zou, Wang, Zha, Habeeb, Rudolph, Suk, Aspegren, Goldman, Zhan, Damlaj, Molybog, Tufanov, Leontiadis, Veliche, Gat, Weissman, Geboski, Kohli, Lam, Asher, Gaya, Marcus, Tang, Chan, Zhen, Reizenstein, Teboul, Zhong, Jin, Yang, Cummings, Carvill, Shepard, McPhie,
  Torres, Ginsburg, Wang, Wu, U, Saxena, Khandelwal, Zand, Matosich, Veeraraghavan, Michelena, Li, Jagadeesh, Huang, Chawla, Huang, Chen, Garg, A, Silva, Bell, Zhang, Guo, Yu, Moshkovich, Wehrstedt, Khabsa, Avalani, Bhatt, Mankus, Hasson, Lennie, Reso, Groshev, Naumov, Lathi, Keneally, Liu, Seltzer, Valko, Restrepo, Patel, Vyatskov, Samvelyan, Clark, Macey, Wang, Hermoso, Metanat, Rastegari, Bansal, Santhanam, Parks, White, Bawa, Singhal, Egebo, Usunier, Mehta, Laptev, Dong, Cheng, Chernoguz, Hart, Salpekar, Kalinli, Kent, Parekh, Saab, Balaji, Rittner, Bontrager, Roux, Dollar, Zvyagina, Ratanchandani, Yuvraj, Liang, Alao, Rodriguez, Ayub, Murthy, Nayani, Mitra, Parthasarathy, Li, Hogan, Battey, Wang, Howes, Rinott, Mehta, Siby, Bondu, Datta, Chugh, Hunt, Dhillon, Sidorov, Pan, Mahajan, Verma, Yamamoto, Ramaswamy, Lindsay, Lindsay, Feng, Lin, Zha, Patil, Shankar, Zhang, Zhang, Wang, Agarwal, Sajuyigbe, Chintala, Max, Chen, Kehoe, Satterfield, Govindaprasad, Gupta, Deng, Cho, Virk, Subramanian, Choudhury,
  Goldman, Remez, Glaser, Best, Koehler, Robinson, Li, Zhang, Matthews, Chou, Shaked, Vontimitta, Ajayi, Montanez, Mohan, Kumar, Mangla, Ionescu, Poenaru, Mihailescu, Ivanov, Li, Wang, Jiang, Bouaziz, Constable, Tang, Wu, Wang, Wu, Gao, Kleinman, Chen, Hu, Jia, Qi, Li, Zhang, Zhang, Adi, Nam, Yu, Wang, Zhao, Hao, Qian, Li, He, Rait, DeVito, Rosnbrick, Wen, Yang, Zhao, and Ma]{llama}
Aaron Grattafiori, Abhimanyu Dubey, Abhinav Jauhri, Abhinav Pandey, Abhishek Kadian, Ahmad Al-Dahle, Aiesha Letman, Akhil Mathur, Alan Schelten, Alex Vaughan, Amy Yang, Angela Fan, Anirudh Goyal, Anthony Hartshorn, Aobo Yang, Archi Mitra, Archie Sravankumar, Artem Korenev, Arthur Hinsvark, Arun Rao, Aston Zhang, Aurelien Rodriguez, Austen Gregerson, Ava Spataru, Baptiste Roziere, Bethany Biron, Binh Tang, Bobbie Chern, Charlotte Caucheteux, Chaya Nayak, Chloe Bi, Chris Marra, Chris McConnell, Christian Keller, Christophe Touret, Chunyang Wu, Corinne Wong, Cristian~Canton Ferrer, Cyrus Nikolaidis, Damien Allonsius, Daniel Song, Danielle Pintz, Danny Livshits, Danny Wyatt, David Esiobu, Dhruv Choudhary, Dhruv Mahajan, Diego Garcia-Olano, Diego Perino, Dieuwke Hupkes, Egor Lakomkin, Ehab AlBadawy, Elina Lobanova, Emily Dinan, Eric~Michael Smith, Filip Radenovic, Francisco Guzmán, Frank Zhang, Gabriel Synnaeve, Gabrielle Lee, Georgia~Lewis Anderson, Govind Thattai, Graeme Nail, Gregoire Mialon, Guan Pang,
  Guillem Cucurell, Hailey Nguyen, Hannah Korevaar, Hu~Xu, Hugo Touvron, Iliyan Zarov, Imanol~Arrieta Ibarra, Isabel Kloumann, Ishan Misra, Ivan Evtimov, Jack Zhang, Jade Copet, Jaewon Lee, Jan Geffert, Jana Vranes, Jason Park, Jay Mahadeokar, Jeet Shah, Jelmer van~der Linde, Jennifer Billock, Jenny Hong, Jenya Lee, Jeremy Fu, Jianfeng Chi, Jianyu Huang, Jiawen Liu, Jie Wang, Jiecao Yu, Joanna Bitton, Joe Spisak, Jongsoo Park, Joseph Rocca, Joshua Johnstun, Joshua Saxe, Junteng Jia, Kalyan~Vasuden Alwala, Karthik Prasad, Kartikeya Upasani, Kate Plawiak, Ke~Li, Kenneth Heafield, Kevin Stone, Khalid El-Arini, Krithika Iyer, Kshitiz Malik, Kuenley Chiu, Kunal Bhalla, Kushal Lakhotia, Lauren Rantala-Yeary, Laurens van~der Maaten, Lawrence Chen, Liang Tan, Liz Jenkins, Louis Martin, Lovish Madaan, Lubo Malo, Lukas Blecher, Lukas Landzaat, Luke de~Oliveira, Madeline Muzzi, Mahesh Pasupuleti, Mannat Singh, Manohar Paluri, Marcin Kardas, Maria Tsimpoukelli, Mathew Oldham, Mathieu Rita, Maya Pavlova, Melanie Kambadur,
  Mike Lewis, Min Si, Mitesh~Kumar Singh, Mona Hassan, Naman Goyal, Narjes Torabi, Nikolay Bashlykov, Nikolay Bogoychev, Niladri Chatterji, Ning Zhang, Olivier Duchenne, Onur Çelebi, Patrick Alrassy, Pengchuan Zhang, Pengwei Li, Petar Vasic, Peter Weng, Prajjwal Bhargava, Pratik Dubal, Praveen Krishnan, Punit~Singh Koura, Puxin Xu, Qing He, Qingxiao Dong, Ragavan Srinivasan, Raj Ganapathy, Ramon Calderer, Ricardo~Silveira Cabral, Robert Stojnic, Roberta Raileanu, Rohan Maheswari, Rohit Girdhar, Rohit Patel, Romain Sauvestre, Ronnie Polidoro, Roshan Sumbaly, Ross Taylor, Ruan Silva, Rui Hou, Rui Wang, Saghar Hosseini, Sahana Chennabasappa, Sanjay Singh, Sean Bell, Seohyun~Sonia Kim, Sergey Edunov, Shaoliang Nie, Sharan Narang, Sharath Raparthy, Sheng Shen, Shengye Wan, Shruti Bhosale, Shun Zhang, Simon Vandenhende, Soumya Batra, Spencer Whitman, Sten Sootla, Stephane Collot, Suchin Gururangan, Sydney Borodinsky, Tamar Herman, Tara Fowler, Tarek Sheasha, Thomas Georgiou, Thomas Scialom, Tobias Speckbacher,
  Todor Mihaylov, Tong Xiao, Ujjwal Karn, Vedanuj Goswami, Vibhor Gupta, Vignesh Ramanathan, Viktor Kerkez, Vincent Gonguet, Virginie Do, Vish Vogeti, Vítor Albiero, Vladan Petrovic, Weiwei Chu, Wenhan Xiong, Wenyin Fu, Whitney Meers, Xavier Martinet, Xiaodong Wang, Xiaofang Wang, Xiaoqing~Ellen Tan, Xide Xia, Xinfeng Xie, Xuchao Jia, Xuewei Wang, Yaelle Goldschlag, Yashesh Gaur, Yasmine Babaei, Yi~Wen, Yiwen Song, Yuchen Zhang, Yue Li, Yuning Mao, Zacharie~Delpierre Coudert, Zheng Yan, Zhengxing Chen, Zoe Papakipos, Aaditya Singh, Aayushi Srivastava, Abha Jain, Adam Kelsey, Adam Shajnfeld, Adithya Gangidi, Adolfo Victoria, Ahuva Goldstand, Ajay Menon, Ajay Sharma, Alex Boesenberg, Alexei Baevski, Allie Feinstein, Amanda Kallet, Amit Sangani, Amos Teo, Anam Yunus, Andrei Lupu, Andres Alvarado, Andrew Caples, Andrew Gu, Andrew Ho, Andrew Poulton, Andrew Ryan, Ankit Ramchandani, Annie Dong, Annie Franco, Anuj Goyal, Aparajita Saraf, Arkabandhu Chowdhury, Ashley Gabriel, Ashwin Bharambe, Assaf Eisenman, Azadeh
  Yazdan, Beau James, Ben Maurer, Benjamin Leonhardi, Bernie Huang, Beth Loyd, Beto~De Paola, Bhargavi Paranjape, Bing Liu, Bo~Wu, Boyu Ni, Braden Hancock, Bram Wasti, Brandon Spence, Brani Stojkovic, Brian Gamido, Britt Montalvo, Carl Parker, Carly Burton, Catalina Mejia, Ce~Liu, Changhan Wang, Changkyu Kim, Chao Zhou, Chester Hu, Ching-Hsiang Chu, Chris Cai, Chris Tindal, Christoph Feichtenhofer, Cynthia Gao, Damon Civin, Dana Beaty, Daniel Kreymer, Daniel Li, David Adkins, David Xu, Davide Testuggine, Delia David, Devi Parikh, Diana Liskovich, Didem Foss, Dingkang Wang, Duc Le, Dustin Holland, Edward Dowling, Eissa Jamil, Elaine Montgomery, Eleonora Presani, Emily Hahn, Emily Wood, Eric-Tuan Le, Erik Brinkman, Esteban Arcaute, Evan Dunbar, Evan Smothers, Fei Sun, Felix Kreuk, Feng Tian, Filippos Kokkinos, Firat Ozgenel, Francesco Caggioni, Frank Kanayet, Frank Seide, Gabriela~Medina Florez, Gabriella Schwarz, Gada Badeer, Georgia Swee, Gil Halpern, Grant Herman, Grigory Sizov, Guangyi, Zhang, Guna
  Lakshminarayanan, Hakan Inan, Hamid Shojanazeri, Han Zou, Hannah Wang, Hanwen Zha, Haroun Habeeb, Harrison Rudolph, Helen Suk, Henry Aspegren, Hunter Goldman, Hongyuan Zhan, Ibrahim Damlaj, Igor Molybog, Igor Tufanov, Ilias Leontiadis, Irina-Elena Veliche, Itai Gat, Jake Weissman, James Geboski, James Kohli, Janice Lam, Japhet Asher, Jean-Baptiste Gaya, Jeff Marcus, Jeff Tang, Jennifer Chan, Jenny Zhen, Jeremy Reizenstein, Jeremy Teboul, Jessica Zhong, Jian Jin, Jingyi Yang, Joe Cummings, Jon Carvill, Jon Shepard, Jonathan McPhie, Jonathan Torres, Josh Ginsburg, Junjie Wang, Kai Wu, Kam~Hou U, Karan Saxena, Kartikay Khandelwal, Katayoun Zand, Kathy Matosich, Kaushik Veeraraghavan, Kelly Michelena, Keqian Li, Kiran Jagadeesh, Kun Huang, Kunal Chawla, Kyle Huang, Lailin Chen, Lakshya Garg, Lavender A, Leandro Silva, Lee Bell, Lei Zhang, Liangpeng Guo, Licheng Yu, Liron Moshkovich, Luca Wehrstedt, Madian Khabsa, Manav Avalani, Manish Bhatt, Martynas Mankus, Matan Hasson, Matthew Lennie, Matthias Reso, Maxim
  Groshev, Maxim Naumov, Maya Lathi, Meghan Keneally, Miao Liu, Michael~L. Seltzer, Michal Valko, Michelle Restrepo, Mihir Patel, Mik Vyatskov, Mikayel Samvelyan, Mike Clark, Mike Macey, Mike Wang, Miquel~Jubert Hermoso, Mo~Metanat, Mohammad Rastegari, Munish Bansal, Nandhini Santhanam, Natascha Parks, Natasha White, Navyata Bawa, Nayan Singhal, Nick Egebo, Nicolas Usunier, Nikhil Mehta, Nikolay~Pavlovich Laptev, Ning Dong, Norman Cheng, Oleg Chernoguz, Olivia Hart, Omkar Salpekar, Ozlem Kalinli, Parkin Kent, Parth Parekh, Paul Saab, Pavan Balaji, Pedro Rittner, Philip Bontrager, Pierre Roux, Piotr Dollar, Polina Zvyagina, Prashant Ratanchandani, Pritish Yuvraj, Qian Liang, Rachad Alao, Rachel Rodriguez, Rafi Ayub, Raghotham Murthy, Raghu Nayani, Rahul Mitra, Rangaprabhu Parthasarathy, Raymond Li, Rebekkah Hogan, Robin Battey, Rocky Wang, Russ Howes, Ruty Rinott, Sachin Mehta, Sachin Siby, Sai~Jayesh Bondu, Samyak Datta, Sara Chugh, Sara Hunt, Sargun Dhillon, Sasha Sidorov, Satadru Pan, Saurabh Mahajan,
  Saurabh Verma, Seiji Yamamoto, Sharadh Ramaswamy, Shaun Lindsay, Shaun Lindsay, Sheng Feng, Shenghao Lin, Shengxin~Cindy Zha, Shishir Patil, Shiva Shankar, Shuqiang Zhang, Shuqiang Zhang, Sinong Wang, Sneha Agarwal, Soji Sajuyigbe, Soumith Chintala, Stephanie Max, Stephen Chen, Steve Kehoe, Steve Satterfield, Sudarshan Govindaprasad, Sumit Gupta, Summer Deng, Sungmin Cho, Sunny Virk, Suraj Subramanian, Sy~Choudhury, Sydney Goldman, Tal Remez, Tamar Glaser, Tamara Best, Thilo Koehler, Thomas Robinson, Tianhe Li, Tianjun Zhang, Tim Matthews, Timothy Chou, Tzook Shaked, Varun Vontimitta, Victoria Ajayi, Victoria Montanez, Vijai Mohan, Vinay~Satish Kumar, Vishal Mangla, Vlad Ionescu, Vlad Poenaru, Vlad~Tiberiu Mihailescu, Vladimir Ivanov, Wei Li, Wenchen Wang, Wenwen Jiang, Wes Bouaziz, Will Constable, Xiaocheng Tang, Xiaojian Wu, Xiaolan Wang, Xilun Wu, Xinbo Gao, Yaniv Kleinman, Yanjun Chen, Ye~Hu, Ye~Jia, Ye~Qi, Yenda Li, Yilin Zhang, Ying Zhang, Yossi Adi, Youngjin Nam, Yu, Wang, Yu~Zhao, Yuchen Hao, Yundi
  Qian, Yunlu Li, Yuzi He, Zach Rait, Zachary DeVito, Zef Rosnbrick, Zhaoduo Wen, Zhenyu Yang, Zhiwei Zhao, and Zhiyu Ma.
\newblock The llama 3 herd of models, 2024.
\newblock URL \url{https://arxiv.org/abs/2407.21783}.

\bibitem[Graves(2012)]{beam_search}
Alex Graves.
\newblock Sequence transduction with recurrent neural networks.
\newblock \emph{CoRR}, abs/1211.3711, 2012.
\newblock URL \url{http://arxiv.org/abs/1211.3711}.

\bibitem[Guo et~al.(2021)Guo, Rush, and Kim]{diff_pruning}
Demi Guo, Alexander Rush, and Yoon Kim.
\newblock Parameter-efficient transfer learning with diff pruning.
\newblock In Chengqing Zong, Fei Xia, Wenjie Li, and Roberto Navigli (eds.), \emph{Proceedings of the 59th Annual Meeting of the Association for Computational Linguistics and the 11th International Joint Conference on Natural Language Processing (Volume 1: Long Papers)}, pp.\  4884--4896, Online, August 2021. Association for Computational Linguistics.
\newblock \doi{10.18653/v1/2021.acl-long.378}.
\newblock URL \url{https://aclanthology.org/2021.acl-long.378/}.

\bibitem[Han et~al.(2015)Han, Pool, Tran, and Dally]{prune1}
Song Han, Jeff Pool, John Tran, and William~J. Dally.
\newblock Learning both weights and connections for efficient neural networks.
\newblock In \emph{Proceedings of the 29th International Conference on Neural Information Processing Systems - Volume 1}, NIPS'15, pp.\  1135–1143, Cambridge, MA, USA, 2015. MIT Press.

\bibitem[Hayou et~al.(2024)Hayou, Ghosh, and Yu]{lora_plus}
Soufiane Hayou, Nikhil Ghosh, and Bin Yu.
\newblock Lo{RA}+: Efficient low rank adaptation of large models.
\newblock In \emph{Forty-first International Conference on Machine Learning}, 2024.
\newblock URL \url{https://openreview.net/forum?id=NEv8YqBROO}.

\bibitem[He et~al.(2025)He, Li, Jiang, and Miller]{smt}
Haoze He, Juncheng~B Li, Xuan Jiang, and Heather Miller.
\newblock {SMT}: Fine-tuning large language models with sparse matrices.
\newblock In \emph{The Thirteenth International Conference on Learning Representations}, 2025.
\newblock URL \url{https://openreview.net/forum?id=GbgCRJedQ7}.

\bibitem[Holtzman et~al.(2020)Holtzman, Buys, Du, Forbes, and Choi]{top_p}
Ari Holtzman, Jan Buys, Li~Du, Maxwell Forbes, and Yejin Choi.
\newblock The curious case of neural text degeneration.
\newblock In \emph{International Conference on Learning Representations}, 2020.
\newblock URL \url{https://openreview.net/forum?id=rygGQyrFvH}.

\bibitem[Hooker(2021)]{hardware_lottery}
Sara Hooker.
\newblock The hardware lottery.
\newblock \emph{Commun. ACM}, 64\penalty0 (12):\penalty0 58–65, November 2021.
\newblock ISSN 0001-0782.
\newblock \doi{10.1145/3467017}.
\newblock URL \url{https://doi.org/10.1145/3467017}.

\bibitem[Houlsby et~al.(2019)Houlsby, Giurgiu, Jastrzebski, Morrone, De~Laroussilhe, Gesmundo, Attariyan, and Gelly]{adapter_orig}
Neil Houlsby, Andrei Giurgiu, Stanislaw Jastrzebski, Bruna Morrone, Quentin De~Laroussilhe, Andrea Gesmundo, Mona Attariyan, and Sylvain Gelly.
\newblock Parameter-efficient transfer learning for {NLP}.
\newblock In \emph{Proceedings of the 36th International Conference on Machine Learning}, 2019.

\bibitem[Hu et~al.(2022)Hu, yelong shen, Wallis, Allen-Zhu, Li, Wang, Wang, and Chen]{lora}
Edward~J Hu, yelong shen, Phillip Wallis, Zeyuan Allen-Zhu, Yuanzhi Li, Shean Wang, Lu~Wang, and Weizhu Chen.
\newblock Lo{RA}: Low-rank adaptation of large language models.
\newblock In \emph{International Conference on Learning Representations}, 2022.
\newblock URL \url{https://openreview.net/forum?id=nZeVKeeFYf9}.

\bibitem[Hu et~al.(2023)Hu, Wang, Lan, Xu, Lim, Bing, Xu, Poria, and Lee]{llm-adapters}
Zhiqiang Hu, Lei Wang, Yihuai Lan, Wanyu Xu, Ee-Peng Lim, Lidong Bing, Xing Xu, Soujanya Poria, and Roy Ka-Wei Lee.
\newblock {LLM}-adapters: An adapter family for parameter-efficient fine-tuning of large language models.
\newblock In \emph{The 2023 Conference on Empirical Methods in Natural Language Processing}, 2023.
\newblock URL \url{https://openreview.net/forum?id=gdUBK65fwn}.

\bibitem[Huang et~al.(2025)Huang, Ko, Zhuang, Tang, and Zhang]{hira}
Qiushi Huang, Tom Ko, Zhan Zhuang, Lilian Tang, and Yu~Zhang.
\newblock Hi{RA}: Parameter-efficient hadamard high-rank adaptation for large language models.
\newblock In \emph{The Thirteenth International Conference on Learning Representations}, 2025.
\newblock URL \url{https://openreview.net/forum?id=TwJrTz9cRS}.

\bibitem[Jaiswal et~al.(2023)Jaiswal, Liu, Chen, and Wang]{sparsity_emergence}
Ajay~Kumar Jaiswal, Shiwei Liu, Tianlong Chen, and Zhangyang Wang.
\newblock The emergence of essential sparsity in large pre-trained models: The weights that matter.
\newblock In \emph{Thirty-seventh Conference on Neural Information Processing Systems}, 2023.
\newblock URL \url{https://openreview.net/forum?id=bU9hwbsVcy}.

\bibitem[Kopiczko et~al.(2024)Kopiczko, Blankevoort, and Asano]{vera}
Dawid~Jan Kopiczko, Tijmen Blankevoort, and Yuki~M Asano.
\newblock Ve{RA}: Vector-based random matrix adaptation.
\newblock In \emph{The Twelfth International Conference on Learning Representations}, 2024.
\newblock URL \url{https://openreview.net/forum?id=NjNfLdxr3A}.

\bibitem[Krishna et~al.(2017)Krishna, Hata, Ren, Fei-Fei, and Niebles]{activitynet}
Ranjay Krishna, Kenji Hata, Frederic Ren, Li~Fei-Fei, and Juan~Carlos Niebles.
\newblock Dense-captioning events in videos.
\newblock In \emph{2017 IEEE International Conference on Computer Vision (ICCV)}, pp.\  706--715, 2017.
\newblock \doi{10.1109/ICCV.2017.83}.

\bibitem[Kumar et~al.(2022)Kumar, Raghunathan, Jones, Ma, and Liang]{FFT_distortion}
Ananya Kumar, Aditi Raghunathan, Robbie~Matthew Jones, Tengyu Ma, and Percy Liang.
\newblock Fine-tuning can distort pretrained features and underperform out-of-distribution.
\newblock In \emph{International Conference on Learning Representations}, 2022.
\newblock URL \url{https://openreview.net/forum?id=UYneFzXSJWh}.

\bibitem[Lee et~al.(2019)Lee, Ajanthan, and Torr]{snip}
Namhoon Lee, Thalaiyasingam Ajanthan, and Philip Torr.
\newblock {SNIP}: {SINGLE}-{SHOT} {NETWORK} {PRUNING} {BASED} {ON} {CONNECTION} {SENSITIVITY}.
\newblock In \emph{International Conference on Learning Representations}, 2019.
\newblock URL \url{https://openreview.net/forum?id=B1VZqjAcYX}.

\bibitem[Liao et~al.(2023)Liao, Meng, and Monz]{pafi}
Baohao Liao, Yan Meng, and Christof Monz.
\newblock Parameter-efficient fine-tuning without introducing new latency.
\newblock In Anna Rogers, Jordan Boyd-Graber, and Naoaki Okazaki (eds.), \emph{Proceedings of the 61st Annual Meeting of the Association for Computational Linguistics (Volume 1: Long Papers)}, pp.\  4242--4260, Toronto, Canada, July 2023. Association for Computational Linguistics.
\newblock \doi{10.18653/v1/2023.acl-long.233}.
\newblock URL \url{https://aclanthology.org/2023.acl-long.233/}.

\bibitem[Lin et~al.(2020)Lin, Madotto, and Fung]{adapter_new}
Zhaojiang Lin, Andrea Madotto, and Pascale Fung.
\newblock Exploring versatile generative language model via parameter-efficient transfer learning.
\newblock In Trevor Cohn, Yulan He, and Yang Liu (eds.), \emph{Findings of the Association for Computational Linguistics: EMNLP 2020}, pp.\  441--459, Online, November 2020. Association for Computational Linguistics.
\newblock \doi{10.18653/v1/2020.findings-emnlp.41}.
\newblock URL \url{https://aclanthology.org/2020.findings-emnlp.41/}.

\bibitem[Mangrulkar et~al.(2022)Mangrulkar, Gugger, Debut, Belkada, Paul, and Bossan]{hf_peft}
Sourab Mangrulkar, Sylvain Gugger, Lysandre Debut, Younes Belkada, Sayak Paul, and Benjamin Bossan.
\newblock {PEFT}: State-of-the-art parameter-efficient fine-tuning methods.
\newblock \url{https://github.com/huggingface/peft}, 2022.

\bibitem[McCloskey \& Cohen(1989)McCloskey and Cohen]{cfr}
Michael McCloskey and Neal~J. Cohen.
\newblock Catastrophic interference in connectionist networks: The sequential learning problem.
\newblock volume~24 of \emph{Psychology of Learning and Motivation}, pp.\  109--165. Academic Press, 1989.
\newblock \doi{https://doi.org/10.1016/S0079-7421(08)60536-8}.
\newblock URL \url{https://www.sciencedirect.com/science/article/pii/S0079742108605368}.

\bibitem[Mihaylov et~al.(2018)Mihaylov, Clark, Khot, and Sabharwal]{obqa}
Todor Mihaylov, Peter Clark, Tushar Khot, and Ashish Sabharwal.
\newblock Can a suit of armor conduct electricity? a new dataset for open book question answering.
\newblock In Ellen Riloff, David Chiang, Julia Hockenmaier, and Jun{'}ichi Tsujii (eds.), \emph{Proceedings of the 2018 Conference on Empirical Methods in Natural Language Processing}, pp.\  2381--2391, Brussels, Belgium, October-November 2018. Association for Computational Linguistics.
\newblock \doi{10.18653/v1/D18-1260}.
\newblock URL \url{https://aclanthology.org/D18-1260/}.

\bibitem[Miller et~al.(2021)Miller, Taori, Raghunathan, Sagawa, Koh, Shankar, Liang, Carmon, and Schmidt]{ID_OOD_study}
John~P Miller, Rohan Taori, Aditi Raghunathan, Shiori Sagawa, Pang~Wei Koh, Vaishaal Shankar, Percy Liang, Yair Carmon, and Ludwig Schmidt.
\newblock Accuracy on the line: on the strong correlation between out-of-distribution and in-distribution generalization.
\newblock In Marina Meila and Tong Zhang (eds.), \emph{Proceedings of the 38th International Conference on Machine Learning}, volume 139 of \emph{Proceedings of Machine Learning Research}, pp.\  7721--7735. PMLR, 18--24 Jul 2021.
\newblock URL \url{https://proceedings.mlr.press/v139/miller21b.html}.

\bibitem[Mosbach et~al.(2021)Mosbach, Andriushchenko, and Klakow]{fft_instability2}
Marius Mosbach, Maksym Andriushchenko, and Dietrich Klakow.
\newblock On the stability of fine-tuning {\{}bert{\}}: Misconceptions, explanations, and strong baselines.
\newblock In \emph{International Conference on Learning Representations}, 2021.
\newblock URL \url{https://openreview.net/forum?id=nzpLWnVAyah}.

\bibitem[Nguyen et~al.(2024)Nguyen, Uhlich, Cardinaux, Mauch, Edraki, and Courville]{saft}
Bac Nguyen, Stefan Uhlich, Fabien Cardinaux, Lukas Mauch, Marzieh Edraki, and Aaron Courville.
\newblock Saft: Towards out-of-distribution generalization in fine-tuning.
\newblock In \emph{Computer Vision – ECCV 2024: 18th European Conference, Milan, Italy, September 29–October 4, 2024, Proceedings, Part LXIX}, pp.\  138–154, Berlin, Heidelberg, 2024. Springer-Verlag.
\newblock ISBN 978-3-031-72889-1.
\newblock \doi{10.1007/978-3-031-72890-7_9}.
\newblock URL \url{https://doi.org/10.1007/978-3-031-72890-7_9}.

\bibitem[Nikdan et~al.(2024)Nikdan, Tabesh, Crn\v{c}evi\'{c}, and Alistarh]{rosa}
Mahdi Nikdan, Soroush Tabesh, Elvir Crn\v{c}evi\'{c}, and Dan Alistarh.
\newblock {R}o{SA}: Accurate parameter-efficient fine-tuning via robust adaptation.
\newblock In Ruslan Salakhutdinov, Zico Kolter, Katherine Heller, Adrian Weller, Nuria Oliver, Jonathan Scarlett, and Felix Berkenkamp (eds.), \emph{Proceedings of the 41st International Conference on Machine Learning}, volume 235 of \emph{Proceedings of Machine Learning Research}, pp.\  38187--38206. PMLR, 21--27 Jul 2024.
\newblock URL \url{https://proceedings.mlr.press/v235/nikdan24a.html}.

\bibitem[Prasanna et~al.(2020)Prasanna, Rogers, and Rumshisky]{lth_bert_sparsity_attn}
Sai Prasanna, Anna Rogers, and Anna Rumshisky.
\newblock {W}hen {BERT} {P}lays the {L}ottery, {A}ll {T}ickets {A}re {W}inning.
\newblock In Bonnie Webber, Trevor Cohn, Yulan He, and Yang Liu (eds.), \emph{Proceedings of the 2020 Conference on Empirical Methods in Natural Language Processing (EMNLP)}, pp.\  3208--3229, Online, November 2020. Association for Computational Linguistics.
\newblock \doi{10.18653/v1/2020.emnlp-main.259}.
\newblock URL \url{https://aclanthology.org/2020.emnlp-main.259/}.

\bibitem[Ramesh et~al.(2024)Ramesh, Ganapathiraman, Laradji, and Schmidt]{blockllm}
Amrutha~Varshini Ramesh, Vignesh Ganapathiraman, Issam~H. Laradji, and Mark Schmidt.
\newblock Block{LLM}: Memory-efficient adaptation of {LLM}s by selecting and optimizing the right coordinate blocks.
\newblock In \emph{OPT 2024: Optimization for Machine Learning}, 2024.
\newblock URL \url{https://openreview.net/forum?id=9pWKXz1CPD}.

\bibitem[Sakaguchi et~al.(2021)Sakaguchi, Bras, Bhagavatula, and Choi]{winogrande}
Keisuke Sakaguchi, Ronan~Le Bras, Chandra Bhagavatula, and Yejin Choi.
\newblock Winogrande: an adversarial winograd schema challenge at scale.
\newblock \emph{Commun. ACM}, 64\penalty0 (9):\penalty0 99–106, August 2021.
\newblock ISSN 0001-0782.
\newblock \doi{10.1145/3474381}.
\newblock URL \url{https://doi.org/10.1145/3474381}.

\bibitem[Sap et~al.(2019)Sap, Rashkin, Chen, Le~Bras, and Choi]{siqa}
Maarten Sap, Hannah Rashkin, Derek Chen, Ronan Le~Bras, and Yejin Choi.
\newblock Social {IQ}a: Commonsense reasoning about social interactions.
\newblock In Kentaro Inui, Jing Jiang, Vincent Ng, and Xiaojun Wan (eds.), \emph{Proceedings of the 2019 Conference on Empirical Methods in Natural Language Processing and the 9th International Joint Conference on Natural Language Processing (EMNLP-IJCNLP)}, pp.\  4463--4473, Hong Kong, China, November 2019. Association for Computational Linguistics.
\newblock \doi{10.18653/v1/D19-1454}.
\newblock URL \url{https://aclanthology.org/D19-1454/}.

\bibitem[Shaham \& Levy(2022)Shaham and Levy]{beam_search_top_p_combined}
Uri Shaham and Omer Levy.
\newblock What do you get when you cross beam search with nucleus sampling?
\newblock In Shabnam Tafreshi, Jo{\~a}o Sedoc, Anna Rogers, Aleksandr Drozd, Anna Rumshisky, and Arjun Akula (eds.), \emph{Proceedings of the Third Workshop on Insights from Negative Results in NLP}, pp.\  38--45, Dublin, Ireland, May 2022. Association for Computational Linguistics.
\newblock \doi{10.18653/v1/2022.insights-1.5}.
\newblock URL \url{https://aclanthology.org/2022.insights-1.5/}.

\bibitem[Song et~al.(2024)Song, Li, Zhang, Zhao, and Du]{sift}
Weixi Song, Zuchao Li, Lefei Zhang, Hai Zhao, and Bo~Du.
\newblock Sparse is enough in fine-tuning pre-trained large language models.
\newblock In Ruslan Salakhutdinov, Zico Kolter, Katherine Heller, Adrian Weller, Nuria Oliver, Jonathan Scarlett, and Felix Berkenkamp (eds.), \emph{Proceedings of the 41st International Conference on Machine Learning}, volume 235 of \emph{Proceedings of Machine Learning Research}, pp.\  46121--46135. PMLR, 21--27 Jul 2024.
\newblock URL \url{https://proceedings.mlr.press/v235/song24e.html}.

\bibitem[Springer et~al.(2025)Springer, Goyal, Wen, Kumar, Yue, Malladi, Neubig, and Raghunathan]{overtraining}
Jacob~Mitchell Springer, Sachin Goyal, Kaiyue Wen, Tanishq Kumar, Xiang Yue, Sadhika Malladi, Graham Neubig, and Aditi Raghunathan.
\newblock Overtrained language models are harder to fine-tune.
\newblock In \emph{Forty-second International Conference on Machine Learning}, 2025.
\newblock URL \url{https://openreview.net/forum?id=YW6edSufht}.

\bibitem[Stahlberg \& Byrne(2019)Stahlberg and Byrne]{beam_search_eos}
Felix Stahlberg and Bill Byrne.
\newblock On {NMT} search errors and model errors: Cat got your tongue?
\newblock In Kentaro Inui, Jing Jiang, Vincent Ng, and Xiaojun Wan (eds.), \emph{Proceedings of the 2019 Conference on Empirical Methods in Natural Language Processing and the 9th International Joint Conference on Natural Language Processing (EMNLP-IJCNLP)}, pp.\  3356--3362, Hong Kong, China, November 2019. Association for Computational Linguistics.
\newblock \doi{10.18653/v1/D19-1331}.
\newblock URL \url{https://aclanthology.org/D19-1331/}.

\bibitem[Sung et~al.(2021)Sung, Nair, and Raffel]{fish_mask}
Yi-Lin Sung, Varun Nair, and Colin Raffel.
\newblock Training neural networks with fixed sparse masks.
\newblock In A.~Beygelzimer, Y.~Dauphin, P.~Liang, and J.~Wortman Vaughan (eds.), \emph{Advances in Neural Information Processing Systems}, 2021.
\newblock URL \url{https://openreview.net/forum?id=Uwh-v1HSw-x}.

\bibitem[Taori et~al.(2020)Taori, Dave, Shankar, Carlini, Recht, and Schmidt]{distr_shift_imagenet}
Rohan Taori, Achal Dave, Vaishaal Shankar, Nicholas Carlini, Benjamin Recht, and Ludwig Schmidt.
\newblock Measuring robustness to natural distribution shifts in image classification.
\newblock In H.~Larochelle, M.~Ranzato, R.~Hadsell, M.F. Balcan, and H.~Lin (eds.), \emph{Advances in Neural Information Processing Systems}, volume~33, pp.\  18583--18599. Curran Associates, Inc., 2020.
\newblock URL \url{https://proceedings.neurips.cc/paper_files/paper/2020/file/d8330f857a17c53d217014ee776bfd50-Paper.pdf}.

\bibitem[Team et~al.(2024)Team, Mesnard, Hardin, Dadashi, Bhupatiraju, Pathak, Sifre, Rivière, Kale, Love, Tafti, Hussenot, Sessa, Chowdhery, Roberts, Barua, Botev, Castro-Ros, Slone, Héliou, Tacchetti, Bulanova, Paterson, Tsai, Shahriari, Lan, Choquette-Choo, Crepy, Cer, Ippolito, Reid, Buchatskaya, Ni, Noland, Yan, Tucker, Muraru, Rozhdestvenskiy, Michalewski, Tenney, Grishchenko, Austin, Keeling, Labanowski, Lespiau, Stanway, Brennan, Chen, Ferret, Chiu, Mao-Jones, Lee, Yu, Millican, Sjoesund, Lee, Dixon, Reid, Mikuła, Wirth, Sharman, Chinaev, Thain, Bachem, Chang, Wahltinez, Bailey, Michel, Yotov, Chaabouni, Comanescu, Jana, Anil, McIlroy, Liu, Mullins, Smith, Borgeaud, Girgin, Douglas, Pandya, Shakeri, De, Klimenko, Hennigan, Feinberg, Stokowiec, hui Chen, Ahmed, Gong, Warkentin, Peran, Giang, Farabet, Vinyals, Dean, Kavukcuoglu, Hassabis, Ghahramani, Eck, Barral, Pereira, Collins, Joulin, Fiedel, Senter, Andreev, and Kenealy]{gemma}
Gemma Team, Thomas Mesnard, Cassidy Hardin, Robert Dadashi, Surya Bhupatiraju, Shreya Pathak, Laurent Sifre, Morgane Rivière, Mihir~Sanjay Kale, Juliette Love, Pouya Tafti, Léonard Hussenot, Pier~Giuseppe Sessa, Aakanksha Chowdhery, Adam Roberts, Aditya Barua, Alex Botev, Alex Castro-Ros, Ambrose Slone, Amélie Héliou, Andrea Tacchetti, Anna Bulanova, Antonia Paterson, Beth Tsai, Bobak Shahriari, Charline~Le Lan, Christopher~A. Choquette-Choo, Clément Crepy, Daniel Cer, Daphne Ippolito, David Reid, Elena Buchatskaya, Eric Ni, Eric Noland, Geng Yan, George Tucker, George-Christian Muraru, Grigory Rozhdestvenskiy, Henryk Michalewski, Ian Tenney, Ivan Grishchenko, Jacob Austin, James Keeling, Jane Labanowski, Jean-Baptiste Lespiau, Jeff Stanway, Jenny Brennan, Jeremy Chen, Johan Ferret, Justin Chiu, Justin Mao-Jones, Katherine Lee, Kathy Yu, Katie Millican, Lars~Lowe Sjoesund, Lisa Lee, Lucas Dixon, Machel Reid, Maciej Mikuła, Mateo Wirth, Michael Sharman, Nikolai Chinaev, Nithum Thain, Olivier Bachem,
  Oscar Chang, Oscar Wahltinez, Paige Bailey, Paul Michel, Petko Yotov, Rahma Chaabouni, Ramona Comanescu, Reena Jana, Rohan Anil, Ross McIlroy, Ruibo Liu, Ryan Mullins, Samuel~L Smith, Sebastian Borgeaud, Sertan Girgin, Sholto Douglas, Shree Pandya, Siamak Shakeri, Soham De, Ted Klimenko, Tom Hennigan, Vlad Feinberg, Wojciech Stokowiec, Yu~hui Chen, Zafarali Ahmed, Zhitao Gong, Tris Warkentin, Ludovic Peran, Minh Giang, Clément Farabet, Oriol Vinyals, Jeff Dean, Koray Kavukcuoglu, Demis Hassabis, Zoubin Ghahramani, Douglas Eck, Joelle Barral, Fernando Pereira, Eli Collins, Armand Joulin, Noah Fiedel, Evan Senter, Alek Andreev, and Kathleen Kenealy.
\newblock Gemma: Open models based on gemini research and technology, 2024.
\newblock URL \url{https://arxiv.org/abs/2403.08295}.

\bibitem[torchtune maintainers \& contributors(2024)torchtune maintainers and contributors]{torchtune}
torchtune maintainers and contributors.
\newblock {torchtune: PyTorch's post-training library}, April 2024.
\newblock URL \url{https//github.com/pytorch/torchtune}.

\bibitem[Valipour et~al.(2023)Valipour, Rezagholizadeh, Kobyzev, and Ghodsi]{dylora}
Mojtaba Valipour, Mehdi Rezagholizadeh, Ivan Kobyzev, and Ali Ghodsi.
\newblock {D}y{L}o{RA}: Parameter-efficient tuning of pre-trained models using dynamic search-free low-rank adaptation.
\newblock In Andreas Vlachos and Isabelle Augenstein (eds.), \emph{Proceedings of the 17th Conference of the European Chapter of the Association for Computational Linguistics}, pp.\  3274--3287, Dubrovnik, Croatia, May 2023. Association for Computational Linguistics.
\newblock \doi{10.18653/v1/2023.eacl-main.239}.
\newblock URL \url{https://aclanthology.org/2023.eacl-main.239/}.

\bibitem[Wang et~al.(2025)Wang, Dimitriadis, Favero, Ortiz-Jimenez, Fleuret, and Frossard]{lines}
Ke~Wang, Nikolaos Dimitriadis, Alessandro Favero, Guillermo Ortiz-Jimenez, Fran{\c{c}}ois Fleuret, and Pascal Frossard.
\newblock Lines: Post-training layer scaling prevents forgetting and enhances model merging.
\newblock In \emph{The Thirteenth International Conference on Learning Representations}, 2025.
\newblock URL \url{https://openreview.net/forum?id=J5sUOvlLbQ}.

\bibitem[Wang et~al.(2024)Wang, Chen, Jiang, Xue, Kong, and Wu]{lora_dropout}
Sheng Wang, Liheng Chen, Jiyue Jiang, Boyang Xue, Lingpeng Kong, and Chuan Wu.
\newblock {L}o{RA} meets dropout under a unified framework.
\newblock In Lun-Wei Ku, Andre Martins, and Vivek Srikumar (eds.), \emph{Findings of the Association for Computational Linguistics: ACL 2024}, pp.\  1995--2008, Bangkok, Thailand, August 2024. Association for Computational Linguistics.
\newblock \doi{10.18653/v1/2024.findings-acl.119}.
\newblock URL \url{https://aclanthology.org/2024.findings-acl.119/}.

\bibitem[Welleck et~al.(2020{\natexlab{a}})Welleck, Kulikov, Kim, Pang, and Cho]{decoding_inconsistency}
Sean Welleck, Ilia Kulikov, Jaedeok Kim, Richard~Yuanzhe Pang, and Kyunghyun Cho.
\newblock Consistency of a recurrent language model with respect to incomplete decoding.
\newblock In Bonnie Webber, Trevor Cohn, Yulan He, and Yang Liu (eds.), \emph{Proceedings of the 2020 Conference on Empirical Methods in Natural Language Processing (EMNLP)}, pp.\  5553--5568, Online, November 2020{\natexlab{a}}. Association for Computational Linguistics.
\newblock \doi{10.18653/v1/2020.emnlp-main.448}.
\newblock URL \url{https://aclanthology.org/2020.emnlp-main.448/}.

\bibitem[Welleck et~al.(2020{\natexlab{b}})Welleck, Kulikov, Roller, Dinan, Cho, and Weston]{likelihood_max_causes_repetition}
Sean Welleck, Ilia Kulikov, Stephen Roller, Emily Dinan, Kyunghyun Cho, and Jason Weston.
\newblock Neural text generation with unlikelihood training.
\newblock In \emph{International Conference on Learning Representations}, 2020{\natexlab{b}}.
\newblock URL \url{https://openreview.net/forum?id=SJeYe0NtvH}.

\bibitem[Wortsman et~al.(2022)Wortsman, Ilharco, Kim, Li, Kornblith, Roelofs, Lopes, Hajishirzi, Farhadi, Namkoong, and Schmidt]{wise_ft}
Mitchell Wortsman, Gabriel Ilharco, Jong~Wook Kim, Mike Li, Simon Kornblith, Rebecca Roelofs, Raphael~Gontijo Lopes, Hannaneh Hajishirzi, Ali Farhadi, Hongseok Namkoong, and Ludwig Schmidt.
\newblock Robust fine-tuning of zero-shot models.
\newblock In \emph{Proceedings of the IEEE/CVF Conference on Computer Vision and Pattern Recognition (CVPR)}, pp.\  7959--7971, June 2022.

\bibitem[yang Liu et~al.(2024)yang Liu, Wang, Yin, Molchanov, Wang, Cheng, and Chen]{dora}
Shih yang Liu, Chien-Yi Wang, Hongxu Yin, Pavlo Molchanov, Yu-Chiang~Frank Wang, Kwang-Ting Cheng, and Min-Hung Chen.
\newblock Do{RA}: Weight-decomposed low-rank adaptation.
\newblock In \emph{Forty-first International Conference on Machine Learning}, 2024.
\newblock URL \url{https://openreview.net/forum?id=3d5CIRG1n2}.

\bibitem[Zellers et~al.(2019)Zellers, Holtzman, Bisk, Farhadi, and Choi]{hellaswag}
Rowan Zellers, Ari Holtzman, Yonatan Bisk, Ali Farhadi, and Yejin Choi.
\newblock {H}ella{S}wag: Can a machine really finish your sentence?
\newblock In Anna Korhonen, David Traum, and Llu{\'i}s M{\`a}rquez (eds.), \emph{Proceedings of the 57th Annual Meeting of the Association for Computational Linguistics}, pp.\  4791--4800, Florence, Italy, July 2019. Association for Computational Linguistics.
\newblock \doi{10.18653/v1/P19-1472}.
\newblock URL \url{https://aclanthology.org/P19-1472/}.

\bibitem[Zhang et~al.(2023)Zhang, Chen, Bukharin, He, Cheng, Chen, and Zhao]{adalora}
Qingru Zhang, Minshuo Chen, Alexander Bukharin, Pengcheng He, Yu~Cheng, Weizhu Chen, and Tuo Zhao.
\newblock Adaptive budget allocation for parameter-efficient fine-tuning.
\newblock In \emph{The Eleventh International Conference on Learning Representations}, 2023.
\newblock URL \url{https://openreview.net/forum?id=lq62uWRJjiY}.

\bibitem[Zhao et~al.(2023)Zhao, Gu, Varma, Luo, Huang, Xu, Wright, Shojanazeri, Ott, Shleifer, Desmaison, Balioglu, Damania, Nguyen, Chauhan, Hao, Mathews, and Li]{fsdp}
Yanli Zhao, Andrew Gu, Rohan Varma, Liang Luo, Chien-Chin Huang, Min Xu, Less Wright, Hamid Shojanazeri, Myle Ott, Sam Shleifer, Alban Desmaison, Can Balioglu, Pritam Damania, Bernard Nguyen, Geeta Chauhan, Yuchen Hao, Ajit Mathews, and Shen Li.
\newblock Pytorch fsdp: Experiences on scaling fully sharded data parallel.
\newblock \emph{Proc. VLDB Endow.}, 16\penalty0 (12):\penalty0 3848–3860, August 2023.
\newblock ISSN 2150-8097.
\newblock \doi{10.14778/3611540.3611569}.
\newblock URL \url{https://doi.org/10.14778/3611540.3611569}.

\bibitem[Zhou et~al.(2025)Zhou, Jacobs, Gadhikar, and Burkholz]{attn_small_wts}
Chao Zhou, Tom Jacobs, Advait Gadhikar, and Rebekka Burkholz.
\newblock Pay attention to small weights, 2025.
\newblock URL \url{https://arxiv.org/abs/2506.21374}.

\end{thebibliography}
